\documentclass[journal, twoside]{IEEEtran}

\usepackage{graphicx} 
\usepackage{amsmath} 
\usepackage{amssymb} 
\usepackage[T1]{fontenc}
\usepackage{microtype}
\usepackage{multirow}
\usepackage{makecell}
\usepackage{capt-of}
\usepackage{adjustbox}
\usepackage{threeparttable}
\usepackage{pifont}
\usepackage{float}
\usepackage{booktabs}
\usepackage[caption=false,font=footnotesize]{subfig}
\usepackage{svg}
\usepackage{cuted}
\usepackage{algorithm}
\usepackage{algorithmic}
\usepackage{dsfont}
\usepackage{arydshln}
\usepackage{diagbox}
\usepackage{longtable}
\usepackage{textcomp}
\usepackage{verbatim}

\usepackage{xcolor}
\definecolor{linkcolor}{HTML}{ED1C24}
\usepackage{xurl}
\usepackage[pagebackref=false,breaklinks=true,colorlinks,urlcolor=blue,citecolor=blue,linkcolor=linkcolor,bookmarks=false, ]{hyperref}
\hypersetup{pdfauthor={Rui Cao et al.}}

\newcommand{\cmark}{\ding{51}}
\newcommand{\xmark}{\ding{55}}
\newcolumntype{Y}{>{\centering\arraybackslash}X}

\author{Rui Cao\textsuperscript{1},
Jia-Hui Pan\textsuperscript{2},
Yin Zhang\textsuperscript{3},
Qiwei Meng\textsuperscript{1},
Jiapeng Li\textsuperscript{1},
Gang Wang\textsuperscript{1},
Runsong Zhu\textsuperscript{2},
Wei Chen\textsuperscript{1},
Chi-Wing Fu\textsuperscript{2},
Xieyuanli Chen\textsuperscript{1,4},
Jiangliu Wang\textsuperscript{1,5},
and Yun-Hui Liu\textsuperscript{1}\textsuperscript{\ding{41}}%
\thanks{This work is supported in part by Research Grants Council (RGC) of Hong Kong under Grant 14207324, in part by the HK RGC AoE under AoE/E-407/24-N, in part by the InnoHK initiative of the Innovation and Technology Commission of the Hong Kong Special Administrative Region Government via the Hong Kong Centre for Logistics Robotics, and in part by the CUHK T Stone Robotics Institute.
\emph{(\ding{41} Corresponding author: Yun-Hui Liu.)}}%
\thanks{\textsuperscript{1} Department of Mechanical and Automation Engineering, The Chinese University of Hong Kong, Hong Kong, China.
E-mails: rcao@mae.cuhk.edu.hk (Rui Cao); yhliu@mae.cuhk.edu.hk (Yun-Hui Liu).}%
\thanks{\textsuperscript{2} Department of Computer Science and Engineering, The Chinese University of Hong Kong, Hong Kong, China.}%
\thanks{\textsuperscript{3} The University of Hong Kong, Hong Kong, China.}%
\thanks{\textsuperscript{4} National University of Defense Technology, Hunan, China.}%
\thanks{\textsuperscript{5} Sino-German College of Intelligent Manufacturing, Shenzhen Technology University, Shenzhen, China.}%
}

\newcommand{\rui}[1]{{\color{black}{#1}}}

\newcommand{\methodname}{UA-Depth}

\usepackage{array}

\newcommand{\currentrowstyle}{}
\newcommand{\rowstyle}[1]{\gdef\currentrowstyle{#1}#1\ignorespaces}

\newcolumntype{L}{>{\global\let\currentrowstyle\relax}l}
\newcolumntype{C}{>{\currentrowstyle}c}
\newcolumntype{M}{>{\currentrowstyle}l}

\newcommand{\ms}[2]{\makecell[c]{#1\\[-0.6ex]{\scriptsize$\pm#2$}}}
\newcommand{\msbest}[2]{\makecell[c]{\textbf{\textcolor{black}{#1}}\\[-0.6ex]{\scriptsize\textcolor{black}{$\pm#2$}}}}
\newcommand{\mssecond}[2]{\makecell[c]{\underline{\textcolor{black}{#1}}\\[-0.6ex]{\scriptsize\textcolor{black}{$\pm#2$}}}}

\begin{document}

\title{\methodname: Uncertainty-Aware Depth Restoration for Material-Agnostic Object Manipulation}

\maketitle


\setlength{\stripsep}{0pt}
\begin{strip}
\vspace{-28mm}
\centering
\includegraphics[width=\textwidth]{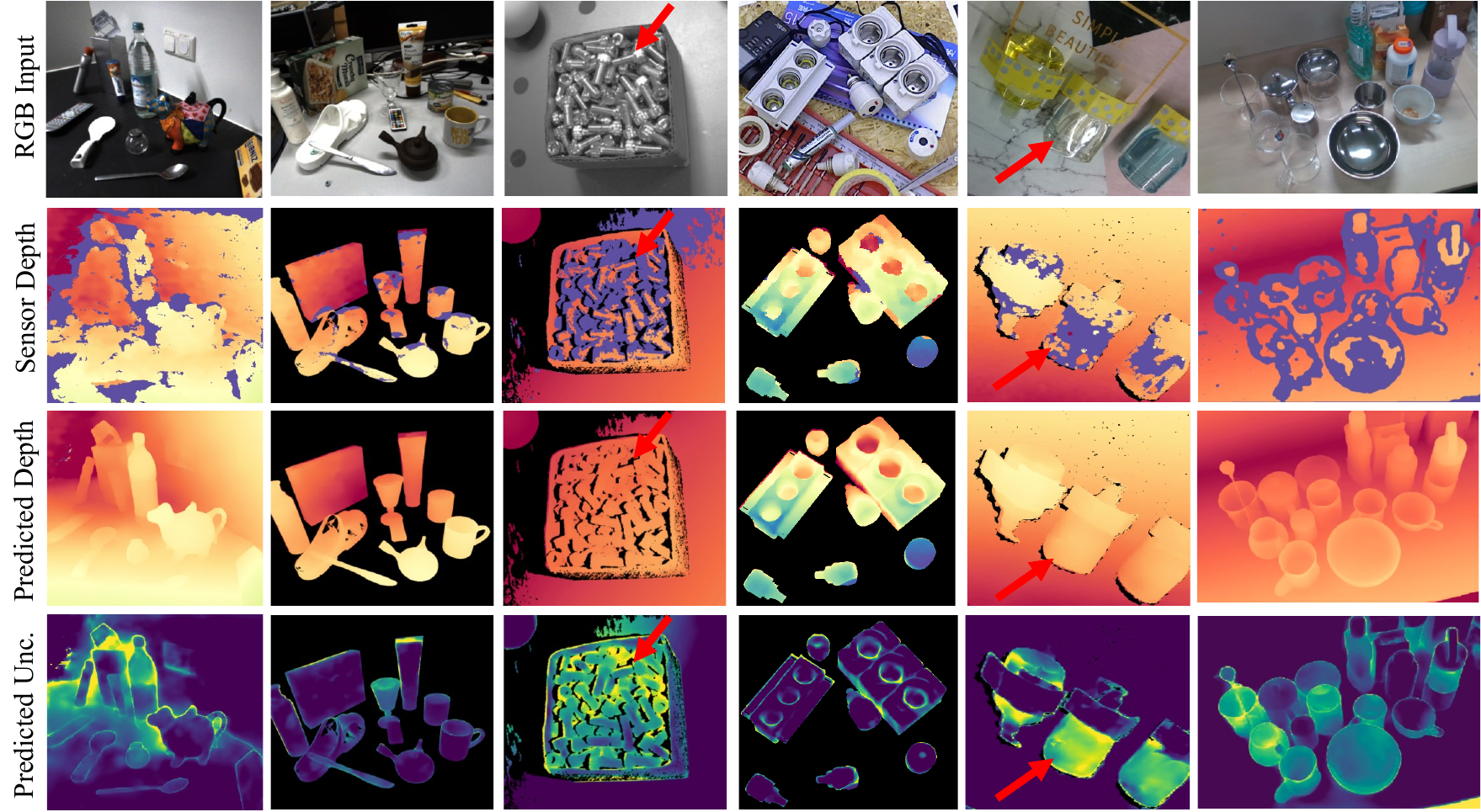}
\vspace{-4mm}
\sloppy
\captionof{figure}{
We propose \methodname, an uncertainty-aware depth restoration framework that takes a single RGB image and noisy sensor depth as input, and outputs \emph{metric} depth together with a pixel-wise uncertainty map.
As highlighted by the red arrows, \methodname\ assigns high uncertainty to unreliable regions on specular or transparent surfaces and restores more complete, accurate depth on these challenging objects.
These examples are \emph{zero-shot}: the datasets are unseen during training.
}
\fussy
\end{strip}

\begin{abstract}
Robotic manipulation requires accurate metric depth, yet commodity RGB-D cameras often fail on non-Lambertian objects such as transparent, reflective, and translucent materials, producing holes, structured noise, and biased depth. 
Most restoration methods produce only a single depth map, treating noisy and reliable regions uniformly for downstream tasks while ignoring regional uncertainty.
We present \methodname, an uncertainty-aware depth restoration framework that iteratively co-refines depth and pixel-wise uncertainty, and uses uncertainty-gated features to suppress unreliable cues for stable updates. 
To strengthen zero-shot behavior, \methodname\ leverages a frozen depth foundation model as a prior to initialize missing or unreliable depth, and applies two-stage MAD-weighted log-depth scale normalization to reduce outlier bias in relative-to-metric alignment. 
We introduce an eight-dataset zero-shot benchmark across diverse scenes, materials, and sensors. 
Across these benchmarks, \methodname\ achieves the best mean rank among representative baselines, with particularly strong gains on datasets dominated by transparent and reflective objects.
Beyond depth metrics, we demonstrate downstream improvements in category-level 6D pose estimation and 6D grasp synthesis by exploiting the predicted uncertainty. 
\emph{Code, models, and data for training and evaluation will be released upon publication.}
\end{abstract}

\begin{IEEEkeywords}
Depth restoration, uncertainty estimation, RGB-D perception, non-Lambertian objects, zero-shot generalization, robotic manipulation.
\end{IEEEkeywords}

%

\IEEEpeerreviewmaketitle

\section{Introduction}
\label{sec:introduction}
\IEEEPARstart{A}{ccurate} 3D geometry perception is essential for robotic manipulation and supports tasks such as 6D pose estimation~\cite{li2022sim,hodan2018bop}, collision-free motion planning~\cite{xu2023real}, and grasp synthesis~\cite{cao2024uncertainty,newbury2023deep,anygrasp}.
These tasks rely on metric depth to reason about object shape, free space, and contact points.
As a result, commodity RGB-D cameras are widely used on robotic platforms to provide per-pixel depth measurements~\cite{anygrasp,shen2023distilled,zhang2024toward}.

In real-world environments, commodity depth cameras often struggle with non-Lambertian materials such as transparent glass and reflective metal~\cite{ramirez2025ntire,costanzino2023learning,dai2022domain}.
On these surfaces, they can produce missing regions, structured noise, and biased measurements due to refraction, reflection, or subsurface scattering~\cite{dai2022domain,zhang2023close}.
Such corruptions are hard to address with simple filtering.
Consequently, robotic perception and manipulation pipelines that rely on reasonably accurate depth—and are typically developed and validated on mostly diffuse objects—can degrade substantially once transparent or reflective items appear in the scene~\cite{dai2022domain,ichnowski2021dexnerf}.


Researchers have proposed many depth completion and restoration methods to mitigate these sensor failures~\cite{dai2022domain,d3roma,cdm}, yet explicit, task-usable uncertainty remains underexplored.
Most approaches output a single restored depth map without explicit uncertainty estimates.
For non-Lambertian objects, errors are spatially localized and view-dependent, producing holes, bias, and boundary distortions within the same observation~\cite{dai2022domain,zhang2023close}.
Downstream modules therefore need depth values \emph{and} a per-pixel signal of where they are trustworthy.
Without uncertainty, downstream components treat all pixels/points as equally reliable, so localized depth artifacts can disproportionately affect decisions and lead to failures in tasks such as path planning~\cite{banfi2022uncertainty_ogm}, collision checking~\cite{florence2018nanomap}, and grasp synthesis~\cite{mahler2015gpgpisopt}.
\rui{These spatially varying errors also pose a challenge within restoration itself: the reliability of RGB and sensor-depth cues may need to be reassessed as depth estimates are progressively refined.}

Beyond reliability, cross-dataset generalization remains a bottleneck for depth restoration~\cite{dai2022domain,d3roma}.
Many methods are designed for sparse but reliable measurements (e.g., LiDAR-style completion)~\cite{zuo2024ogni,omni-dc} or near-clean RGB-D inputs~\cite{promptda,priorda}.
In contrast, commodity RGB-D often yields depth that is dense in format but incomplete and biased in practice, with material-dependent holes and spatially structured distortions on non-Lambertian surfaces, so performance can degrade under domain shifts.
Other approaches rely on specialized sensing (e.g., raw IR stereo)~\cite{d3roma,shi2024asgrasp} or camera-specific models~\cite{cdm}; while effective in their target setups, they introduce extra assumptions and can be sensitive to changes in camera configuration or capture conditions.

To tackle these challenges, we introduce \textbf{\methodname}, an \textbf{U}ncertainty-\textbf{A}ware \textbf{Depth} restoration framework for material-agnostic object perception in robotic manipulation.
\rui{Our key insight is to treat uncertainty as an evolving signal within depth restoration, rather than solely as an auxiliary prediction.
Under material-dependent RGB-D corruption, RGB and sensor-depth cues can be unreliable in different regions, and the uncertainty guiding refinement should be updated as the depth estimate changes.
Accordingly, \methodname\ jointly updates depth and a recurrent uncertainty state through Depth-Uncertainty Co-Refinement.
At each iteration, Uncertainty-aware Context Feature Calibration (UCFC) uses this state to recalibrate both RGB and sensor-depth context before the next depth update, allowing their contributions to adapt as refinement progresses.}

\rui{To strengthen zero-shot generalization, we leverage a frozen pretrained monocular depth model to provide a dense relative-depth prior and image features when sensor measurements are missing or corrupted.
Because this prior lacks metric scale, Aligned Depth Filling (ADF) aligns it with valid sensor depth measurements and uses it to fill missing regions, providing a dense initialization.
However, material-dependent biases in sensor depth can distort the alignment, while heterogeneous depth ranges across datasets complicate metric refinement.
Inspired by~\cite{omni-dc}, we therefore employ Robust Scale Normalization (Robust-SN), which uses two-stage MAD-weighted log-depth alignment and normalization to mitigate these effects during mixed-dataset training and inference.
These components serve complementary roles: the frozen prior and ADF support dense metric initialization, Robust-SN handles scale alignment and normalization under corrupted sensor measurements, and UCFC uses evolving uncertainty to recalibrate context throughout recurrent refinement.}

To evaluate zero-shot generalization, we further propose an eight-dataset benchmark spanning diverse scene types (e.g., tabletop~\cite{fang2022transcg}, indoor~\cite{jung2023importance,housecat6d,ycbv,tless}, and industrial~\cite{xyz-ibd,robi}), object materials (e.g., diffuse, transparent, reflective, and texture-less), and RGB-D sensors, providing a unified testbed for cross-dataset evaluation.


The main contributions of this work are as follows:
\begin{itemize}
\item We propose \methodname, an uncertainty-aware framework for metric depth restoration under non-Lambertian failures in commodity RGB-D sensing.
\item \rui{We introduce UCFC for depth restoration under material-dependent RGB-D corruption, using uncertainty updated jointly with depth to recalibrate RGB and sensor-depth context at each recurrent step.}
\item We support zero-shot depth restoration with a \rui{frozen monocular depth prior}, ADF, and Robust-SN, which provide dense metric initialization and robust log-depth alignment/normalization under noisy measurements.
\item We establish an eight-dataset benchmark for zero-shot evaluation across diverse scenes, objects, and sensors.
\end{itemize}

Experiments on three in-domain datasets show consistent overall gains in metric-depth accuracy over strong baselines, while producing informative uncertainty estimates.
On the proposed zero-shot benchmark, \methodname\ achieves the best overall performance, with particularly pronounced gains on datasets rich in non-Lambertian objects (e.g., HAMMER~\cite{jung2023importance}, HouseCat6D~\cite{housecat6d}, and TransCG~\cite{fang2022transcg}).
\rui{Ablation studies in both in-domain and zero-shot evaluations further support the core contribution of UCFC, showing the overall benefit of dynamically updated uncertainty feedback for depth refinement.}
Beyond depth metrics, we demonstrate that the predicted uncertainty is useful in the evaluated manipulation-oriented perception pipelines: uncertainty-guided sampling improves category-level 6D pose estimation, while uncertainty-guided reweighting improves mean AP in 6D grasp synthesis.
The rest of this paper is organized as follows: Sec.~\ref{sec:related_work} reviews related work; Sec.~\ref{sec:method} presents \methodname; Sec.~\ref{sec:exp} introduces the benchmark and experimental setup and reports quantitative results and ablations; Sec.~\ref{sec:app} evaluates downstream robotics applications; Sec.~\ref{sec:limitation} discusses limitations; and Sec.~\ref{sec:conclusion} concludes the paper.

\begin{figure*}[th]
    \centering
    \includegraphics[width=\linewidth]{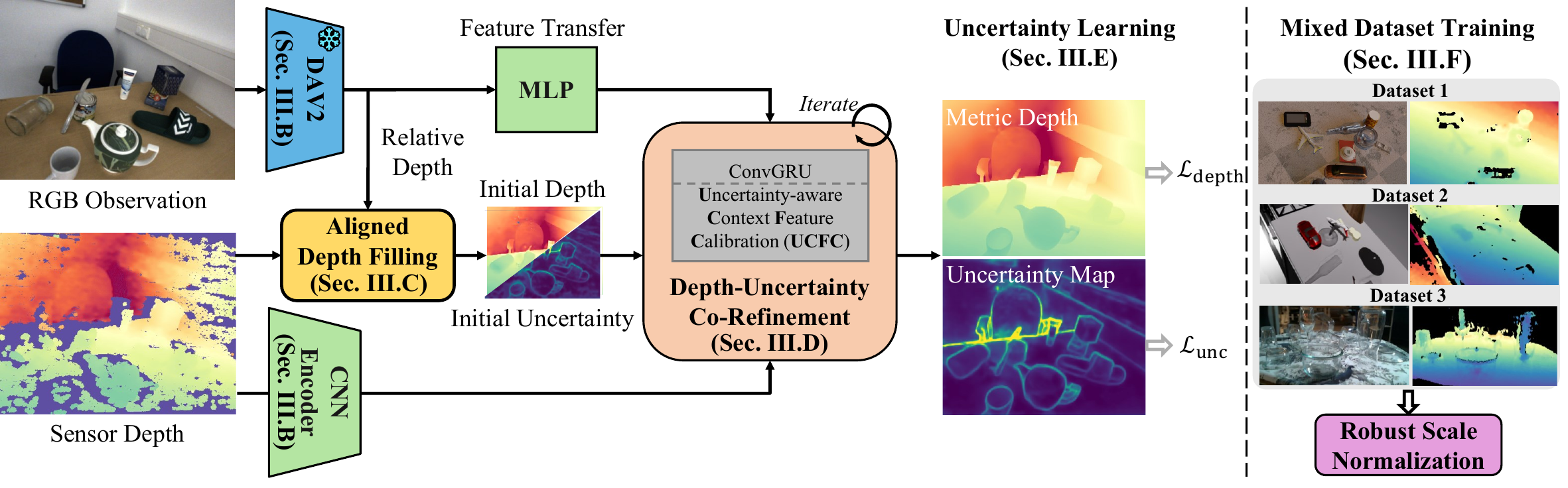}
    \vspace{-5mm}
    \caption{
    Overview of \methodname.
    Given an RGB observation and a sensor depth map, we first extract RGB features with relative depth using a frozen DAV2 backbone~\cite{depthanythingv2}.
    We then derive two feature streams for refinement: an RGB context feature obtained by projecting DAV2 features through a lightweight feature transfer MLP, and a sensor-depth context feature produced by a context network (Sec.~\ref{sec:feat_ext}).
    In parallel, the proposed ADF module (Sec.~\ref{sec:adf}) initializes metric depth and uncertainty by aligning the relative depth to valid sensor measurements and filling missing regions.
    Starting from this initialization, the Depth-Uncertainty Co-Refinement module (Sec.~\ref{sec:depth_unc_opt}) jointly updates depth and uncertainty over several iterations, and finally upsamples the predictions to the input resolution via convex sampling~\cite{teed2020raft}. 
    We discuss uncertainty modeling in Sec.~\ref{sec:unc_modeling}.
    \rui{For the mixed-dataset model, Robust-SN (Sec.~\ref{sec:robust_sn}) replaces the scale--shift alignment in ADF with robust log-depth alignment and normalization, and is applied during both training and inference.}
    }
    \label{fig:overview}
\end{figure*}

\section{Related Work}
\label{sec:related_work}
\subsection{Material-Agnostic Depth Restoration}
Material-agnostic depth restoration aims to recover accurate metric depth without relying on material-specific priors, or target-material fine-tuning, remaining robust across objects with diverse optical effects such as transparency and specularity.
SwinDR~\cite{dai2022domain} first formalizes this setting and introduces DREDS, a large-scale RGB-D dataset with diverse transparent and reflective objects. TransCG~\cite{fang2022transcg} further contributes a large-scale dataset for transparent object manipulation and proposes a transformer-based depth completion baseline that leverages cross-modal attention between RGB and sensor depth. D3RoMa~\cite{d3roma} adopts a diffusion-based framework conditioned on sensor depth or infrared (IR) stereo images to estimate metric depth, while ASGrasp~\cite{shi2024asgrasp} directly regresses dense geometry from raw IR stereo for grasp synthesis. \rui{ClearDepth~\cite{bai2024cleardepth} uses a vision transformer and Sim2Real training for stereo depth recovery of transparent objects, with evaluation on robotic grasping.} CDM~\cite{cdm} instead models camera-specific noise and synthesizes realistic depth, but requires large-scale data preparation for each sensor type. 
\rui{DCIRNet~\cite{xie2026dcirnet} combines RGB-D feature fusion with spatial-propagation-based depth refinement for transparent and reflective objects, and evaluates its application to dexterous grasping.}
\rui{SeeClear~\cite{wang2026seeclear} addresses monocular transparent-object depth estimation by generating opaque appearances for transparent regions before applying a pretrained depth estimator.}
Overall, existing approaches improve depth perception under non-Lambertian materials, but many either assume access to low-level or multi-view cues such as raw IR stereo signals~\cite{d3roma,shi2024asgrasp}, or require per-sensor data preparation and device-specific pretraining to match target cameras~\cite{cdm}.
Meanwhile, \rui{systematic zero-shot evaluation of a single model across diverse datasets, sensor types, and material conditions remains limited}~\cite{dai2022domain,d3roma,xie2026dcirnet,bai2024cleardepth,wang2026seeclear}.

\subsection{Depth Completion}
Depth completion aims to infer a dense metric depth map from RGB and sparse depth inputs. Classical learning-based approaches such as SPN~\cite{spn} refine initial depth with learnable affinity propagation, while NLSPN~\cite{nlspn} employs non-local spatial propagation to exploit long-range context. More recent methods focus on generalization: Depth Prompting (DP)~\cite{depthprompting} uses a monocular depth foundation model with prompting to adapt across sensor types, OGNI-DC~\cite{zuo2024ogni} formulates completion as differentiable optimization with neural iterations, and OMNI-DC~\cite{omni-dc} enhances cross-dataset robustness through scale-equivariant normalization in log-depth space and mixed-dataset training.
Many depth completion methods are designed around sparse, relatively clean measurements and primarily model simple, Gaussian noise~\cite{omni-dc}. As a result, they may be less effective when the input exhibits more complex degradations that often appear in commodity RGB-D sensing, such as large missing regions and material-dependent biases on non-Lambertian surfaces, which motivates depth restoration tailored to robotic manipulation.

\subsection{Uncertainty-Aware Depth Estimation}
Modeling uncertainty in depth estimation is important for safe robotic decision-making. Kendall and Gal~\cite{kendall2017uncertainties} introduce a Bayesian formulation that models depth as a heteroscedastic Gaussian and learns per-pixel aleatoric uncertainty via the loss.
Poggi~\textit{et al.}~\cite{poggi2020uncertainty} study uncertainty estimation for self-supervised monocular depth prediction and benchmark several practical uncertainty strategies.
Hornauer~\textit{et al.}~\cite{hornauer2025revisiting} demonstrate that gradient-based supervision can lead to sharper uncertainty estimates around depth discontinuities.
STEPS~\cite{STEPS} further uses uncertain pixel masking to improve self-supervised depth learning under degraded nighttime observations, while Ctrl-U~\cite{ctrlu} shows a broader use of uncertainty-aware weighting for robust conditional generation.
Uncertainty and confidence have also been explored in depth completion.
DDP~\cite{ddp} estimates dense depth posterior distributions from RGB and sparse range measurements.
pNCNN~\cite{pNCNN} explicitly distinguishes propagated confidence from probabilistic uncertainty: it notes that the propagated output confidence in normalized convolution has no probabilistic interpretation and is not significantly correlated with prediction error, and then introduces a probabilistic formulation to estimate output uncertainty for completed depth.
SparseDC~\cite{sparsedc} supervises pseudo uncertainty and uses it to guide feed-forward feature fusion between RGB and sparse depth features.
A related line of work uses confidence maps as internal reliability weights: FusionNet~\cite{fusionnet} learns confidence maps to fuse depth predictions from different branches, DeepLiDAR~\cite{deeplidar} predicts confidence masks to handle mixed LiDAR signals near foreground boundaries, and NLSPN~\cite{nlspn} uses predicted confidence to suppress unreliable pixels during non-local spatial propagation.
\rui{MUIR~\cite{wang2024matchability} predicts disparity uncertainty but gates stereo features with a soft matchability mask, while TCG-Depth~\cite{huang2026tcg} modulates refinement features using fixed confidence. In contrast, \methodname\ uses UCFC to feed evolving depth uncertainty back to recalibrate RGB and sensor-depth context at every recurrent step.}

\subsection{Generalizable Depth Estimation}
Generalizable depth estimation aims to develop models that remain robust under unseen domains~\cite{dpt}. Depth Anything V1~\cite{depthanythingv1} and V2~\cite{depthanythingv2} (DAV1/2) are monocular foundation models trained on large-scale, diverse datasets and provide strong zero-shot relative-depth priors. PromptDA~\cite{promptda} and PriorDA~\cite{priorda} further improve adaptability by conditioning such foundation models on learnable prompts or geometric priors. OMNI-DC~\cite{omni-dc} achieves strong cross-dataset performance for depth completion by operating in a normalized log-depth space.
MOMA~\cite{guo2025monocular} further suggests that relative-depth priors from DAV1/2 can be leveraged for downstream robotic grasping after metric alignment.
\rui{Using a frozen monocular depth prior}, we focus on restoring \emph{metric} depth from noisy RGB-D rather than monocular prediction alone.
Inspired by OMNI-DC~\cite{omni-dc}, we operate in a normalized log-depth space and further introduce a robust scale normalization tailored to noisy sensor depth, enabling stable mixed-dataset training across heterogeneous sensors and materials without domain-specific fine-tuning.

\begin{figure*}[t]
\centering
\includegraphics[width=\linewidth]{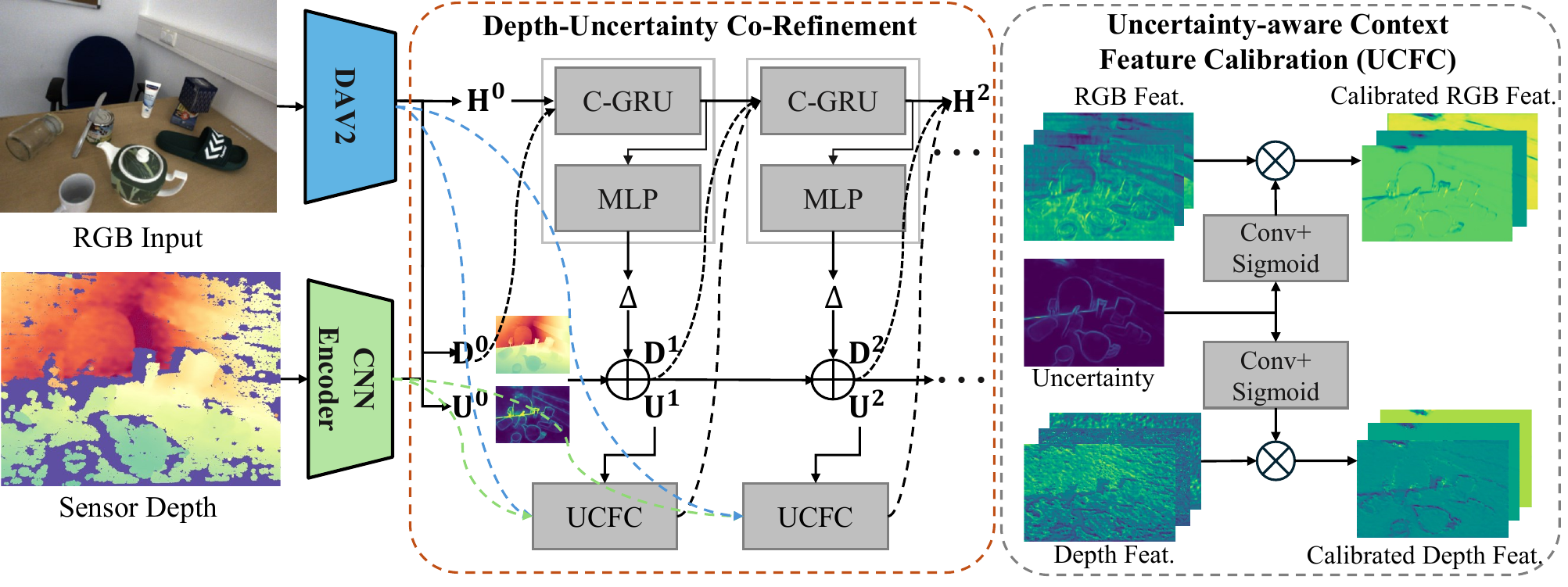}
\caption{
Detailed overview of Depth-Uncertainty Co-Refinement with Uncertainty-aware Context Feature Calibration (UCFC).
We run $T$ ConvGRU refinement steps: at step $t$, the ConvGRU updates the hidden state $\mathbf{H}^{t}$ and predicts residual updates that refine depth $\mathbf{D}^{t}$ and uncertainty $\mathbf{U}^{t}$.
UCFC is applied at each iteration to calibrate both RGB and depth context features using the predicted uncertainty, gating unreliable regions before they are fused into the ConvGRU update.
}
\label{fig:unc_optimization}
\end{figure*}

\section{Method}
\label{sec:method}
\subsection{Overview}
Given an RGB image $\mathbf{I} \in \mathbb{R}^{H \times W \times 3}$ and sensor depth $\mathbf{D}_\mathrm{sen} \in \mathbb{R}^{H \times W}$, our goal is to recover a metric depth map $\mathbf{D} \in \mathbb{R}^{H \times W}$ together with a pixel-wise uncertainty map $\mathbf{U} \in \mathbb{R}^{H \times W}$. We formulate depth restoration as an iterative optimization process, where $\mathbf{D}$ and $\mathbf{U}$ are progressively refined using the RGB image and sensor depth. An overview is shown in Fig.~\ref{fig:overview}.

Commodity RGB-D sensors often produce missing or biased depth on non-Lambertian surfaces, making refinement directly from $\mathbf{D}_\mathrm{sen}$ unstable.
To provide robust guidance, we use \rui{a frozen pretrained monocular depth model} to extract a relative depth $\mathbf{D}_\mathrm{rel}$ and RGB features $\mathbf{F}_\mathrm{img}$, and transfer $\mathbf{F}_\mathrm{img}$ into our refinement domain (Sec.~\ref{sec:feat_ext}).
Since $\mathbf{D}_\mathrm{rel}$ \rui{provides dense geometric structure but lacks metric scale}, we introduce Aligned Depth Filling (ADF) to obtain a metrically grounded starting point (Sec.~\ref{sec:adf}).
ADF aligns $\mathbf{D}_\mathrm{rel}$ to \rui{sensor-valid} pixels of $\mathbf{D}_\mathrm{sen}$, fills missing regions with the aligned prior, and outputs $(\mathbf{D}^{0}, \mathbf{U}^{0})$ for refinement.
\rui{ADF is used in both in-domain and mixed-dataset settings.
For mixed-dataset pretraining, heterogeneous depth ranges challenge metric-space refinement, while biased sensor measurements undermine simple scale estimates.
We therefore enable Robust Scale Normalization (Robust-SN; Sec.~\ref{sec:robust_sn}), which replaces the metric scale-and-shift alignment with robust log-bias estimation and applies the same ADF filling rule in normalized log-depth space, where subsequent refinement also operates.}

Starting from $(\mathbf{D}^{0}, \mathbf{U}^{0})$, our refinement stage aims to correct corrupted depth while \rui{updating the uncertainty state used to recalibrate context features at each iteration}.
We iteratively co-refine $(\mathbf{D}, \mathbf{U})$ with a ConvGRU using RGB--D context features (Sec.~\ref{sec:depth_unc_opt}).
At each iteration $k$, depth and uncertainty \textit{implicitly} couple through the recurrent
hidden state, and the \rui{recurrent uncertainty state} $\mathbf{U}^{k}$ \textit{explicitly} calibrates the context features via Uncertainty-aware Context Feature Calibration (UCFC) to suppress unreliable cues.
After $K$ iterations, $(\mathbf{D}^{K}, \mathbf{U}^{K})$ are upsampled to full resolution using convex sampling~\cite{teed2020raft}. 
We further specify the uncertainty learning objective and its error term tailored to depth restoration (Sec.~\ref{sec:unc_modeling}).


\subsection{Feature Extraction}
\label{sec:feat_ext}
\noindent\textbf{Incorporating Monocular Depth Foundation Model Prior.} 
\rui{We adopt DAV2~\cite{depthanythingv2} to extract a relative-depth prior $\mathbf{D}_\mathrm{rel}$ and RGB features $\mathbf{F}_\mathrm{img}$.
The relative-depth prior lacks metric scale and must be aligned with sensor depth for robotic manipulation.
We keep the DINOv2~\cite{oquab2024dinov2} encoder and DPT~\cite{dpt} decoder frozen.
}
In our framework, we use the feature maps at three scales:
\(\mathbf{F}_{\mathrm{img},j} \in \mathbb{R}^{C_j \times \tfrac{H}{j} \times \tfrac{W}{j}}\),
where \(j \in \{4,8,16\}\).
These multi-scale features are further transformed into task-specific representations using a multi-scale MLP:
\begin{equation}
\mathbf{F}_{\mathrm{img},j}' = \text{Conv}_j\bigl(\mathbf{F}_{\mathrm{img},j}\bigr), \quad j \in \{4,8,16\}.
\end{equation}

\noindent\textbf{Sensor Depth Feature Extraction.} 
In parallel, the sensor depth map $\mathbf{D}_{\mathrm{sen}}$ is encoded using a lightweight CNN encoder similar to RAFT-Stereo~\cite{lipson2021raft}. 
This encoder consists of residual blocks~\cite{he2016deep} interleaved with downsampling layers, and produces multi-scale depth features:
\begin{equation}
\mathbf{F}_{\mathrm{sen},j} \in \mathbb{R}^{C_j \times \tfrac{H}{j} \times \tfrac{W}{j}}, \quad j \in \{4,8,16\}.
\end{equation}

The transformed RGB features $\mathbf{F}'_{\mathrm{img}}$ and the depth features $\mathbf{F}_{\mathrm{sen}}$ are then concatenated to form the context features. 
These context features encode both appearance information from RGB and geometric information from the sensor depth, and are subsequently exploited to initialize and guide the Depth-Uncertainty Co-Refinement.

\subsection{Depth Initialization via Aligned Depth Filling}
\label{sec:adf}

We first propose ADF to initialize a reliable input for Depth-Uncertainty Co-Refinement.
Raw sensor depth $\mathbf{D}_{\mathrm{sen}}$ often contains missing values due to material-dependent reflection, refraction, or translucency. Directly using such degraded input leads to error propagation in subsequent refinement. On the other hand, the relative depth $\mathbf{D}_{\mathrm{rel}}$ predicted by a vision foundation model provides complete and smooth depth structures but lacks metric scale. To exploit the complementary advantages of the two sources, we combine them into a unified initialization: \rui{sensor-valid} values are preserved, while missing regions are replaced with the aligned relative depth.

Formally, we define the \rui{sensor-valid} set as
\begin{equation}
\label{eq:valid_set}
\mathcal{V} = \{(x,y)\mid \mathbf{D}_{\mathrm{sen}}(x,y) \rui{> d_{\min}} \}.
\end{equation}
\rui{Here $d_{\min}$ denotes the minimum valid sensor depth.}
\rui{For the default in-domain setting, }the relative depth is then aligned \rui{in metric depth} via a scale--shift transformation:
\begin{equation}
\mathbf{D}'_{\mathrm{rel}}(x,y) = s\,\mathbf{D}_{\mathrm{rel}}(x,y) + t,
\end{equation}
where $s$ and $t$ are optimized by minimizing the least-squares error over \rui{sensor-valid} pixels:
\begin{equation}
s,t = \arg\min_{s,t} \sum_{(x,y)\in \mathcal{V}} \bigl(s\,\mathbf{D}_{\mathrm{rel}}(x,y)+t-\mathbf{D}_{\mathrm{sen}}(x,y)\bigr)^2.
\end{equation}

The initialization depth $\mathbf{D}^0$ is then constructed as
\begin{equation}
\label{eq:init_depth}
\mathbf{D}^{0}(x,y) =
\begin{cases}
\mathbf{D}_{\mathrm{sen}}(x,y), & (x,y)\in \mathcal{V}, \\
\mathbf{D}'_{\mathrm{rel}}(x,y), & (x,y)\notin \mathcal{V}.
\end{cases}
\end{equation}
\rui{For mixed-dataset training, Robust-SN replaces the above scale--shift alignment with robust log-bias estimation while retaining the same ADF filling rule in normalized log-depth space (see Sec.~\ref{sec:robust_sn} for details).}

Meanwhile, we exploit the foundation model's prior to initialize a heuristic uncertainty map $\mathbf{U}^0$, \rui{which is computed for sensor-valid pixels as}
\begin{equation}
\mathbf{U}^{0}(x,y) =\frac{\lvert \mathbf{D}'_{\mathrm{rel}}(x,y)-\mathbf{D}_{\mathrm{sen}}(x,y)\rvert
}{
\rui{\max\!\left(\mathbf{D}'_{\mathrm{rel}}(x,y),\varepsilon \right)}
},
\end{equation}
\rui{where $\varepsilon$ is used for numerical stability.}
For pixels not in the \rui{sensor-valid} set $\mathcal{V}$ (i.e., \rui{$\mathbf{D}_{\mathrm{sen}}(x,y)\leq d_{\min}$}), \rui{we set} $\mathbf{U}^{0}(x,y)=1$.
Further ablation studies and analysis are provided in Sec.~\ref{sec:ablations}.

\subsection{Depth-Uncertainty Co-Refinement}
\label{sec:depth_unc_opt}
\noindent \textbf{Uncertainty-aware Context Feature Calibration (UCFC).} 
\rui{In standard ConvGRU refinement, the same context features are typically reused across iterations~\cite{lipson2021raft}, without explicitly recalibrating them according to the evolving uncertainty state.}
In our setting, \rui{this can be problematic}: sensor depth can be severely corrupted on non-Lambertian surfaces, and RGB evidence can be weak in textureless regions.
Reusing static context under such corruptions can repeatedly propagate misleading signals, amplifying errors and destabilizing refinement.
To mitigate this, we propose UCFC, which uses \rui{the recurrent uncertainty state jointly updated with depth} at each iteration to \emph{gate} the \rui{RGB and sensor-depth} context features, yielding iteration-dependent and reliability-aware guidance.

Given the \rui{recurrent uncertainty state} $\mathbf{U}^{k}$ at iteration $k$, we derive two gating maps for the sensor-depth and RGB streams, respectively, using separate convolutional layers followed by a sigmoid activation $\sigma(\cdot)$:
\begin{equation}
\mathbf{M}^{k}_{\mathrm{sen}} = \sigma\bigl(\text{Conv}_{\mathrm{sen}}(\mathbf{U}^{k})\bigr),
\quad
\mathbf{M}^{k}_{\mathrm{img}} = \sigma\bigl(\text{Conv}_{\mathrm{img}}(\mathbf{U}^{k})\bigr).
\end{equation}
The gating maps \rui{modulate the contributions of sensor-depth and RGB features using separate learned weights} through element-wise multiplication (Hadamard product $\odot$):
\begin{equation}
\tilde{\mathbf{F}}_{\mathrm{sen}}^{k} = \mathbf{M}^{k}_{\mathrm{sen}} \odot \mathbf{F}_{\mathrm{sen}}, 
\quad 
\tilde{\mathbf{F}}_{\mathrm{img}}^{k} = \mathbf{M}^{k}_{\mathrm{img}} \odot \mathbf{F}'_{\mathrm{img}},
\end{equation}
where $\tilde{\mathbf{F}}_{\mathrm{sen}}^{k}$ and $\tilde{\mathbf{F}}_{\mathrm{img}}^{k}$ are the uncertainty-gated sensor-depth and image features at iteration $k$.

\noindent \textbf{Iteration.}
We adopt an iterative refinement scheme based on multi-scale ConvGRU blocks operating at \(\tfrac{1}{r}\) of the original resolution (where \(r\) is the refine-downsample rate), which effectively reduces the computational cost of recurrent updates on high-resolution inputs.

As shown in Fig.~\ref{fig:unc_optimization}, given the initialized depth $\mathbf{D}^0$ and uncertainty $\mathbf{U}^0$, we iteratively refine them through ConvGRU updates.
Let $\mathbf{X}^{k} = [\mathbf{D}^{k}, \mathbf{U}^{k}]$ denote the two-channel input composed of the current \rui{depth and recurrent uncertainty state}, where $[\cdot,\cdot]$ denotes channel-wise concatenation.
\rui{Here, $\mathbf{U}^{k}$ denotes the unconstrained latent state initialized from $\mathbf{U}^{0}$.}
The ConvGRU update at step \rui{$k=0,\ldots,K-1$} is
\begin{align}
\mathbf{Z}^{k} &= \sigma\!\left(\text{Conv}\!\left([\rui{\mathbf{H}^{k}},\ \mathbf{X}^{k}],\, W_z\right) + \mathbf{c}_z^{k}\right), \\
\mathbf{R}^{k} &= \sigma\!\left(\text{Conv}\!\left([\rui{\mathbf{H}^{k}},\ \mathbf{X}^{k}],\, W_r\right) + \mathbf{c}_r^{k}\right), \\
\rui{\tilde{\mathbf{H}}^{k+1}} &= \tanh\!\left(\text{Conv}\!\left([\mathbf{R}^{k} \!\odot\! \rui{\mathbf{H}^{k}},\ \mathbf{X}^{k}],\, W_h\right) + \mathbf{c}_h^{k}\right), \\
\rui{\mathbf{H}^{k+1}} &= (1-\mathbf{Z}^{k}) \odot \rui{\mathbf{H}^{k}} + \mathbf{Z}^{k} \odot \rui{\tilde{\mathbf{H}}^{k+1}}.
\end{align}
where $W_z$, $W_r$ and $W_h$ are parameters of convolutional layers, $\mathbf{c}_z^{k}, \mathbf{c}_r^{k}, \mathbf{c}_h^{k}$ are gated context features $\mathbf{F}^{k} = [\tilde{\mathbf{F}}_{\mathrm{sen}}^{k}, \tilde{\mathbf{F}}_{\mathrm{img}}^{k}]$, i.e.:
\begin{equation}
( \mathbf{c}_z^{k},\ \mathbf{c}_r^{k},\ \mathbf{c}_h^{k} ) = \text{Conv}\left(\mathbf{F}^{k}\right).
\end{equation}
Finally, two prediction heads applied to the hidden state $\rui{\mathbf{H}^{k+1}}$ predict the residual depth $\Delta \mathbf{D}^{k}$ and residual uncertainty $\Delta \mathbf{U}^{k}$.
We then update
\begin{equation}
\mathbf{D}^{k+1} = \mathbf{D}^{k} + \Delta \mathbf{D}^{k}, 
\qquad 
\mathbf{U}^{k+1} = \mathbf{U}^{k} + \Delta \mathbf{U}^{k},
\end{equation}
at each iteration $k$.
\rui{The updated uncertainty state $\mathbf{U}^{k+1}$ is used to recompute the feature gates at the next iteration, closing the feedback loop between depth refinement and context calibration.}
We adopt a three-level hierarchy of ConvGRU blocks that update hidden states in a coarse-to-fine manner within each iteration, with their initial states initialized from the context features as $\mathbf{H}^{0}_{j} = \tanh(\mathbf{F}_{j})$ for $j \in \{4,8,16\}$.
After the final iteration, the refined depth $\mathbf{D}^K$ \rui{and recurrent uncertainty state $\mathbf{U}^K$ are} upsampled to the full image resolution using convex sampling~\cite{teed2020raft}.

\subsection{Uncertainty Modelling and Learning Objective}
\label{sec:unc_modeling}

In this section, we define an uncertainty learning objective and an error proxy $\mathcal{E}$ that encourages informative uncertainty estimates on object regions.
We model input-dependent \emph{aleatoric} uncertainty associated with missing, noisy, or material-corrupted RGB-D observations and ambiguous depth discontinuities. We do not estimate epistemic/model uncertainty, because the network uses a single deterministic set of parameters. Inspired by heteroscedastic uncertainty learning~\cite{kendall2017uncertainties}, we predict a per-pixel uncertainty map $\mathbf{U}$ and optimize it with the following objective:
\begin{equation}
\label{eq:l_unc_general}
\mathcal{L}_{\text{unc}}(\mathbf{D}, \hat{\mathbf{D}}, \mathbf{U}) 
= \frac{\lambda_1 \,\mathcal{E}(\mathbf{D}, \hat{\mathbf{D}})}{\mathbf{U}^2} + \lambda_2 \log \mathbf{U},
\end{equation}
where ${\mathbf{D}}$ denotes the GT depth, and $\hat{\mathbf{D}}$ denotes the predicted depth after the refinement; $\mathcal{E}(\mathbf{D}, \hat{\mathbf{D}})$ is an error proxy computed from the depth residuals between $\mathbf{D}$ and $\hat{\mathbf{D}}$.
The $\log \mathbf{U}$ term discourages the trivial solution $\mathbf{U}\!\to\!\infty$, and $\lambda_1$ and $\lambda_2$ are balance parameters.


\begin{figure}[t]
    \centering
    \includegraphics[width=\linewidth]{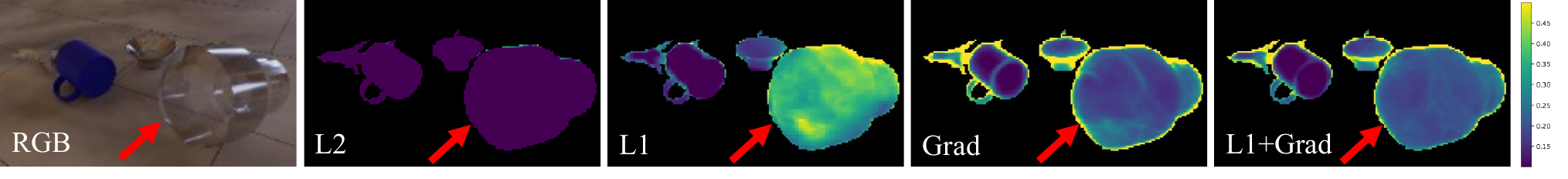}
    \caption{Effect of different choices of error proxy $\mathcal{E}$ for learning the uncertainty map on samples of DREDS~\cite{dai2022domain}. From left to right: RGB input, uncertainty maps trained with $L_2$, $L_1$, Grad, and $L_1{+}$Grad losses. Red arrows highlight a transparent object where $L_1$ tends to emphasize interior details, Grad emphasizes only the boundary, and $L_1{+}$Grad responds to both. All uncertainty maps are normalized with the same per-image percentile scaling for visualization.
    }
    \label{fig:unc_models}
\end{figure}

\noindent \textbf{Error Proxy for Uncertainty Supervision.}  
Eq.~\eqref{eq:l_unc_general} requires specifying the error proxy $\mathcal{E}(\mathbf{D}, \hat{\mathbf{D}})$ used to learn uncertainty.
A standard choice is the $L_2$ error~\cite{kendall2017uncertainties}, but in our scenes it is often dominated by large residuals around strong depth jumps (e.g., object--background transitions and severe outliers), 
which can cause the learned uncertainty to be dominated by a small set of large outliers, reducing its sensitivity to fine object geometry and mild discontinuities.
Using an $L_1$ error is more robust to sensor noise and stabilizes training, yet $L_1$ alone provides weak structural constraints and may blur object boundaries.

\begin{figure}[t]
    \centering
    \includegraphics[width=\linewidth]{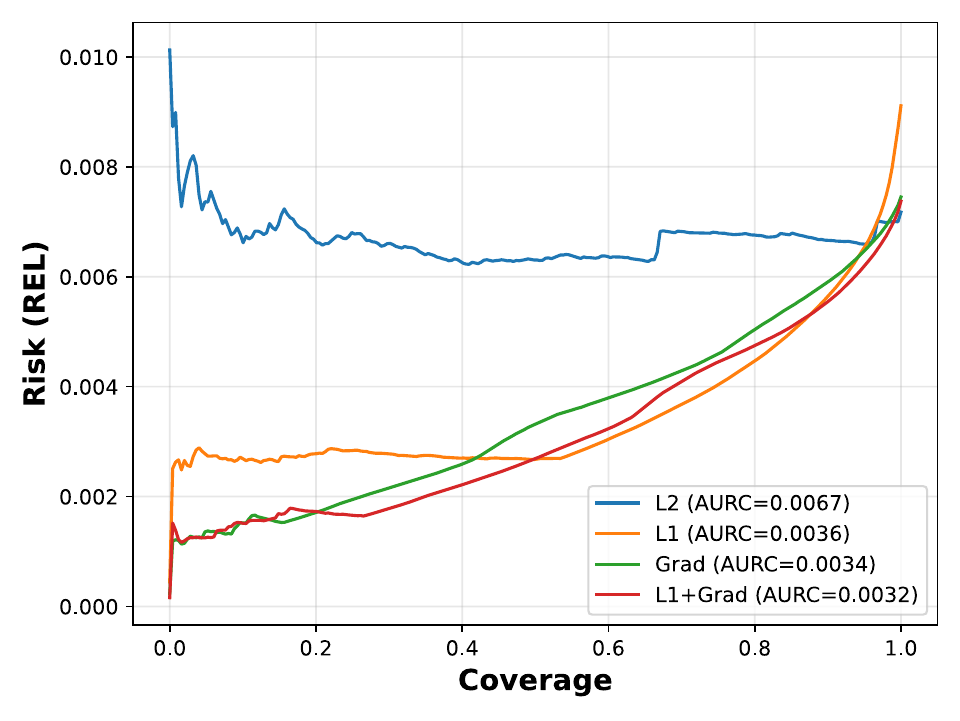}
    \caption{Risk--coverage curves on object pixels of the DREDS test split~\cite{dai2022domain} for uncertainty learned with different error proxies.
    Pixels are ranked by predicted uncertainty and retained from low to high uncertainty; we plot mean REL (risk) versus retained fraction (coverage).
    Lower AURC indicates better uncertainty ranking.}
    \label{fig:risk_coverage_obj}
\end{figure}

To address this limitation, we define $\mathcal{E}$ as a combination of a pixel-wise term and a gradient-matching term:
\begin{equation}
\label{eq:l_unc_final}
\mathcal{E}(\mathbf{D}, \hat{\mathbf{D}})
= \|\mathbf{D} - \mathrm{sg}(\hat{\mathbf{D}})\|_1 
+ \alpha \,\mathcal{L}_{\text{grad}}(\mathbf{D}, \mathrm{sg}(\hat{\mathbf{D}})),
\end{equation}
where $\alpha$ balances pixel-wise and structural contributions.
Here, $\mathcal{L}_{\text{grad}}$ enforces gradient consistency~\cite{ranftl2020towards,depthanythingv2}:
\begin{equation}
\label{eq:gradient_matching}
\mathcal{L}_{\text{grad}}(\mathbf{D},\hat{\mathbf{D}}) 
= \|\nabla_x \mathbf{D} - \nabla_x \hat{\mathbf{D}}\|_1 
+ \|\nabla_y \mathbf{D} - \nabla_y \hat{\mathbf{D}}\|_1,
\end{equation}
and $\mathrm{sg}(\cdot)$ is the stop-gradient operator that blocks gradients from flowing into the depth branch, ensuring that $\mathbf{U}$ is learned to match the residual-based proxy $\mathcal{E}$ rather than directly influencing the depth branch.
Thus, $\mathcal{E}$ shapes only the uncertainty target, while depth is still supervised solely by $\mathcal{L}_{\text{depth}}$.
\rui{The effect of stop-gradient is evaluated in Appendix Table~\ref{tab:stop_grad_hammer}.}
Since $\mathcal{E}$ combines residual and gradient terms rather than using a standard squared Gaussian residual, the raw values of $\mathbf{U}$ are \textbf{not} interpreted as calibrated posterior standard deviations.
\rui{The scale of $\mathbf{U}$ also depends on the error proxy $\mathcal{E}$ and the loss-weight ratio $\lambda_1/\lambda_2$. To assess whether this learned scale remains meaningful under domain shift, we evaluate its calibration on unseen datasets. Although a source-domain temperature can adjust its overall magnitude, calibration is not uniform across target datasets and depth ranges: NLL improves on seven of eight zero-shot datasets but worsens on GraspNet-Trans (GN-Trans), while depth-bin ECE varies across datasets (Appendix Table~\ref{tab:unc_calibration}).}

As illustrated in Fig.~\ref{fig:unc_models}, $L_1$ tends to highlight material-induced degradations (e.g., transparent/reflective regions) but often underemphasizes boundary-related errors, whereas the gradient term emphasizes contours but is less sensitive to interior degradations.
Our $L_1{+}$Grad formulation (Eq.~\eqref{eq:l_unc_final}) complements these two behaviors, enabling the model to capture both types of uncertainty patterns in a unified manner.
We further assess whether the predicted uncertainty provides a meaningful ranking of object pixels using risk--coverage curves (Fig.~\ref{fig:risk_coverage_obj}).
Here, \emph{coverage} is the fraction of pixels retained after ranking by uncertainty and keeping the lowest-uncertainty ones, and \emph{risk} is the mean absolute relative error (REL) on the retained pixels.
A better uncertainty estimate discards unreliable pixels first, yielding a lower-risk curve at the same coverage.
As shown, an $L_2$ proxy ranks pixels poorly, while our $L_1{+}$Grad proxy yields the most informative ranking and the best overall risk--coverage behavior.
\rui{An additional uncertainty thresholding analysis at $50\%$ pixel coverage is provided in Appendix Table~\ref{tab:unc_threshold}.}

\noindent \textbf{Total Loss Function.}
We employ an $L_1$ loss and a gradient matching loss (Eq.~\eqref{eq:gradient_matching}) to supervise depth prediction, i.e.,
\begin{equation}
\mathcal{L}_{\text{depth}} = \mathcal{L}_1 + \mathcal{L}_{\text{grad}}.
\end{equation}
Combining this with the uncertainty supervision loss $\mathcal{L}_{\text{unc}}$, we follow the iterative supervision strategy of~\cite{lipson2021raft} and define the total loss as
\begin{equation}
\label{eq:l_total}
\mathcal{L}_{\text{total}} = 
\sum_{k=1}^{K} \gamma^{K-k}\big(\mathcal{L}_{\text{depth}}^{k} + \mathcal{L}_{\text{unc}}^{k}\big),
\end{equation}
where $K$ is the number of iterations and $\gamma$ is a decay factor. 
This formulation applies exponentially increasing weights to later iterations, thereby progressively emphasizing accurate depth and uncertainty refinement as the recurrent updates proceed.

\subsection{Robust Scale Normalization for Mixed-Dataset Training}
\label{sec:robust_sn}
Depth ranges vary drastically across datasets, for example, from millimeter-level industrial bin-picking scenes~\cite{robi,xyz-ibd} to meter-scale indoor environments~\cite{jung2023importance,housecat6d}. Direct supervision or inference in metric space across such heterogeneous ranges can introduce scale bias and instability.

\begin{figure}[t]
    \centering
    \includegraphics[width=\linewidth]{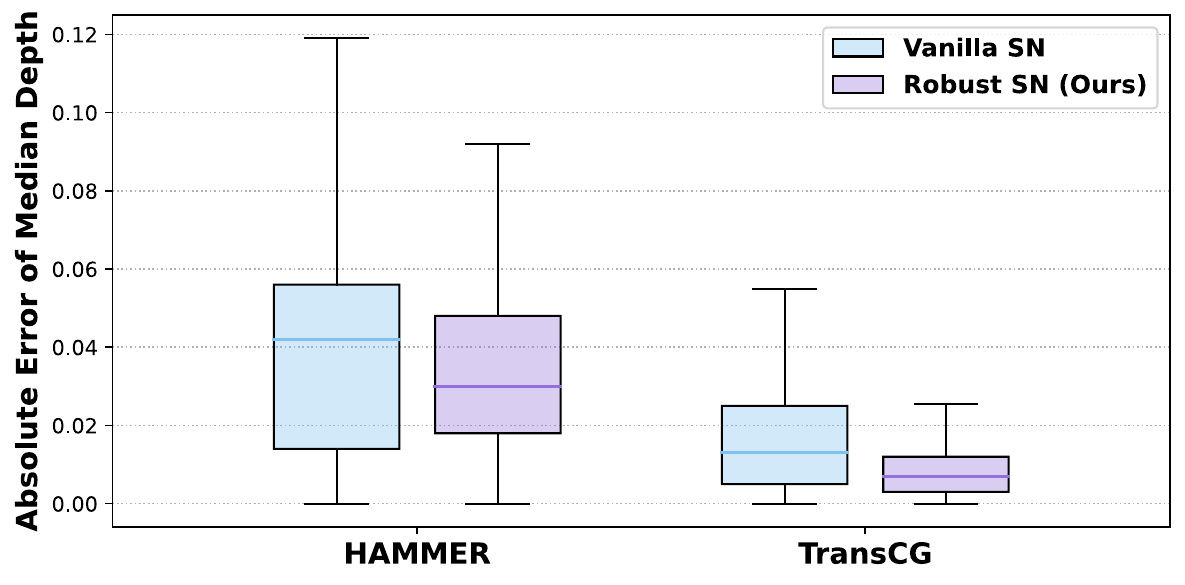}
    \caption{Box plots of the per-frame mean absolute error of the estimated median depth (in meters) on testset of HAMMER~\cite{jung2023importance} and TransCG~\cite{fang2022transcg}. We observe that the vanilla scale normalization used in OMNI-DC~\cite{omni-dc} is sensitive to noisy or partially missing sensor depth, which is common in datasets containing transparent or reflective objects, leading to larger errors and higher variance compared to our robust scale normalization.}
\label{fig:sn_motivation}
\end{figure}

Prior work, such as OMNI-DC~\cite{omni-dc}, addresses this by normalizing depth in log space,
\begin{equation}
\label{eq:raw_sn}
\tilde{\mathbf{D}} = \log(\mathbf{D}) - \log(\operatorname{median}(\mathbf{D})).
\end{equation}
This encourages scale equivariance learning in depth completion. In our restoration setting, however, sensor depth can be corrupted or partially missing, making the per-frame median in Eq.~\eqref{eq:raw_sn} sensitive to noise and causing scale drift, as shown in Fig.~\ref{fig:sn_motivation}. We therefore extend scale normalization to robustly anchor the metric scale while aligning the \rui{relative-depth} prior to the noisy sensor input.

\noindent\textbf{Normalization in Log-depth Space.}
\rui{For mixed-dataset training, Robust-SN replaces the metric scale--shift alignment of Sec.~\ref{sec:adf} with a multiplicative log-depth bias while retaining the ADF filling rule.} 
For each frame, we compute
\begin{equation}
\log\mathbf{D}_{\mathrm{rel}}' = \log\mathbf{D}_{\mathrm{rel}} + a,
\end{equation}
where $a$ denotes the log-depth bias that shifts the relative prior toward the noisy sensor depth.
We also estimate an anchor depth $s_{\text{anc}}$ to normalize both sources,
\begin{align}
\tilde{\mathbf{D}}_{\mathrm{sen}} &= \log\mathbf{D}_{\mathrm{sen}} - \log s_{\text{anc}}, \\
\tilde{\mathbf{D}}_{\mathrm{rel}} &= \log\mathbf{D}_{\mathrm{rel}}' - \log s_{\text{anc}} .
\end{align}
\rui{Applying the ADF filling rule in Eq.~\eqref{eq:init_depth} to these normalized depths yields $\tilde{\mathbf{D}}^0$.}
Operating in log-depth decouples metric scale from scene content and is essential for training on datasets with differing depth ranges.

\noindent\textbf{Two-stage Robust Estimation.}
Simple medians, as used in OMNI-DC, are fragile under heterogeneous sensor noise and non-Lambertian artifacts. We therefore estimate $(a, s_{\text{anc}})$ through a two-stage robust procedure. A provisional log-bias,
\begin{equation}
a_0 =
\operatorname{median}_{(x,y)\in \rui{\mathcal{V}_{\mathrm{scr}}}}
\bigl(\log\mathbf{D}_{\mathrm{sen}}(x,y) - \log\mathbf{D}_{\mathrm{rel}}(x,y)\bigr),
\end{equation}
provides an initial alignment, \rui{where $\mathcal{V}_{\mathrm{scr}}\subseteq\mathcal{V}$ denotes the screened sensor-valid pixels described in Sec.~\ref{sec:implementation_details}. We set $a_0=0$ if fewer than 16 candidates remain.} Residuals under this alignment,
\begin{equation}
r(x,y)=\Big|\log\mathbf{D}_{\mathrm{sen}}(x,y)-(\log\mathbf{D}_{\mathrm{rel}}(x,y)+a_0)\Big|,
\end{equation}
are summarized using the Median Absolute Deviation (MAD):
\begin{align}
m_r &= \operatorname{median}_{(x,y)\in\mathcal{V}} r(x,y),\\
\tau &= \rui{\kappa\max\!\left(\operatorname{median}_{(x,y)\in\mathcal{V}}|r(x,y)-m_r|,\varepsilon
\right)}
\end{align}
Normalized residuals define per-pixel weights,
\begin{equation}
w(x,y)=\exp\!\bigl(-r(x,y)/\tau\bigr),
\end{equation}
such that pixels where prior and sensor agree contribute more strongly. Final estimates of log-bias and anchor depth are obtained via weighted medians:
\begin{align}
a &= \operatorname{median}^{w}\!\bigl(\log\mathbf{D}_{\mathrm{sen}} - \log\mathbf{D}_{\mathrm{rel}}\bigr), \\
s_{\text{anc}} &= \operatorname{median}^{w}\!\bigl(\mathbf{D}_{\mathrm{sen}}\bigr).
\end{align}

All ConvGRU refinement operates in the normalized log-depth domain, and metric depth is recovered only at the output:
\begin{equation}
\mathbf{D}(x,y)
= s_{\text{anc}} \cdot \exp\!\bigl(\tilde{\mathbf{D}}(x,y)\bigr).
\end{equation}
Note that in mixed-dataset training, the total training loss in Eq.~\eqref{eq:l_total} is computed in the metric depth domain.
\rui{The estimation of $a_0$, the choice of $\kappa$, and numerical safeguards for degenerate cases are detailed in Sec.~\ref{sec:implementation_details}.}

\begin{table*}[t]
\centering
\caption{Training/testing datasets used in zero-shot generalization evaluation. All dataset statistics are taken from the corresponding original papers unless otherwise specified.}
\label{tab:datasets_overview}
\setlength{\tabcolsep}{6pt}
\begin{tabular}{l l l l l l}
\toprule
Split & Dataset & Size & Object & Scene & Depth Sensor \\
\midrule
\multirow{3}{*}{Training}
& DREDS-CatKnown~\cite{dai2022domain} & 100K & 1536 objects, 7 categories & $\sim$74 scenes & Synthetic \\
& ClearPose~\cite{chen2022clearpose} & 222K & 63 objects, 7 categories & 24 scenes & RealSense L515 \\
& HISS~\cite{d3roma} & 65K & 350 objects & 160 scenes & Synthetic \\
\midrule
\multirow{8}{*}{Testing}
& HAMMER~\cite{jung2023importance} & 1414 & 20 objects, 8 categories  & 9 scenes & RealSense D435 \\
& HouseCat6D~\cite{housecat6d} & 2929 & 50 objects, 10 categories & 5 scenes & RealSense D435  \\
& TransCG~\cite{fang2022transcg} & 11K & 46 objects & 52 scenes & RealSense D435 \\
& \rui{GN-Trans} & 23K & 76 objects & 90 scenes & Synthetic \\
& YCB-V~\cite{ycbv} & 20K & 21 objects & 12 scenes & Xtion Pro Live RGB-D \\
& T-LESS~\cite{tless} & 10K & 30 objects & 20 scenes & Primesense Carmine 1.09 \\
& XYZ-IBD~\cite{xyz-ibd} & 45 & 15 objects & 15 scenes & XYZ Robotic DLP \\
& ROBI~\cite{robi} & 3129 & 7 objects & 63 scenes & Ensenso N35 \\
\bottomrule
\end{tabular}

\end{table*}

Compared with vanilla scale normalization, our formulation additionally aligns a dense relative prior to noisy sensor depth and estimates normalization parameters via a robust MAD-based procedure, which reduces the influence of corrupted readings on scale estimation and improves cross-dataset consistency and generalization within a single model.
\rui{The inference procedure is summarized in the Appendix.}

\section{Experiments}
\label{sec:exp}
In this section, we first describe implementation details (Sec.~\ref{sec:implementation_details}) and the evaluation protocol (Sec.~\ref{sec:datasets_metrics}). We then report in-domain results to evaluate the model’s fitting ability (Sec.~\ref{sec:in_domain_eval}), followed by zero-shot generalization across eight diverse benchmarks (Sec.~\ref{sec:zs_results}). Finally, ablation studies isolate the contribution of each technical innovations (Sec.~\ref{sec:ablations}).

\subsection{Implementation Details}
\label{sec:implementation_details}
We implement our framework in PyTorch and train it on NVIDIA RTX~3090 GPUs.
The DAV2 backbone is initialized from the official pretrained ViT-L checkpoint and kept frozen throughout training to preserve its generalization ability.
\rui{As DAV2 predicts a disparity-like relative depth map $\mathbf{q}$, with larger values corresponding to nearer surfaces, we obtain the relative-depth prior as $\mathbf{D}_{\mathrm{rel}}(x,y)=1/\mathbf{q}(x,y)$ for $\mathbf{q}(x,y)>0$, and zero otherwise.}
We perform \(K{=}5\) recurrent ConvGRU updates per sample with $\gamma = 0.9$, operating at \(\tfrac{1}{r}\) of the input resolution, where the refine-downsample rate \(r\) is set to 2 for \(320\times180\) inputs and 4 for \(518\times518\) inputs.
The valid depth range is restricted to \([0.2,\,5.0]\) meters.
\rui{For ADF, the sensor-valid set is defined by $\mathbf{D}_{\mathrm{sen}}>d_{\min}$, with $d_{\min}=0.2$ m.}
\rui{To estimate the provisional log-bias $a_0$ in Robust-SN, we retain sensor-valid pixels with fully valid $3\times3$ neighborhoods and sensor log-depth gradients below the 80th percentile. Their log-depth differences are clipped to the 20th--80th percentile interval before taking the median; if fewer than 16 candidate pixels remain, we set $a_0=0$.
In the second stage, the final log-bias $a$ and anchor scale $s_{\mathrm{anc}}$ are estimated using weighted medians over all sensor-valid pixels, without applying the first-stage filtering.}
\rui{
We set $\kappa=1.5\times1.4826=2.2239$ and $\varepsilon=10^{-6}$, applying the numerical floor to MAD before computing $\tau$.
If the total weight is at most $\varepsilon$, weighted medians fall back to ordinary medians.
When no valid sensor pixels are available, we set $a=0$ and use the clamped median of the monocular depth prior as $s_{\mathrm{anc}}$.
Logarithmic inputs and relative-error denominators are clamped from below by $\varepsilon$ for numerical stability.}

\rui{For uncertainty supervision (Eqs.~\eqref{eq:l_unc_general} and~\eqref{eq:l_unc_final}), we set $\lambda_1=0.5$, $\lambda_2=0.2$, and $\alpha=1$. We selected $\lambda_1$ and $\lambda_2$ empirically to balance the numerical scales of the two terms in Eq.~\eqref{eq:l_unc_general}, and kept all three coefficients fixed across datasets. The error proxy is computed from the detached depth prediction, while the depth branch is supervised by $\mathcal{L}_{\mathrm{depth}}$.}
Optimization uses the AdamW optimizer (\(\beta_1{=}0.9,\ \beta_2{=}0.999\)) with a base learning rate of \(1\times10^{-4}\), a minimum learning rate of \(2\times10^{-5}\), and weight decay of \(0.01\).
A polynomial decay schedule is applied to the learning rate.
We parameterize uncertainty as \rui{$\mathbf{U}=\max\bigl(\mathrm{softplus}(\tilde{\mathbf{U}}),\mathbf{U}_{\min}\bigr)$} with $\mathbf{U}_{\min}=0.1$ to ensure positivity and numerical stability, where $\mathrm{softplus}(x)=\log(1+\exp(x))$ is applied element-wise.
\rui{Here, $\tilde{\mathbf{U}}$ denotes the upsampled recurrent uncertainty state $\mathbf{U}^{k}$ defined in Sec.~\ref{sec:depth_unc_opt}. 
We apply the positivity mapping to each uncertainty prediction for the loss in Eq.~\eqref{eq:l_total}.}
For mixed-dataset training, in each epoch, we randomly sample 50K images from each training dataset and merge them for optimization to balance data sources.
We train for 30 epochs with batch size 4 using \(518\times518\) RGB--D inputs, which takes about 4 days on 6 RTX 3090 GPUs.
Data augmentation is applied only to DREDS following~\cite{dai2022domain}, while no augmentation is used for ClearPose and HISS.

\subsection{Datasets and Metrics}
\label{sec:datasets_metrics}
\noindent\textbf{Datasets.}
We evaluate \methodname\ in two settings: in-domain and zero-shot generalization.
\textbf{In-domain:} we train a separate model on the official training split of each dataset (HAMMER~\cite{jung2023importance}, DREDS~\cite{dai2022domain}, and ClearPose~\cite{chen2022clearpose}) and report results on the corresponding test split. \textbf{Zero-shot:} following the mixed-dataset pretraining protocol of~\cite{d3roma}, we pretrain on DREDS-CatKnown~\cite{dai2022domain}, ClearPose~\cite{chen2022clearpose}, and HISS~\cite{d3roma}, and evaluate \emph{directly} on the test sets of HAMMER~\cite{jung2023importance}, HouseCat6D~\cite{housecat6d}, TransCG~\cite{fang2022transcg}, GraspNet-Trans, XYZ-IBD~\cite{xyz-ibd}, ROBI~\cite{robi}, YCB-V~\cite{ycbv}, and T-LESS~\cite{tless} without any fine-tuning of \methodname.
In this paper, zero-shot means that \methodname\ is not trained or adapted on the target benchmarks. 
This does not rule out possible upstream exposure of the frozen DAV2 prior to semantically related public images.
As shown in Table~\ref{tab:datasets_overview}, we select these target benchmarks to cover complementary test settings, including different object categories, object materials, and sensors.
We additionally build GraspNet-Trans for manipulation-centric transfer: it shares the tabletop grasping context of GraspNet-1B~\cite{graspnet1b} but introduces transparent objects and challenging material effects, allowing us to probe robustness under non-Lambertian corruptions in a grasping-oriented setting.
Appendix~\ref{app:zs_benchmark} provides additional object-overlap and sensor-depth analyses of the testing datasets.

For YCB-V~\cite{ycbv} and T-LESS~\cite{tless}, which do not provide per-pixel GT metric depth, we evaluate on object pixels using metric depth rendered from the provided 6D poses and CAD models.
For HouseCat6D~\cite{housecat6d}, we use the provided object-region GT metric depth.
Additional dataset preparation details and analyses are provided in the Appendix~\ref{app:zs_benchmark}.

\noindent\textbf{Metrics.}
Following prior work~\cite{dai2022domain,d3roma}, we report standard metric depth accuracy metrics over the evaluation pixel set $V$.
Let $d$ and $d^*$ denote predicted and GT depth at a valid pixel $(x,y)\in V$, respectively.
We compute mean absolute error $\text{MAE}=\frac{1}{|V|}\sum_{(x,y)\in V}\lvert d-d^*\rvert$, root mean squared error $\text{RMSE}=\sqrt{\frac{1}{|V|}\sum_{(x,y)\in V}(d-d^*)^2}$, and mean absolute relative error $\text{REL}=\frac{1}{|V|}\sum_{(x,y)\in V}\frac{\lvert d-d^*\rvert}{d^*}$.
Threshold accuracy $\delta_\tau=\frac{1}{|V|}\sum_{(x,y)\in V}\mathbf{1}\!\left(\max\!\left(\frac{d}{d^*},\frac{d^*}{d}\right)<\tau\right)$ reports the fraction of pixels within tolerance $\tau\in\{1.05,1.10,1.25\}$ (higher is better).

For depth uncertainty metrics, we follow the sparsification protocol~\cite{poggi2020uncertainty,hornauer2025revisiting} and report AUSE (Area Under the Sparsification Error; lower is better, closer to oracle) and AURG (Area Under the Random Gain; higher is better, larger gain over random), computed by sparsifying pixels according to predicted uncertainty and measuring the resulting curves under REL, RMSE, and $\delta_{1.05}$.
More details about the metric definitions are provided in the Appendix~\ref{app:metrics}.

\begin{table}[t]
\centering
\caption{Quantitative results on HAMMER~\cite{jung2023importance}. Metrics are computed over \textbf{all valid} pixels.}
\resizebox{\linewidth}{!}{

\begin{tabular}{L C C C C C C}
\toprule
Method & RMSE$\downarrow$ & REL$\downarrow$ & MAE$\downarrow$ 
& $\delta_{1.05}\uparrow$ & $\delta_{1.10}\uparrow$ & $\delta_{1.25}\uparrow$ \\
\midrule
SwinDR~\cite{dai2022domain} & 0.0433 & 0.0364 & 0.0282 & 78.13 & 90.06 & 98.24 \\
$\text{D}^3\text{RoMa}$~\cite{d3roma} & 0.0405 & 0.0494 & 0.0265 & 76.95 & 91.26 & 98.22 \\
DP~\cite{depthprompting} & 0.0587 & 0.0448 & 0.0274 & 82.98 & 89.97 & 95.89 \\
OGNI-DC~\cite{zuo2024ogni} & 0.0491 & 0.0338 & 0.0242 & 82.17 & 92.73 & 98.44 \\
PromptDA~\cite{promptda} & 0.0523 & 0.0546 & 0.0317 & 69.73 & 85.89 & 95.39 \\
PPFT~\cite{ikemura2024robust} & \underline{0.0351} & \underline{0.0302} & \underline{0.0201} & \underline{85.23} & \underline{94.23} & \underline{98.83} \\
\midrule
\rowstyle{\color{black}}
Ours 
& \msbest{0.0334}{0.0051}
& \msbest{0.0236}{0.0004}
& \msbest{0.0174}{0.0024}
& \msbest{89.78}{0.18}
& \msbest{96.72}{0.38}
& \msbest{99.48}{0.08} \\
\bottomrule
\end{tabular}
}
\label{tab:hammer_results}
\end{table}

\begin{table}[t]
\centering
\caption{Quantitative results on DREDS-CatKnown~\cite{dai2022domain}. Metrics are computed over \textbf{object} pixels.}
\resizebox{\linewidth}{!}{



\begin{tabular}{L C C C C C C}
\toprule
Methods & \makecell{RMSE$\downarrow$} & \makecell{REL$\downarrow$} & \makecell{MAE$\downarrow$}
& \makecell{$\delta_{1.05}$$\uparrow$} & \makecell{$\delta_{1.10}$$\uparrow$} & \makecell{$\delta_{1.25}$$\uparrow$} \\
\midrule
NLSPN~\cite{nlspn}       & 0.010  & 0.009  & 0.006  & 97.48 & 99.51 & 99.97 \\
LIDF~\cite{zhu2021rgb}   & 0.016  & 0.018  & 0.011  & 93.60 & 98.71 & 99.92 \\
SwinDR~\cite{dai2022domain} & 0.010  & 0.008  & 0.005  & 98.04 & \textbf{99.62} & \textbf{99.98} \\
ASGrasp~\cite{shi2024asgrasp} & \textbf{0.007} & \textbf{0.006} & \textbf{0.004} & - & - & - \\
$\text{D}^3$RoMa~\cite{d3roma} & 0.0101 & 0.0075 & 0.0050 & \underline{98.11} & 99.49 & \underline{99.97} \\

\rowstyle{\color{black}}
PromptDA~\cite{promptda} & 0.0138 & 0.0141 & 0.0090 & 95.96 & 99.12 & 99.93 \\

\midrule
Ours 
& \mssecond{0.0092}{0.0002}
& \mssecond{0.0069}{0.0002}
& \mssecond{0.0045}{0.0001}
& \msbest{98.22}{0.09}
& \mssecond{99.61}{0.03}
& \msbest{99.98}{0.002} \\
\bottomrule
\end{tabular}
}
\label{tab:dreds_results}
\end{table}

\begin{table}[t]
\centering
\caption{Quantitative results on STD~\cite{dai2022domain}. Metrics are computed over \textbf{object} pixels.}
\resizebox{\linewidth}{!}{





\begin{tabular}{L C C C C C C}
\toprule
Methods
& \makecell{RMSE$\downarrow$} & \makecell{REL$\downarrow$} & \makecell{MAE$\downarrow$}
& \makecell{$\delta_{1.05}$$\uparrow$} & \makecell{$\delta_{1.10}$$\uparrow$} & \makecell{$\delta_{1.25}$$\uparrow$} \\
\midrule
\multicolumn{7}{c}{\textsc{STD-CatKnown}} \\
\midrule
SwinDR~\cite{dai2022domain} & \textbf{0.015} & \underline{0.013} & \textbf{0.008} & \underline{96.66} & \textbf{99.03} & \textbf{99.92} \\
$\text{D}^3$RoMa~\cite{d3roma} & 0.0176 & 0.0133 & 0.0096 & 96.04 & 98.42 & 99.74 \\

\rowstyle{\color{black}}
PromptDA~\cite{promptda} & 0.3427 & 0.5497 & 0.3353 & 71.63 & 77.29 & 82.26 \\
\midrule
Ours 
& \mssecond{0.0155}{0.0003}
& \msbest{0.0125}{0.0002}
& \mssecond{0.0083}{0.0001}
& \msbest{96.92}{0.12}
& \mssecond{98.87}{0.06}
& \mssecond{99.89}{0.01} \\
\midrule
\multicolumn{7}{c}{\textsc{STD-CatNovel}} \\
\midrule
SwinDR~\cite{dai2022domain} & \underline{0.025} & \underline{0.033} & \underline{0.017} & \underline{81.55} & \underline{93.10} & \textbf{99.84} \\
$\text{D}^3$RoMa~\cite{d3roma} & 0.0283 & 0.0344 & 0.0189 & 79.60 & 91.43 & 99.58 \\

\rowstyle{\color{black}}
PromptDA~\cite{promptda} & \rui{0.1923} & 0.3468 & 0.1819 & 47.45 & 64.66 & 85.34 \\
\midrule
Ours 
& \msbest{0.0221}{0.0031}
& \msbest{0.0273}{0.0043}
& \msbest{0.0145}{0.0022}
& \msbest{86.93}{3.63}
& \msbest{94.97}{2.46}
& \mssecond{99.81}{0.02} \\
\bottomrule
\end{tabular}
}
\label{tab:std_results}
\end{table}

\subsection{In-domain Evaluation}
\label{sec:in_domain_eval}
We evaluate our model in an in-domain setting on three benchmarks: \textsc{HAMMER}~\cite{jung2023importance}, \textsc{DREDS}~\cite{dai2022domain}, and \textsc{ClearPose}~\cite{chen2022clearpose}.
For \methodname, we report mean$\pm$standard deviation over \textbf{three} independent training seeds, and determine \textbf{best}/\underline{second-best} rankings based on the mean values. 
Baseline results are reported as single-run results following the corresponding papers or reproduced implementations.
More implementation details for the baselines and additional results are provided in the Appendix~\ref{app:in_domain_results}.

\noindent\textbf{HAMMER.}
We evaluate on the D435 samples of HAMMER.
Following~\cite{ikemura2024robust}, we use an input and evaluation resolution of \(272\times208\), and compute metrics over all valid pixels for evaluating all methods.  
As shown in Table~\ref{tab:hammer_results}, our method achieves lower mean REL and higher mean $\delta_{1.05}$ than six representative methods~\cite{dai2022domain,d3roma,depthprompting,zuo2024ogni,promptda,ikemura2024robust}. 
Compared with PPFT~\cite{ikemura2024robust}, our model \rui{achieves comparable RMSE ($0.0334\pm0.0051$ versus $0.0351$), while reducing REL and MAE by 21.7\% and 13.6\%, respectively, and improving $\delta_{1.05}$ by 4.6 percentage points (pp).}

\begin{table}[t]
\centering
\caption{Quantitative results on ClearPose~\cite{chen2022clearpose}. Metrics are computed over \textbf{all valid} pixels.}
\resizebox{\linewidth}{!}{

\begin{tabular}{L M C C C C C C}
\toprule
\multicolumn{1}{c}{Testset} & Method 
& RMSE$\downarrow$ & REL$\downarrow$ & MAE$\downarrow$ 
& $\delta_{1.05}\uparrow$ & $\delta_{1.10}\uparrow$ & $\delta_{1.25}\uparrow$ \\
\midrule
\multirow{5}{*}{\makecell[c]{New\\Background}} 
& ImplicitDepth~\cite{zhu2021rgb} & 0.07 & 0.05 & 0.04 & 67.00 & 87.03 & 97.50 \\
& TransCG~\cite{fang2022transcg}       & \textbf{0.03} & 0.03 & 0.02 & 86.50 & 97.02 & \textbf{99.74} \\
& $\text{D}^3\text{RoMa}$~\cite{d3roma} & \underline{0.05} & \underline{0.01} & \textbf{0.01} & \textbf{96.71} & \textbf{98.84} & \textbf{99.74} \\
\rowstyle{\color{black}}
& PromptDA~\cite{promptda} & 0.240 & 0.045 & 0.061 & 85.03 & 93.66 & 97.49 \\
\cmidrule(lr){2-8}
\rowstyle{\color{black}}
& Ours  & \ms{0.057}{0.008} & \msbest{0.009}{0.002} & \mssecond{0.011}{0.004} & \mssecond{96.18}{1.13} & \mssecond{98.55}{0.19} & \mssecond{99.55}{0.09} \\
\midrule
\multirow{5}{*}{\makecell[c]{Heavy\\Occlusion}} 
& ImplicitDepth~\cite{zhu2021rgb} & 0.11 & 0.09 & 0.08 & 41.43 & 66.52 & 91.96 \\
& TransCG~\cite{fang2022transcg}       & \textbf{0.06} & 0.04 & \underline{0.04} & 72.03 & 90.61 & 98.73 \\
& $\text{D}^3\text{RoMa}$~\cite{d3roma} & 0.10 & \underline{0.03} & \underline{0.04} & \underline{83.97} & \underline{93.69} & \underline{98.79} \\
\rowstyle{\color{black}}
& PromptDA~\cite{promptda} & 0.268 & 0.058 & 0.087 & 70.76 & 86.19 & 96.09 \\
\cmidrule(lr){2-8}
\rowstyle{\color{black}}
& Ours  & \mssecond{0.085}{0.011} & \msbest{0.020}{0.003} & \msbest{0.021}{0.006} & \msbest{89.39}{1.87} & \msbest{96.10}{0.40} & \msbest{99.18}{0.20} \\
\midrule
\multirow{5}{*}{\makecell[c]{Translucent\\Cover}} 
& ImplicitDepth~\cite{zhu2021rgb} & \underline{0.16} & 0.16 & 0.13 & 22.85 & 41.17 & 73.11 \\
& TransCG~\cite{fang2022transcg}       & \underline{0.16} & 0.15 & 0.14 & 23.44 & 39.75 & 67.56 \\
& $\text{D}^3\text{RoMa}$~\cite{d3roma} & \textbf{0.13} & \textbf{0.06} & \underline{0.07} & \underline{63.07} & \textbf{82.78} & \textbf{95.80} \\
\rowstyle{\color{black}}
& PromptDA~\cite{promptda} & 0.289 & 0.107 & 0.117 & 56.04 & 71.49 & \underline{89.76} \\
\cmidrule(lr){2-8}
\rowstyle{\color{black}}
& Ours  & \ms{0.170}{0.010} & \mssecond{0.066}{0.002} & \msbest{0.066}{0.004} & \msbest{67.34}{0.85} & \mssecond{73.56}{0.85} & \ms{87.99}{0.52} \\
\midrule
\multirow{5}{*}{\makecell[c]{Opaque\\Distractor}} 
& ImplicitDepth~\cite{zhu2021rgb} & 0.14 & 0.13 & 0.10 & 34.41 & 55.59 & 83.23 \\
& TransCG~\cite{fang2022transcg}       & \textbf{0.08} & 0.06 & 0.06 & 52.43 & 75.52 & 97.53 \\
& $\text{D}^3\text{RoMa}$~\cite{d3roma} & 0.11 & \underline{0.03} & \underline{0.05} & \underline{82.46} & \underline{91.46} & \underline{97.97} \\
\rowstyle{\color{black}}
& PromptDA~\cite{promptda} & 0.307 & 0.088 & 0.104 & 70.65 & 83.16 & 92.94 \\
\cmidrule(lr){2-8}
\rowstyle{\color{black}}
& Ours  & \mssecond{0.092}{0.006} & \msbest{0.026}{0.002} & \msbest{0.024}{0.003} & \msbest{85.25}{0.97} & \msbest{92.07}{0.40} & \msbest{98.47}{0.19} \\
\midrule
\multirow{5}{*}{\makecell[c]{Filled\\Liquid}} 
& ImplicitDepth~\cite{zhu2021rgb} & 0.14 & 0.13 & 0.11 & 32.84 & 53.44 & 84.84 \\
& TransCG~\cite{fang2022transcg}       & \textbf{0.04} & 0.04 & \underline{0.03} & 77.65 & 93.81 & \textbf{99.50} \\
& $\text{D}^3\text{RoMa}$~\cite{d3roma} & 0.09 & \underline{0.03} & 0.04 & \underline{87.58} & \underline{94.85} & 99.15 \\
\rowstyle{\color{black}}
& PromptDA~\cite{promptda} & 0.312 & 0.078 & 0.106 & 73.02 & 84.73 & 93.45 \\
\cmidrule(lr){2-8}
\rowstyle{\color{black}}
& Ours  & \mssecond{0.075}{0.006} & \msbest{0.014}{0.002} & \msbest{0.015}{0.004} & \msbest{93.67}{1.18} & \msbest{97.35}{0.25} & \mssecond{99.22}{0.10} \\
\midrule
\multirow{5}{*}{\makecell[c]{Non\\Planar}} 
& ImplicitDepth~\cite{zhu2021rgb} & 0.18 & 0.16 & 0.15 & 20.34 & 38.57 & 74.02 \\
& TransCG~\cite{fang2022transcg}       & 0.09 & 0.07 & 0.07 & 55.31 & 76.47 & 94.88 \\
& $\text{D}^3\text{RoMa}$~\cite{d3roma} & \underline{0.08} & \underline{0.03} & \underline{0.04} & \underline{84.67} & \underline{92.85} & \underline{98.21} \\
\rowstyle{\color{black}}
& PromptDA~\cite{promptda} & 0.302 & 0.103 & 0.140 & 55.90 & 72.53 & 88.32 \\
\cmidrule(lr){2-8}
\rowstyle{\color{black}}
& Ours  & \msbest{0.059}{0.006} & \msbest{0.016}{0.001} & \msbest{0.016}{0.002} & \msbest{91.88}{0.53} & \msbest{96.14}{0.18} & \msbest{98.95}{0.10} \\
\bottomrule
\end{tabular}
}
\label{tab:clearpose_results}
\end{table}

\begin{table*}[t]
\centering
\caption{Zero-shot generalization results on \textbf{object} pixels. 
We report REL and $\delta_{1.05}$ on eight datasets. 
Bold and underlined entries indicate the best and second-best results, respectively.}
\setlength{\tabcolsep}{4pt}
\resizebox{\textwidth}{!}{

\begin{tabular}{L*{16}{C}C}
\toprule
\multirow{2}{*}{Method}
& \multicolumn{2}{c}{HAMMER} & \multicolumn{2}{c}{HouseCat6D} & \multicolumn{2}{c}{TransCG} & \multicolumn{2}{c}{\rui{GN-Trans}}
& \multicolumn{2}{c}{YCB-V} & \multicolumn{2}{c}{T-LESS} & \multicolumn{2}{c}{XYZ-IBD} & \multicolumn{2}{c}{ROBI}
& \multirow{2}{*}{Mean Rank} \\
\cmidrule(lr){2-3} \cmidrule(lr){4-5} \cmidrule(lr){6-7} \cmidrule(lr){8-9}
\cmidrule(lr){10-11} \cmidrule(lr){12-13} \cmidrule(lr){14-15} \cmidrule(lr){16-17}
& REL$\downarrow$ & $\delta_{1.05}\uparrow$
& REL$\downarrow$ & $\delta_{1.05}\uparrow$
& REL$\downarrow$ & $\delta_{1.05}\uparrow$
& REL$\downarrow$ & $\delta_{1.05}\uparrow$
& REL$\downarrow$ & $\delta_{1.05}\uparrow$
& REL$\downarrow$ & $\delta_{1.05}\uparrow$
& REL$\downarrow$ & $\delta_{1.05}\uparrow$
& REL$\downarrow$ & $\delta_{1.05}\uparrow$ & \\
\midrule
SwinDR~\cite{dai2022domain}
& 0.0429 & 76.35 & 0.0463 & \underline{76.84} & 0.0307 & 81.10 & 0.1078 & \underline{61.75}
& \underline{0.0241} & 88.82 & \underline{0.0151} & \underline{93.53} & 0.0247 & 85.19 & 0.0318 & 77.95
& 3.2 \\
$\text{D}^3\text{RoMa}$~\cite{d3roma}
& 0.0655 & 58.43 & 0.0498 & 76.45 & 0.0419 & 67.56 & 0.1231 & 50.27
& 0.0494 & 81.61 & 0.0511 & 73.36 & 0.0650 & 47.27 & 0.0575 & 50.07
& 5.3 \\
PromptDA~\cite{promptda}
& 0.5156 & 6.62 & 0.3444 & 24.75 & 0.2130 & 8.97 & 0.3535 & 17.28
& 0.4402 & 15.52 & 0.1351 & 62.97 & 0.3060 & 38.45 & 0.6614 & 10.03
& 7.0 \\
OMNI-DC~\cite{omni-dc}
& 0.0586 & 67.47 & 0.0530 & 71.34 & 0.0360 & 79.06 & 0.1642 & 52.00
& \textbf{0.0215} & \textbf{94.59} & 0.0260 & 91.52 & \textbf{0.0047} & \textbf{99.03} & \textbf{0.0063} & \textbf{99.40}
& 3.1 \\
PriorDA~\cite{priorda}
& 0.0762 & 41.42 & 0.0720 & 45.51 & 0.0418 & 71.37 & 0.1492 & 46.55
& 0.0338 & 87.47 & 0.0420 & 73.41 & 0.0205 & 95.26 & 0.0182 & 95.15
& 4.9 \\
CDM~\cite{cdm}
& \underline{0.0427} & \underline{77.95} & \underline{0.0440} & 73.10 & \underline{0.0278} & \underline{84.28} & \underline{0.0919} & 57.94
& 0.0575 & 67.17 & 0.0250 & 88.92 & 0.0145 & 96.60 & 0.0146 & 98.04
& 3.1 \\
\midrule
\rowstyle{\color{black}}
Ours
& \msbest{0.0325}{0.0002} & \msbest{82.99}{0.71}
& \msbest{0.0347}{0.0014} & \msbest{86.50}{0.39}
& \msbest{0.0198}{0.0004} & \msbest{89.47}{0.22}
& \msbest{0.0687}{0.0012} & \msbest{67.84}{1.04}
& \ms{0.0288}{0.0047} & \mssecond{93.73}{0.76}
& \msbest{0.0143}{0.0007} & \msbest{95.73}{0.10}
& \mssecond{0.0070}{0.0009} & \mssecond{98.76}{0.46}
& \mssecond{0.0098}{0.0005} & \mssecond{99.05}{0.28}
& \textbf{1.4} \\
\bottomrule
\end{tabular}

}
\label{tab:zs_obj_results}
\end{table*}

\noindent\textbf{DREDS and STD.}
Following~\cite{dai2022domain}, we evaluate all methods at \(224\times126\) over foreground object regions.  
As shown in Table~\ref{tab:dreds_results} and Table~\ref{tab:std_results}, our model achieves consistently strong results, ranking first or second across nearly all metrics.  
On DREDS-CatKnown, our method performs comparably to D$^3$RoMa~\cite{d3roma}, reducing RMSE and REL by 8.6\% and 8.1\%, respectively, and slightly improving $\delta_{1.05}$ by 0.1~pp.  
On STD-CatKnown, our model achieves competitive performance across all metrics. On STD-CatNovel, it \rui{achieves comparable RMSE to SwinDR~\cite{dai2022domain} ($0.0221\pm0.0031$ versus $0.0250$), while reducing REL by 17.2\% and improving $\delta_{1.05}$ by 5.4~pp.}  
These results demonstrate that our method remains robust under domain shifts and provides clear advantages in generalizing to unseen object categories.

\noindent\textbf{ClearPose.}
Following~\cite{d3roma}, both input and evaluation resolutions are set to \(320\times240\), and metrics are computed over all valid pixels.
As shown in Table~\ref{tab:clearpose_results}, our method achieves the best or second-best performance on most metrics across ClearPose subsets, with particularly strong gains under \emph{Heavy Occlusion} and \emph{Non-Planar} settings.
In these challenging cases, RMSE drops from 0.10 to 0.085 and from 0.08 to 0.059, respectively.
On the \emph{Translucent Cover} subset, our model obtains comparable results due to a transparent bin partially covering the objects, which interferes with the relative depth prediction of DAV2~\cite{depthanythingv2}.  
Overall, our method demonstrates robust depth estimation across diverse materials and occlusion conditions, achieving competitive performance on this real-world benchmark.

\subsection{Zero-shot Generalization}
\label{sec:zs_results}
Table~\ref{tab:zs_obj_results} reports a quantitative comparison on our zero-shot benchmark. 
Because object-region geometry is most relevant to object-centric
manipulation,
we report REL and $\delta_{1.05}$ computed over object pixels. 
For Table~\ref{tab:zs_obj_results}, all methods use $518{\times}518$ RGB--D inputs, with predictions evaluated at $320{\times}180$ on object pixels.
Results on all valid pixels and different input resolutions are provided in Appendix~\ref{app:zero_shot_results}.

\begin{figure*}[t]
\centering
\includegraphics[width=\textwidth]{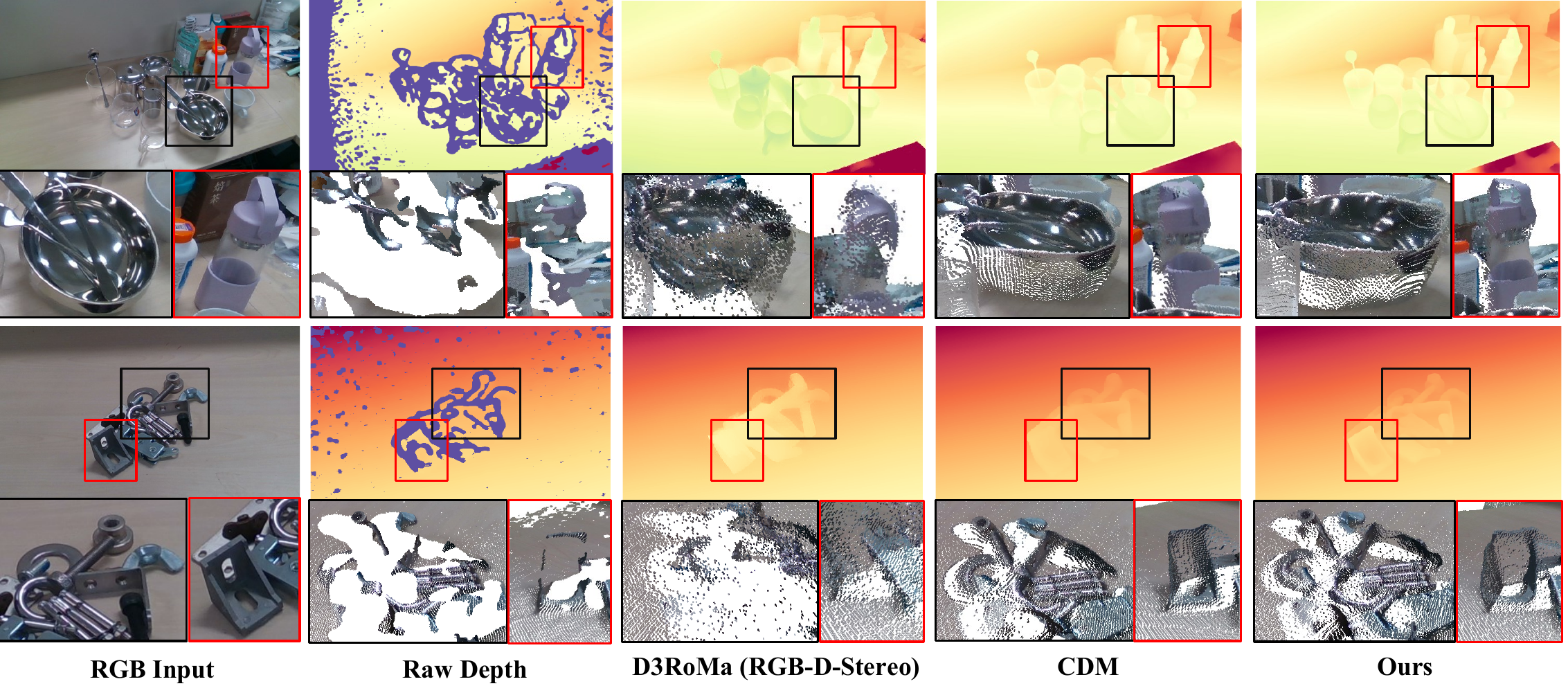}
\caption{
Qualitative zero-shot comparison on in-the-wild RGB-D scenes captured with a RealSense D435.
We show the raw depth and the back-projected point clouds using the same intrinsics, with insets highlighting challenging regions.
Our method reduces holes and structured noise on transparent/reflective objects and better preserves fine-scale geometry and boundaries compared with baselines.
}
\label{fig:in_the_wild}
\end{figure*}

\noindent\textbf{Baselines.}
We compare against six generalized methods. SwinDR~\cite{dai2022domain}, D$^3$RoMa~\cite{d3roma}, and CDM~\cite{cdm} are depth restoration approaches tailored to robotic RGB-D sensing, with CDM additionally learning camera-specific noise models to synthesize large-scale training data. PromptDA~\cite{promptda} and PriorDA~\cite{priorda} perform prompt-based depth completion/super-resolution (e.g., using low-resolution LiDAR or generic geometric prompts), while OMNI-DC~\cite{omni-dc} addresses sparse-LiDAR depth completion. Although PromptDA, PriorDA, and OMNI-DC are not explicitly designed for manipulation benchmarks, they exhibit strong zero-shot ability on completion/super-resolution tasks and, together with CDM, are all built on DAV2~\cite{depthanythingv2}, making them suitable baselines for our study. More implementation details are provided in the Appendix~\ref{app:zero_shot_results}.

\noindent\textbf{Results.}
As shown in Table~\ref{tab:zs_obj_results}, our method generalizes best across the eight zero-shot benchmarks, achieving a substantially lower mean rank than all other methods (1.4 vs.\ $\ge$3.1).
It ranks first on HAMMER, HouseCat6D, TransCG, GraspNet-Trans, and T-LESS, improving both REL and $\delta_{1.05}$ (e.g., reducing REL by 21.1\% on HouseCat6D and 28.8\% on TransCG compared with CDM).
On industrial-style benchmarks, it attains the best performance on T-LESS and reaches near-saturated accuracy on XYZ-IBD and ROBI, with $\delta_{1.05}$ of 98.76 and 99.05, respectively.
\rui{The percentage reductions above are calculated from Table~\ref{tab:zs_obj_results}. 
Results using each method's official or preferred input resolution are reported in Appendix Table~\ref{tab:zs_obj_results_official_resolution}.}



Among baselines, SwinDR~\cite{dai2022domain} is consistently strong but still trails ours on harder transparent and cluttered benchmarks such as HouseCat6D and TransCG.
OMNI-DC~\cite{omni-dc} is the strongest baseline by mean rank and performs best on the industrial datasets (e.g., XYZ-IBD and ROBI), likely benefiting from its scale normalization; however, it degrades on noisier benchmarks.
By contrast, our robust scale normalization is designed to reduce outlier bias under commodity RGB-D noise, improving stability under domain shifts.
A more detailed analysis of OMNI-DC is provided in the Appendix~\ref{app:zero_shot_results}.

\begin{table*}[t]
	\centering\footnotesize
	\caption{Ablation study on our framework components on the test set of HAMMER~\cite{jung2023importance} and DREDS-CatKnown~\cite{dai2022domain}. Metrics are computed over \textbf{Object} pixels.}
    \resizebox{\textwidth}{!}{


\begin{tabular}{l*{11}{c}}
\toprule
\multirow{2}{*}{Model Variant} & \multirow{2}{*}{ADF} & \multirow{2}{*}{Unc. Modeling} & \multirow{2}{*}{UCFC} & \multicolumn{4}{c}{HAMMER} & \multicolumn{4}{c}{DREDS-CatKnown} \\
\cmidrule(lr){5-8} \cmidrule(lr){9-12}
 & & & & $\delta_{1.05}\uparrow$ & MAE$\downarrow$ & RMSE$\downarrow$ & REL$\downarrow$ & $\delta_{1.05}\uparrow$ & MAE$\downarrow$ & RMSE$\downarrow$ & REL$\downarrow$ \\
\midrule
Baseline & \xmark & \xmark & \xmark & \rui{78.20} & \rui{0.0217} & \rui{0.0359} & \rui{0.0365} & 98.15 & 0.0046 & 0.0094 & 0.0070 \\
+ ADF & \cmark & \xmark & \xmark & \rui{82.55} & \rui{0.0190} & \rui{0.0325} & \rui{0.0320} & 98.17 & 0.0045 & 0.0095 & 0.0069 \\
+ Unc. Modeling & \cmark & \cmark & \xmark & \rui{80.16} & \rui{0.0206} & \rui{0.0349} & \rui{0.0340} & 98.19 & 0.0045 & 0.0094 & 0.0069 \\
+ UCFC (Ours) & \cmark & \cmark & \cmark & \rui{\textbf{85.75}} & \rui{\textbf{0.0173}} & \rui{\textbf{0.0312}} & \rui{\textbf{0.0285}} & \textbf{98.32} & \textbf{0.0044} & \textbf{\rui{0.0090}} & \textbf{0.0067} \\
\bottomrule
\end{tabular}}
\label{tab:ablation_component}
\end{table*}

\begin{table*}[t]
\centering
\caption{Effect of different error proxy $\mathcal{E}$ on depth estimation and uncertainty estimation performance on DREDS-CatKnown~\cite{dai2022domain}. 
Metrics are computed over \textbf{Object} pixels.
}
\begin{tabular}{lccccccccc}
\toprule
\multirow{2}{*}{Model Variant} 
& \multirow{2}{*}{REL$\downarrow$} & \multirow{2}{*}{RMSE$\downarrow$}  & \multirow{2}{*}{$\delta_{1.05}\uparrow$} 
& \multicolumn{2}{c}{REL} & \multicolumn{2}{c}{RMSE} & \multicolumn{2}{c}{$\delta_{1.05}$} \\
\cmidrule(lr){5-6} \cmidrule(lr){7-8} \cmidrule(lr){9-10}
& & & & AUSE$\downarrow$ & AURG$\uparrow$
& AUSE$\downarrow$ & AURG$\uparrow$
& AUSE$\downarrow$ & AURG$\uparrow$ \\
\midrule
Baseline + ADF
& 0.00685 & 0.00945 & 98.17
& -- & -- & -- & -- & -- & -- \\
+ $L_1$
& 0.00794 & 0.01038 & 97.70
& 0.00021 & \textbf{0.00055} & 0.00036 & \textbf{0.00075} & 0.00040 & \textbf{0.00259} \\
+ $L_2$
& \textbf{0.00642} & 0.00924 & 98.26
& 0.00059 & 0.00003 & 0.00090 & 0.00011 & 0.00200 & 0.00020 \\
+ Grad
& 0.00653 & 0.00921 & 98.26
& 0.00021 & 0.00041 & 0.00028 & 0.00073 & 0.00043 & 0.00181 \\
+ $L_1$ + Grad (Ours)
& 0.00672 & \textbf{0.00898} & \textbf{98.32}
& \textbf{0.00019} & 0.00046 
& \textbf{0.00024} & 0.00073
& \textbf{0.00032} & 0.00188 \\
\bottomrule
\end{tabular}
\label{tab:unc_modelling}
\end{table*}

CDM~\cite{cdm} leverages camera-specific noise modeling and can be competitive when a matching sensor model is available (e.g., D435), but it transfers less reliably across sensors.
D$^3$RoMa~\cite{d3roma} (RGB-D) improves over raw depth on several benchmarks, yet it is less consistent than our method on harder cross-domain settings.
PromptDA~\cite{promptda} and PriorDA~\cite{priorda} target completion/super-resolution from cleaner or structured inputs and are less robust to the noisy, incomplete depth typical of commodity RGB-D.
More detailed comparisons are provided in the Appendix~\ref{app:zero_shot_results}.

Fig.~\ref{fig:in_the_wild} shows qualitative zero-shot restoration on in-the-wild scenes with reflective and mixed-material objects.
Raw depth exhibits large missing regions and structured noise on challenging surfaces.
D$^3$RoMa and CDM partially suppress these artifacts, but can still leave oversmoothed surfaces or residual speckle and holes in the highlighted regions.
Our method produces more coherent surfaces with fewer artifacts while retaining finer structures and sharper boundaries.

\subsection{Ablation Analysis}
\label{sec:ablations}
\noindent\textbf{Effect of Each Design in \methodname.}
We conduct ablations on HAMMER and DREDS-CatKnown to analyze the contribution of each module.
\rui{For HAMMER, we follow the same in-domain object-pixel protocol as Appendix Table~\ref{tab:ucfc_indomain_hammer_app}; the baseline uses its epoch-20 checkpoint.}
The baseline model initializes depth directly from the raw sensor map and uses a DAV2 encoder with context networks for ConvGRU refinement, without ADF, uncertainty modeling, or UCFC.
As shown in Table~\ref{tab:ablation_component}, adding ADF provides a cleaner initialization by replacing unreliable sensor regions with the aligned DAV2 prior.
\rui{On HAMMER, ADF reduces MAE from 0.0217 to 0.0190 and REL from 0.0365 to 0.0320.}
Introducing uncertainty modeling alone results in little or slightly negative impact, indicating that injecting an uncertainty branch does not improve depth prediction unless it participates in refinement through UCFC.
\rui{Once UCFC is enabled, uncertainty gates context features, improving all four HAMMER depth metrics relative to Uncertainty-only.}
\rui{Compared with the baseline on HAMMER, UCFC reduces MAE, RMSE, and REL by 20.4\%, 13.1\%, and 21.9\%, respectively, and increases $\delta_{1.05}$ by 7.55~pp.}
\rui{On DREDS-CatKnown, UCFC reduces RMSE from 0.0094 to 0.0090 and REL from 0.0070 to 0.0067 relative to the baseline.}
These component ablations show that ADF improves initialization and that uncertainty prediction alone is insufficient.

\noindent\textbf{Different Modeling of Error Proxy $\mathcal{E}$.}
We further investigate how different formulations of the error proxy $\mathcal{E}$ affect depth accuracy and uncertainty quality on the DREDS-CatKnown (Table~\ref{tab:unc_modelling}). 
Using only the $L_1$ loss produces smoother but less precise depth estimates, whereas the gradient-based loss improves local structural consistency and yields competitive depth accuracy. 
Interestingly, the $L_2$ loss achieves the lowest REL, but its uncertainty estimates are poor uncertainty ranking, reflected by significantly higher AUSE and lower AURG. 
When $L_1$ and Grad are combined, the model achieves the best balance across metrics, producing the lowest overall uncertainty error and strong depth performance.
\rui{Across all three error metrics, the $L_1{+}$Grad variant has the lowest AUSE, indicating the smallest area between its sparsification curve and the oracle curve. The $L_1$-only variant has the highest AURG on all three metrics, indicating the largest gain over random pixel removal.}
We adopt the $L_1{+}$Grad formulation in our final model, as it delivers the most consistent and reliable performance across both depth estimation and uncertainty evaluation.

\noindent\textbf{Ablation of UCFC Feedback.}
We further examine whether the performance gain comes from recurrent uncertainty feedback, rather than from extra parameters or generic feature fusion.
All variants keep the same DAV2 backbone, ADF initialization, Robust-SN, recurrent refinement backbone, training data, input resolution, and evaluation protocol; they only modify how reliability information is injected into refinement.
We compare five variants: \emph{Plain refinement}, which removes uncertainty supervision and feedback; \emph{Residual gate}, which replaces the predicted uncertainty map with the normalized absolute residual between the current depth estimate $\mathbf{D}^{k}$ and the original sensor depth $\mathbf{D}_{\mathrm{sen}}$; \emph{UCFC-PM}, which uses the same calibration modules but replaces the uncertainty map with a constant map as a parameter-matched no-feedback variant; \emph{Simple conv fusion}, which directly concatenates depth, uncertainty, RGB features, and sensor-depth features through convolutional fusion; and \emph{UCFC}, which feeds the evolving predicted uncertainty back to recalibrate context features at each recurrent step.
\rui{We additionally include \emph{Frozen-U}, which reuses the uncertainty state predicted after the first refinement step for subsequent feedback, to distinguish the benefit of an informative spatial uncertainty map from that of refreshing it at every iteration.}
\rui{We used a different training seed for the UCFC ablation in Table~\ref{tab:zs_ucfc_ablation} than for the scale-normalization ablation in Table~\ref{tab:sn_ablation}.}

\begin{table*}[t]
\centering
\small
\caption{Ablation of UCFC feedback. Metrics are computed over \textbf{Object} pixels.}
\setlength{\tabcolsep}{4pt}
\resizebox{\textwidth}{!}{
\begin{tabular}{l*{16}{c}c}
\toprule
\multirow{2}{*}{Model Variant} 
& \multicolumn{2}{c}{HAMMER} 
& \multicolumn{2}{c}{HouseCat6D} 
& \multicolumn{2}{c}{TransCG} 
& \multicolumn{2}{c}{\rui{GN-Trans}} 
& \multicolumn{2}{c}{YCB-V} 
& \multicolumn{2}{c}{T-LESS} 
& \multicolumn{2}{c}{XYZ-IBD} 
& \multicolumn{2}{c}{ROBI} 
& \multirow{2}{*}{Mean Rank} \\
\cmidrule(lr){2-3} 
\cmidrule(lr){4-5} 
\cmidrule(lr){6-7} 
\cmidrule(lr){8-9} 
\cmidrule(lr){10-11} 
\cmidrule(lr){12-13} 
\cmidrule(lr){14-15} 
\cmidrule(lr){16-17}
& REL$\downarrow$ & $\delta_{1.05}\uparrow$
& REL$\downarrow$ & $\delta_{1.05}\uparrow$
& REL$\downarrow$ & $\delta_{1.05}\uparrow$
& REL$\downarrow$ & $\delta_{1.05}\uparrow$
& REL$\downarrow$ & $\delta_{1.05}\uparrow$
& REL$\downarrow$ & $\delta_{1.05}\uparrow$
& REL$\downarrow$ & $\delta_{1.05}\uparrow$
& REL$\downarrow$ & $\delta_{1.05}\uparrow$
& \\
\midrule
Plain refinement 
& 0.0365 & 78.05
& 0.0346 & 85.14
& 0.0221 & 86.98
& 0.0566 & 72.23
& 0.0295 & 91.30
& 0.0149 & 95.33
& 0.0090 & 97.87
& 0.0124 & 98.17
& \rui{4.84} \\

Residual gate 
& 0.0393 & 75.92
& 0.0391 & 81.85
& 0.0228 & 86.24
& 0.0682 & 66.67
& 0.0267 & 93.31
& 0.0148 & 94.67
& 0.0065 & 99.38
& 0.0116 & 98.48
& \rui{4.69} \\

Simple conv fusion 
& 0.0351 & 82.36
& 0.0330 & 85.80
& 0.0214 & 87.75
& 0.0727 & 64.39
& \textbf{0.0223} & \textbf{94.11}
& 0.0144 & 95.07
& \textbf{0.0057} & \textbf{99.59}
& 0.0114 & 98.47
& \rui{2.97} \\

UCFC-PM
& 0.0345 & 79.46
& \textbf{0.0322} & 86.83
& 0.0220 & 87.43
& \textbf{0.0545} & \textbf{72.47}
& 0.0404 & 79.91
& 0.0146 & 95.40
& 0.0081 & 99.01
& 0.0121 & 98.54
& \rui{3.28} \\

\rui{Frozen-U}
& \rui{0.0351} & \rui{79.17}
& \rui{0.0346} & \rui{85.54}
& \rui{0.0213} & \rui{87.99}
& \rui{0.0603} & \rui{70.93}
& \rui{0.0256} & \rui{93.66}
& \rui{0.0141} & \rui{95.45}
& \rui{0.0079} & \rui{98.66}
& \rui{0.0117} & \rui{98.54}
& \rui{3.09} \\

UCFC (Ours) 
& \textbf{0.0326} & \textbf{82.38}
& 0.0334 & \textbf{86.86}
& \textbf{0.0199} & \textbf{89.52}
& 0.0701 & 67.28
& 0.0262 & 93.94
& \textbf{0.0137} & \textbf{95.73}
& 0.0069 & 98.35
& \textbf{0.0103} & \textbf{98.74}
& \rui{\textbf{2.13}} \\
\bottomrule

\end{tabular}
}
\label{tab:zs_ucfc_ablation}
\vspace{-2mm}
\end{table*}

\begin{table*}[th]
\centering
\small
\caption{Effect of different scale normalization models. Metrics are computed over \textbf{Object} pixels.}
\setlength{\tabcolsep}{4pt}
\resizebox{\textwidth}{!}{

\begin{tabular}{lcccccccccccccccccc}
\toprule
\multirow{2}{*}{Model Variant} 
& \multicolumn{2}{c}{HAMMER} 
& \multicolumn{2}{c}{HouseCat6D} 
& \multicolumn{2}{c}{TransCG} 
& \multicolumn{2}{c}{\rui{GN-Trans}} 
& \multicolumn{2}{c}{YCB-V} 
& \multicolumn{2}{c}{T-LESS} 
& \multicolumn{2}{c}{XYZ-IBD} 
& \multicolumn{2}{c}{ROBI} 
& \multirow{2}{*}{Mean Rank} 
\\
\cmidrule(lr){2-3} \cmidrule(lr){4-5} \cmidrule(lr){6-7} 
\cmidrule(lr){8-9} \cmidrule(lr){10-11} \cmidrule(lr){12-13} \cmidrule(lr){14-15} \cmidrule(lr){16-17}
& REL$\downarrow$ & $\delta_{1.05}\uparrow$
& REL$\downarrow$ & $\delta_{1.05}\uparrow$
& REL$\downarrow$ & $\delta_{1.05}\uparrow$
& REL$\downarrow$ & $\delta_{1.05}\uparrow$
& REL$\downarrow$ & $\delta_{1.05}\uparrow$
& REL$\downarrow$ & $\delta_{1.05}\uparrow$
& REL$\downarrow$ & $\delta_{1.05}\uparrow$
& REL$\downarrow$ & $\delta_{1.05}\uparrow$
&  \\
\midrule
Vanilla (w/o SN)      & 0.0349 & 80.84 & 0.0424 & 83.05 & 0.0221 & 87.13 & 0.0671 & 66.67 & \textbf{0.0252} & 93.60 & \textbf{0.0120} & \textbf{96.32} & 0.0078 & 99.20 & 0.0199 & 90.36 & 3.25 \\
SN-Median             & 0.0327 & \textbf{83.19} & 0.0351 & \textbf{86.84} & 0.0198 & 89.46 & 0.0657 & 69.16 & 0.0285 & 93.85 & 0.0159 & 95.45 & 0.0077 & 98.90 & 0.0109 & 98.49 & 2.63 \\
SN-LogBias            & 0.0343 & 81.52 & 0.0350 & 85.68 & 0.0204 & 88.95 & \textbf{0.0648} & \textbf{69.21} & 0.0255 & 94.01 & 0.0175 & 95.64 & 0.0073 & 99.06 & 0.0105 & 98.66 & 2.44 \\
SN-LogBias-Robust     & \textbf{0.0324} & 82.96 & \textbf{0.0345} & \rui{86.09} & \textbf{0.0193} & \textbf{89.66} & 0.0677 & 67.20 & 0.0259 & \textbf{94.35} & 0.0143 & 95.83 & \textbf{0.0063} & \textbf{99.25} & \textbf{0.0096} & \textbf{99.12} & \textbf{1.69} \\
\bottomrule
\end{tabular}

}
\label{tab:sn_ablation}
\vspace{-2mm}
\end{table*}

\rui{As shown in Table~\ref{tab:zs_ucfc_ablation}, UCFC achieves the best mean rank (2.13) among the six variants. Compared with Plain refinement, UCFC improves both REL and $\delta_{1.05}$ on seven of the eight zero-shot datasets. It also achieves lower REL and higher $\delta_{1.05}$ than Frozen-U on six datasets for each metric, supporting the overall benefit of uncertainty feedback during recurrent refinement.

GraspNet-Trans is the main exception: UCFC obtains 0.0701 REL and 67.28\% $\delta_{1.05}$, compared with 0.0566/72.23\% for Plain refinement, 0.0545/72.47\% for UCFC-PM, and 0.0603/70.93\% for Frozen-U. The failure analysis in Appendix~\ref{app:failure_taxonomy} shows that biased sensor measurements account for 77.5\% of the overall Mean REL gap to UCFC-PM. Missing regions with inaccurate initialization contribute further, whereas reliable sensor pixels favor UCFC. Thus, the degradation cannot be explained by missing-depth prevalence alone.}
More details, including the in-domain HAMMER analysis, are provided in Appendix~\ref{app:ucfc_ablation}.

\rui{%
\noindent\textbf{Generalization to Different Depth Foundation Models.}
To examine whether the benefits of UCFC extend beyond DAV2~\cite{depthanythingv2}, we evaluate MiDaS v3.1 with a BEiT-L encoder~\cite{midasv31} and Depth Anything 3 Mono (DA3Mono)~\cite{depthanything3}.
Each model provides its own depth prior and RGB features, using MiDaS DPT features or intermediate DA3Mono features. 
We adopt lightweight feature adaptation using single-layer $1\times1$ convolutional projections and spatial resizing where needed.
The pretrained models remain frozen, while the feature projections and restoration modules are trained without changing ADF, Robust-SN, recurrent refinement, or the loss functions.
All configurations follow the same mixed-dataset training protocol and are evaluated on eight unseen datasets.

The variant without UCFC retains ADF, Robust-SN, uncertainty supervision, and recurrent refinement, but does not use uncertainty to calibrate context features. As shown in Table~\ref{tab:cross_foundation_ucfc}, enabling UCFC with MiDaS BEiT-L reduces REL on six of eight datasets and improves $\delta_{1.05}$ on all eight, while DA3Mono-L achieves improvements on five and six datasets, respectively. These results show that UCFC is effective across different pretrained depth priors, although its benefits vary with the backbone and target dataset.
}

\begin{table*}[t]
\centering
\footnotesize
\caption{\rui{Effect of UCFC across different depth foundation models. Metrics are computed over \textbf{Object} pixels.
Parentheses show the change from without to with UCFC (REL / pp), with \textcolor{red!75!black}{red} indicating improvement and \textcolor{green!50!black}{green} degradation.}}
\label{tab:cross_foundation_ucfc}
\begingroup
\setlength{\tabcolsep}{3pt}
\renewcommand{\arraystretch}{1.15}
\resizebox{\textwidth}{!}{%
\rui{
\begin{tabular}{lc*{16}{c}}
\toprule
\multirow{2}{*}{Depth Model} & \multirow{2}{*}{UCFC} & \multicolumn{2}{c}{HAMMER} & \multicolumn{2}{c}{HouseCat6D} & \multicolumn{2}{c}{TransCG} & \multicolumn{2}{c}{\rui{GN-Trans}} & \multicolumn{2}{c}{YCB-V} & \multicolumn{2}{c}{T-LESS} & \multicolumn{2}{c}{XYZ-IBD} & \multicolumn{2}{c}{ROBI} \\
\cmidrule(lr){3-4} \cmidrule(lr){5-6} \cmidrule(lr){7-8} \cmidrule(lr){9-10} \cmidrule(lr){11-12} \cmidrule(lr){13-14} \cmidrule(lr){15-16} \cmidrule(lr){17-18}
 & & REL$\downarrow$ & $\delta_{1.05}\uparrow$ & REL$\downarrow$ & $\delta_{1.05}\uparrow$ & REL$\downarrow$ & $\delta_{1.05}\uparrow$ & REL$\downarrow$ & $\delta_{1.05}\uparrow$ & REL$\downarrow$ & $\delta_{1.05}\uparrow$ & REL$\downarrow$ & $\delta_{1.05}\uparrow$ & REL$\downarrow$ & $\delta_{1.05}\uparrow$ & REL$\downarrow$ & $\delta_{1.05}\uparrow$ \\
\midrule
\multirow{2}{*}{MiDaS BEiT-L} & \xmark  & 0.0380 & 76.37 & 0.0387 & 79.48 & 0.0239 & 85.30 & 0.0709 & 67.08 & 0.0339 & 85.70 & 0.0184 & 90.29 & 0.0102 & 94.75 & 0.0192 & 89.29 \\
& \cmark  & \shortstack[c]{0.0381\\[-0.5ex]{\scriptsize\textcolor{green!50!black}{(+0.00016)}}} & \shortstack[c]{77.78\\[-0.5ex]{\scriptsize\textcolor{red!75!black}{(+1.40)}}} & \shortstack[c]{0.0387\\[-0.5ex]{\scriptsize\textcolor{green!50!black}{(+0.00002)}}} & \shortstack[c]{82.31\\[-0.5ex]{\scriptsize\textcolor{red!75!black}{(+2.83)}}} & \shortstack[c]{0.0216\\[-0.5ex]{\scriptsize\textcolor{red!75!black}{(-0.00232)}}} & \shortstack[c]{87.45\\[-0.5ex]{\scriptsize\textcolor{red!75!black}{(+2.15)}}} & \shortstack[c]{0.0683\\[-0.5ex]{\scriptsize\textcolor{red!75!black}{(-0.00264)}}} & \shortstack[c]{68.29\\[-0.5ex]{\scriptsize\textcolor{red!75!black}{(+1.21)}}} & \shortstack[c]{0.0256\\[-0.5ex]{\scriptsize\textcolor{red!75!black}{(-0.00833)}}} & \shortstack[c]{94.17\\[-0.5ex]{\scriptsize\textcolor{red!75!black}{(+8.47)}}} & \shortstack[c]{0.0138\\[-0.5ex]{\scriptsize\textcolor{red!75!black}{(-0.00459)}}} & \shortstack[c]{95.51\\[-0.5ex]{\scriptsize\textcolor{red!75!black}{(+5.22)}}} & \shortstack[c]{0.0057\\[-0.5ex]{\scriptsize\textcolor{red!75!black}{(-0.00454)}}} & \shortstack[c]{99.31\\[-0.5ex]{\scriptsize\textcolor{red!75!black}{(+4.57)}}} & \shortstack[c]{0.0116\\[-0.5ex]{\scriptsize\textcolor{red!75!black}{(-0.00751)}}} & \shortstack[c]{98.17\\[-0.5ex]{\scriptsize\textcolor{red!75!black}{(+8.87)}}} \\
\midrule
\multirow{2}{*}{DA3Mono-L} & \xmark  & 0.0654 & 76.71 & 0.0341 & 86.51 & 0.0234 & 85.28 & 0.0618 & 68.78 & 0.0351 & 90.98 & 0.0175 & 95.06 & 0.0082 & 98.07 & 0.0148 & 95.43 \\
& \cmark  & \shortstack[c]{0.0530\\[-0.5ex]{\scriptsize\textcolor{red!75!black}{(-0.01241)}}} & \shortstack[c]{80.85\\[-0.5ex]{\scriptsize\textcolor{red!75!black}{(+4.13)}}} & \shortstack[c]{0.0314\\[-0.5ex]{\scriptsize\textcolor{red!75!black}{(-0.00264)}}} & \shortstack[c]{87.54\\[-0.5ex]{\scriptsize\textcolor{red!75!black}{(+1.04)}}} & \shortstack[c]{0.0236\\[-0.5ex]{\scriptsize\textcolor{green!50!black}{(+0.00022)}}} & \shortstack[c]{85.82\\[-0.5ex]{\scriptsize\textcolor{red!75!black}{(+0.53)}}} & \shortstack[c]{0.0552\\[-0.5ex]{\scriptsize\textcolor{red!75!black}{(-0.00659)}}} & \shortstack[c]{72.04\\[-0.5ex]{\scriptsize\textcolor{red!75!black}{(+3.26)}}} & \shortstack[c]{0.0295\\[-0.5ex]{\scriptsize\textcolor{red!75!black}{(-0.00556)}}} & \shortstack[c]{93.14\\[-0.5ex]{\scriptsize\textcolor{red!75!black}{(+2.16)}}} & \shortstack[c]{0.0212\\[-0.5ex]{\scriptsize\textcolor{green!50!black}{(+0.00373)}}} & \shortstack[c]{92.89\\[-0.5ex]{\scriptsize\textcolor{green!50!black}{(-2.17)}}} & \shortstack[c]{0.0104\\[-0.5ex]{\scriptsize\textcolor{green!50!black}{(+0.00214)}}} & \shortstack[c]{96.80\\[-0.5ex]{\scriptsize\textcolor{green!50!black}{(-1.27)}}} & \shortstack[c]{0.0137\\[-0.5ex]{\scriptsize\textcolor{red!75!black}{(-0.00109)}}} & \shortstack[c]{97.08\\[-0.5ex]{\scriptsize\textcolor{red!75!black}{(+1.65)}}} \\
\bottomrule
\end{tabular}
}
}
\endgroup
\end{table*}

\begin{figure}[t]
    \centering
    \includegraphics[width=\linewidth]{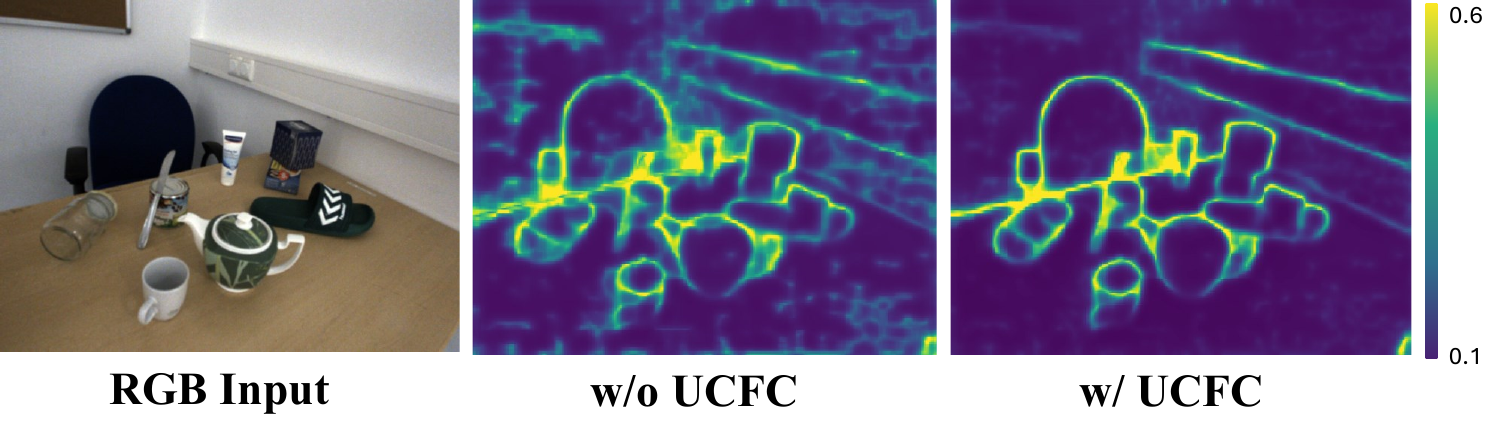}
    \caption{Effect of UCFC on predicted uncertainty.
    We show the RGB input and the uncertainty maps predicted without and with UCFC.
    With UCFC, the uncertainty map is less noisy and exhibits more coherent high-uncertainty responses near transparent surfaces and depth discontinuities.
    All uncertainty maps use identical visualization scaling.}
    \label{fig:ucfc_vis}
\end{figure}

\noindent\textbf{UCFC Stabilizes Refinement and Improves Convergence.}
We further study the effect of UCFC on uncertainty quality and convergence behavior.
Fig.~\ref{fig:ucfc_vis} shows that UCFC reduces speckle-like uncertainty artifacts in the background and yields more spatially coherent uncertainty around transparent regions and depth discontinuities.
This suggests that uncertainty-gated context can suppress misleading cues during iterative refinement and reduce error accumulation.

\begin{figure}[t]
  \centering
\includegraphics[width=\linewidth]{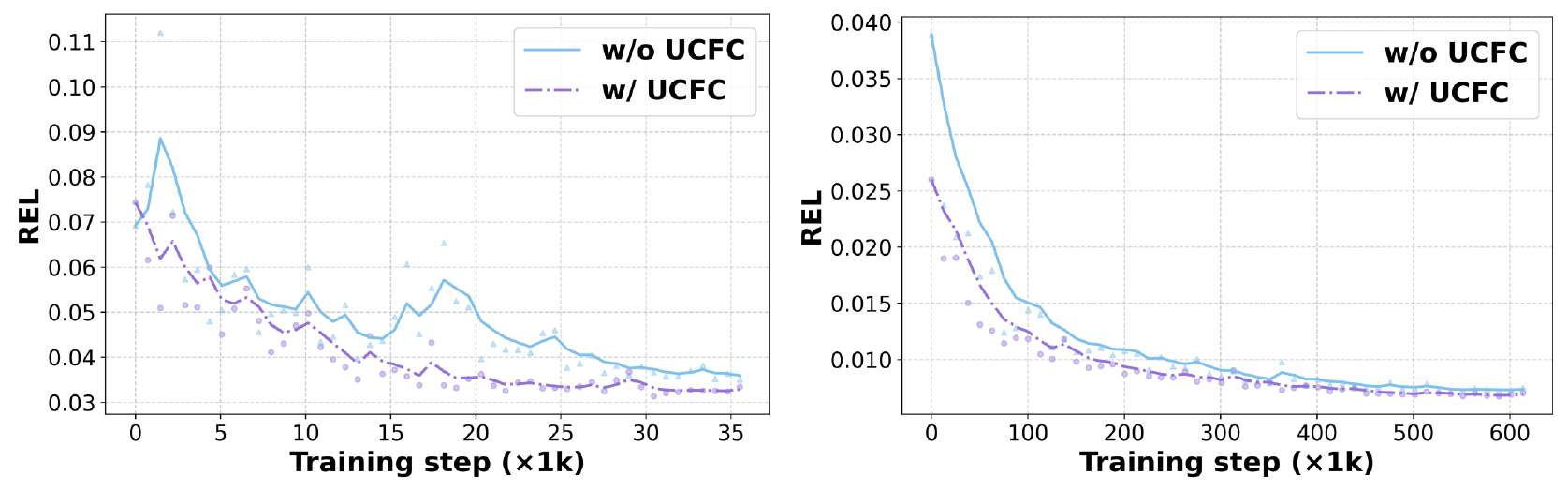}
\caption{Training REL curves on HAMMER (\textbf{left}) and DREDS-CatKnown (\textbf{right}).
UCFC yields lower training error and smoother convergence.
Curves are shown with a moving average for readability.}
\label{fig:UCFC_learning}
\end{figure}

Fig.~\ref{fig:UCFC_learning} shows the training REL curves (smoothed for visualization) on HAMMER and DREDS-CatKnown.
With UCFC, the model converges faster and remains lower in REL throughout training.
At the end of training, REL decreases from 0.034 to 0.030 on HAMMER and from 0.009 to 0.008 on DREDS-CatKnown, suggesting that UCFC improves optimization stability and final accuracy under noisy RGB-D inputs.

\begin{figure}[th]
  \centering
  \includegraphics[width=\linewidth]{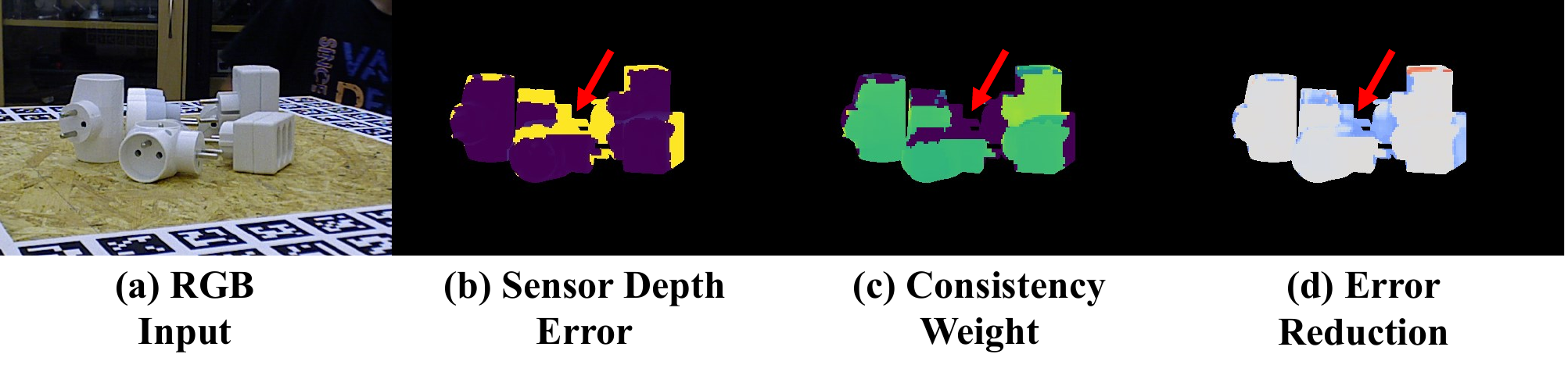}
  \caption{
    Visualization of the two–stage robust scale normalization on T-LESS~\cite{tless}.
    (a) RGB input.
    (b) Absolute error of sensor depth (brighter indicates larger errors).
    (c) Consistency weight map $w(x,y)$, where high values (green) keep reliable sensor pixels and low values (purple) suppress inconsistent measurements. The red arrow highlights a multi-object junction where sensor depth exhibits large errors and is therefore down-weighted by $w(x,y)$.
    (d) Per-pixel absolute error comparison. 
    \textcolor{blue!60}{blue} indicates SN-LogBias-Robust has lower error, \textcolor{red!60}{red} indicates SN-LogBias has lower error.
  }
  \label{fig:sn_vis}
\end{figure}

\noindent\textbf{Effect of Scale Normalization.}
We study the role of scale normalization (SN) using four variants in Table~\ref{tab:sn_ablation}.
(i) Vanilla performs restoration directly in metric depth without SN.
(ii) SN-Median follows Eq.~\eqref{eq:raw_sn} and normalizes log-depth using the sensor median, while directly taking the initial log-depth from the ADF output.
(iii) SN-LogBias further aligns the DAV2 relative prior to the sensor by estimating a per-frame log-bias and an anchor depth using simple medians.
(iv) SN-LogBias-Robust replaces these median estimates with our two-stage consistency-weighted initialization.

Overall, introducing SN is beneficial for mixed-dataset training. Even the simplest SN-Median improves the mean rank from 3.25 to 2.63, but the improvement is not consistent across datasets, indicating that a single anchor statistic can be sensitive to material-dependent corruption and missing depth.
Adding explicit prior-to-sensor alignment in SN-LogBias stabilizes the normalization and further reduces the mean rank to 2.44.
Our SN-LogBias-Robust achieves the lowest mean rank among these variants and improves both REL and $\delta_{1.05}$ on several benchmarks. 
For instance, it reduces REL by 18.6\% on HouseCat6D and by 51.8\% on ROBI, while increasing $\delta_{1.05}$ by 3.04 and 8.76 points, respectively.
\rui{We further examine the contribution of ADF+Robust-SN by independently enabling or disabling ADF+Robust-SN and UCFC (Appendix~\ref{app:adf_sn_ucfc_decomposition}).
When UCFC is disabled, adding ADF+Robust-SN improves mean rank from 3.41 to 2.78. With UCFC enabled, the additional improvement is smaller (1.94 to 1.88).
In the latter setting, both REL and $\delta_{1.05}$ improve on five datasets but degrade on GraspNet-Trans, YCB-V, and XYZ-IBD.
These results show that ADF+Robust-SN improves the overall ranking even without UCFC, while its additional benefit alongside UCFC varies across datasets.}

Fig.~\ref{fig:sn_vis} illustrates why robust initialization matters.
We compute a consistency weight map $w(x,y)$ from sensor--prior disagreement, so pixels with large inconsistencies receive low weights (purple) and contribute less when estimating the normalization parameters (scale and anchor depth).
As a result, the normalization is less biased by local depth failures (red arrow), leading to lower restoration error compared with the median-based initialization.
\rui{To evaluate the robustness of ADF alignment to biased sensor measurements, we conduct a synthetic corruption stress test on valid depth pixels. MAD-weighted alignment achieves lower initialization and final REL than LS on TransCG and GraspNet-Trans across all tested corruption levels, while the two methods yield comparable final REL on HAMMER. Full results are provided in Appendix Table~\ref{tab:structured_corruption}.}

\noindent\textbf{Uncertainty Modeling Improves Object-region Depth Learning.}
Table~\ref{tab:unc_scene_object} compares models trained with and without uncertainty modeling under mixed-dataset pretraining.
\rui{Across HAMMER, TransCG, and ROBI, uncertainty modeling reduces REL by 6.9--54.1\% on object pixels versus 4.8--54.7\% on full scenes, and increases $\delta_{1.05}$ by 3.3--10.1\% on objects versus 2.4--2.8\% on scenes. Object-region REL gains are larger on HAMMER and TransCG but slightly smaller on ROBI.}
Notably, on non-Lambertian-surface--rich benchmarks such as HAMMER and TransCG, uncertainty modeling yields pronounced object-region gains and also provides effective ranking of high-error pixels, as reflected by low AUSE(REL).
These results indicate that learned uncertainty provides an inductive bias that prioritizes challenging object pixels during optimization, which is most critical for robotic manipulation.

\begin{table}[t]
\centering
\caption{Effect of uncertainty modeling on zero-shot benchmarks.
values in red denote improvements.
}
\begingroup
\small
\setlength{\tabcolsep}{3pt}

\begin{tabular}{@{}lccccc@{}}
\toprule
& \multicolumn{2}{c}{w/o Unc} & \multicolumn{3}{c}{w/ Unc} \\
\cmidrule(lr){2-3}\cmidrule(lr){4-6}
Dataset & REL$\downarrow$ & $\delta_{1.05}\uparrow$ & REL$\downarrow$ & $\delta_{1.05}\uparrow$ & AUSE(REL)$\downarrow$ \\
\midrule
\multicolumn{6}{l}{\textbf{Scene pixels}} \\
HAMMER  & 0.0398 & 79.05 & \makecell[c]{0.0379\\[-0.4ex]{\footnotesize\textcolor{red}{($\downarrow$4.77\%)}}} & \makecell[c]{80.98\\[-0.4ex]{\footnotesize\textcolor{red}{($\uparrow$2.44\%)}}} & 0.00131 \\
TransCG & 0.0115 & 94.55 & \makecell[c]{0.0085\\[-0.4ex]{\footnotesize\textcolor{red}{($\downarrow$26.09\%)}}} & \makecell[c]{96.81\\[-0.4ex]{\footnotesize\textcolor{red}{($\uparrow$2.39\%)}}} & 0.00069 \\
ROBI    & 0.0053 & 96.94 & \makecell[c]{0.0024\\[-0.4ex]{\footnotesize\textcolor{red}{($\downarrow$54.72\%)}}} & \makecell[c]{99.61\\[-0.4ex]{\footnotesize\textcolor{red}{($\uparrow$2.75\%)}}} & 0.00021 \\
\midrule
\multicolumn{6}{l}{\textbf{Object pixels}} \\
HAMMER  & 0.0348 & 80.34 & \makecell[c]{0.0324\\[-0.4ex]{\footnotesize\textcolor{red}{($\downarrow$6.90\%)}}} & \makecell[c]{82.96\\[-0.4ex]{\footnotesize\textcolor{red}{($\uparrow$3.26\%)}}} & 0.00105 \\
TransCG & 0.0279 & 81.42 & \makecell[c]{0.0193\\[-0.4ex]{\footnotesize\textcolor{red}{($\downarrow$30.82\%)}}} & \makecell[c]{89.66\\[-0.4ex]{\footnotesize\textcolor{red}{($\uparrow$10.12\%)}}} & 0.00060 \\
ROBI    & 0.0209 & 90.01 & \makecell[c]{0.0096\\[-0.4ex]{\footnotesize\textcolor{red}{($\downarrow$54.07\%)}}} & \makecell[c]{99.12\\[-0.4ex]{\footnotesize\textcolor{red}{($\uparrow$10.12\%)}}} & 0.00028 \\
\bottomrule
\end{tabular}

\endgroup
\label{tab:unc_scene_object}
\end{table}

\noindent\textbf{Inference Time and Deployment-cost Analysis.}
We benchmark all methods on an Intel(R) Xeon(R) Gold 6226R CPU @ 2.90GHz with a single NVIDIA GeForce RTX 3090, using one $518{\times}518$ RGB-D pair as input. 
Our model uses five ConvGRU iterations. 
For all DAV2-based methods, we adopt the ViT-L backbone~\cite{depthanythingv2} and keep all other implementation details aligned with their official releases.
Table~\ref{tab:inference_time}(a) shows that our method requires 225.6 ms per image, which is slower than PromptDA at 103.4 ms, but clearly faster than OMNI-DC at 277.0 ms, PriorDA at 297.0 ms, and CDM at 383.4 ms among DAV2-based approaches, and is roughly an order of magnitude faster than the diffusion-based D$^3$RoMa at 2116.6 ms. 
In our pick-and-place robotic platform, motion planning takes approximately 1.3 s, and prior bin-picking systems often operate with second-scale pick cycles even when motion planning is optimized~\cite{ichnowski2020gomp}.
In this context, the 225.6 ms depth-restoration latency, although not strictly real-time, is adequate for low-speed or quasi-static manipulation where perception updates at a few Hz are sufficient.
It is therefore practical as a pre-planning perception step, but is not intended for high-rate closed-loop replanning or visual servoing.

Table~\ref{tab:inference_time}(b) further reports the deployment cost of our model, including 349.2M parameters, 13.8M trainable parameters, a 1.43 GB checkpoint, 1367.2G FLOPs, 2.09 GB inference peak VRAM, 19.4 GB/GPU training peak VRAM, and 24 GPU-days of training cost.
Table~\ref{tab:inference_time}(c) further breaks down our runtime: about three quarters of the cost comes from DAV2 feature extraction and RGB and sensor context features preparation (118.5 ms in the backbone plus 57.4 ms in the context networks), while robust scale normalization and ConvGRU refinement together account for only about one quarter (6.1 ms and 43.5 ms). 
Thus, most overhead stems from the underlying DAV2 backbone, and the proposed uncertainty-aware refinement adds only a modest extra cost on top of a standard DAV2 forward pass.

\begin{table*}[t]
\centering
\caption{Inference time and deployment-cost.
(a) End-to-end latency of different methods.
(b) Model size, FLOPs, memory, and training cost of \methodname.
(c) Runtime breakdown of \methodname.}
\begingroup
\small
\setlength{\tabcolsep}{3pt}
\begin{adjustbox}{max width=\textwidth}

\begin{tabular}{@{}c@{\hspace{8pt}}c@{\hspace{8pt}}c@{}}

\begin{tabular}[t]{l c}
\multicolumn{2}{c}{(a) Overall methods} \\[2pt]
\toprule
Method & Time (ms) \\
\midrule
SwinDR~\cite{dai2022domain}    & 73.0 \\
PromptDA~\cite{promptda}       & 103.4 \\
OMNI-DC~\cite{omni-dc}         & 277.0 \\
PriorDA~\cite{priorda}         & 297.0 \\
CDM~\cite{cdm}                 & 383.4 \\
D$^3$RoMa~\cite{d3roma}        & 2116.6 \\
\midrule
Ours                           & 225.6 \\
\bottomrule
\end{tabular}
&
\begin{tabular}[t]{l c}
\multicolumn{2}{c}{(b) Deployment cost} \\[2pt]
\toprule
Item & Value \\
\midrule
Params             & 349.2M \\
Trainable params   & 13.8M \\
Checkpoint size    & 1.43 GB \\
FLOPs              & 1367.2G \\
Infer. peak VRAM   & 2.09 GB \\
Train peak VRAM    & 19.4 GB/GPU \\
Training cost      & 24 GPU-days \\
\bottomrule
\end{tabular}
&
\begin{tabular}[t]{l c}
\multicolumn{2}{c}{(c) Latency breakdown} \\[2pt]
\toprule
Component & Time (ms) \\
\midrule
DAV2 backbone    & 118.5 \\
Robust SN        & 6.1 \\
Context networks & 57.4 \\
ConvGRU          & 43.5 \\
\midrule
Total            & 225.6 \\
\bottomrule
\end{tabular}
\end{tabular}%

\end{adjustbox}
\endgroup
\label{tab:inference_time}
\end{table*}

\section{Applications}
\label{sec:app}
To assess impact beyond depth metrics, we evaluate our restored depth and predicted uncertainty in three downstream tasks that are tightly coupled to robotic manipulation. 
We start with category-level 6D pose estimation (Sec.~\ref{sec:cat_6d}), a core component for object-centric manipulation that relies on depth-derived point clouds and is therefore sensitive to missing or biased measurements on transparent and reflective surfaces. 
We then study 6D grasp synthesis (Sec.~\ref{sec:grasp_syn}), where depth quality directly affects grasp ranking and execution feasibility. 
Finally, in Sec.~\ref{sec:real_world_eval}, we report real-robot results in two manipulation scenarios, including a quantitative pick-and-place evaluation and a bimanual pouring demonstration, to examine whether the restored depth improves robotic manipulation in these settings.

\subsection{Category-level 6D Pose Estimation}
\label{sec:cat_6d}
We first investigate how our restored depth supports downstream manipulation through the task of category-level 6D pose estimation.
Given an RGB-D frame, the goal is to estimate the 6D poses of object instances that are unseen at training time but belong to categories seen during training.

\noindent\textbf{Metrics.}
We evaluate on HouseCat6D~\cite{housecat6d}. Following common practice, we report 3D detection at intersection over union (IoU) thresholds \{25\%, 50\%, 75\%\} and 6D pose accuracy under joint translation/rotation criteria of \{5°2\,cm, 5°5\,cm, 10°2\,cm, 10°5\,cm\}~\cite{nocs}.

\noindent\textbf{Uncertainty-guided Sampling.}
Beyond improving depth quality, we also exploit the predicted uncertainty maps in downstream manipulation tasks.
Given an RGB-D input, a 6D pose estimator usually takes as input a fixed-size point cloud obtained by sampling from the depth map.
We replace random sampling with an uncertainty-guided strategy driven by our predicted per-pixel uncertainty.

Let $\mathbf{U}(x,y) \ge 0$ denote the predicted uncertainty at pixel $(x,y)$.
We first normalize $\mathbf{U}$ to a unit interval via per-image min-max normalization, obtaining $\tilde{\mathbf{U}}(x,y) \in [0,1]$, and define a confidence map
\begin{equation}
\label{eq:confidence_map}
\mathbf{C}(x,y) = 1 - \tilde{\mathbf{U}}(x,y).
\end{equation}
\rui{Because the minimum and maximum are computed separately for each image, $\mathbf{C}$ represents relative sampling weights within that image rather than confidence scores calibrated across frames, and extreme pixels can affect the normalized weights.}
For each pixel $(x,y)$ in the non-zero set $\mathcal{V}$, we interpret $\mathbf{C}(x,y)$ as the sampling probability and draw a binary selection variable
\[
z(x,y) \sim \mathrm{Bernoulli}\bigl(\mathbf{C}(x,y)\bigr).
\]
Only pixels with $z(x,y)=1$ are converted to 3D points and passed to the pose estimator, while the total number of sampled points is kept fixed by random subsampling if necessary.
In this way, regions with lower predicted uncertainty (higher confidence) are more likely to contribute to the point cloud, biasing the pose estimator toward geometrically reliable structures while still preserving spatial coverage.

\begin{table}[t]
\centering
\caption{Category-level 6D pose accuracy of VI-Net~\cite{vi-net} on HouseCat6D~\cite{housecat6d} using different depth inputs.}
\label{tab:cat_6d_pose}
\begingroup
\small
\setlength{\tabcolsep}{1.5pt}

\begin{tabular}{@{}L@{\hspace{3pt}}*{7}{C}@{}}
\toprule
\multirow{2}{*}{Depth} & \multicolumn{3}{c}{\textbf{IoU (\%)}} & \multicolumn{4}{c}{\textbf{Pose accuracy (\%)}} \\
\cmidrule(lr){2-4}\cmidrule(l){5-8}
& @25 & @50 & @75 & $5^\circ$2cm & $5^\circ$5cm & $10^\circ$2cm & $10^\circ$5cm \\
\midrule
Raw                           & 77.3 & 52.6 & 17.2 & 6.7 & 8.9 & 17.6 & 25.4 \\
SwinDR~\cite{dai2022domain}   & 77.3 & 57.8 & 21.9 & 7.5 & 9.9 & 20.9 & 30.1 \\
D$^3$RoMa~\cite{d3roma}       & 71.7 & 42.6 & 12.4 & 2.8 & 4.6 & 10.6 & 17.9 \\
\rowstyle{\color{black}}
CDM~\cite{cdm}                & \textbf{80.7} & 58.4 & 18.4 & 10.0 & 14.7 & 22.4 & 35.6 \\
\midrule
\rowstyle{\color{black}}
Ours
& \rui{\makecell[c]{\underline{79.9}\\[-0.6ex]{\footnotesize$\pm3.1$}}}
& \rui{\makecell[c]{\textbf{60.5}\\[-0.6ex]{\footnotesize$\pm4.3$}}}
& \rui{\makecell[c]{\underline{30.3}\\[-0.6ex]{\footnotesize$\pm2.8$}}}
& \rui{\makecell[c]{\underline{13.3}\\[-0.6ex]{\footnotesize$\pm0.6$}}}
& \rui{\makecell[c]{\underline{15.0}\\[-0.6ex]{\footnotesize$\pm0.5$}}}
& \rui{\makecell[c]{\underline{28.9}\\[-0.6ex]{\footnotesize$\pm1.9$}}}
& \rui{\makecell[c]{\underline{36.7}\\[-0.6ex]{\footnotesize$\pm2.1$}}} \\
\rowstyle{\color{black}}
\makecell[l]{Ours w/\\Unc-Sampling}
& \rui{\makecell[c]{79.4\\[-0.6ex]{\footnotesize$\pm1.1$}}}
& \rui{\makecell[c]{\underline{60.2}\\[-0.6ex]{\footnotesize$\pm1.5$}}}
& \rui{\makecell[c]{\textbf{31.9}\\[-0.6ex]{\footnotesize$\pm2.3$}}}
& \rui{\makecell[c]{\textbf{14.3}\\[-0.6ex]{\footnotesize$\pm1.8$}}}
& \rui{\makecell[c]{\textbf{16.4}\\[-0.6ex]{\footnotesize$\pm1.9$}}}
& \rui{\makecell[c]{\textbf{30.8}\\[-0.6ex]{\footnotesize$\pm2.1$}}}
& \rui{\makecell[c]{\textbf{39.7}\\[-0.6ex]{\footnotesize$\pm2.8$}}} \\
\bottomrule
\end{tabular}
\endgroup
\end{table}

\noindent\textbf{Baselines.}
We use VI-Net~\cite{vi-net} and vary only the depth source and the point-sampling policy for the input cloud, training and evaluating each variant with the same depth type. 
We compare: (i) \emph{Raw} sensor depth; (ii) \emph{SwinDR}~\cite{dai2022domain}; (iii) \emph{D$^3$RoMa}~\cite{d3roma}; (iv) \emph{CDM} (CDM-D435)~\cite{cdm}; (v) \emph{Ours} (restored); and (vi) \emph{Ours (Unc-Sampling)}. 
Variants (i)--(v) use uniform random sampling; (vi) replaces it with uncertainty-guided sampling via the confidence map ($\mathbf{C}(x,y)$ in Eq.~\eqref{eq:confidence_map}). 

\begin{figure}[t]
  \centering
  \includegraphics[width=\linewidth]{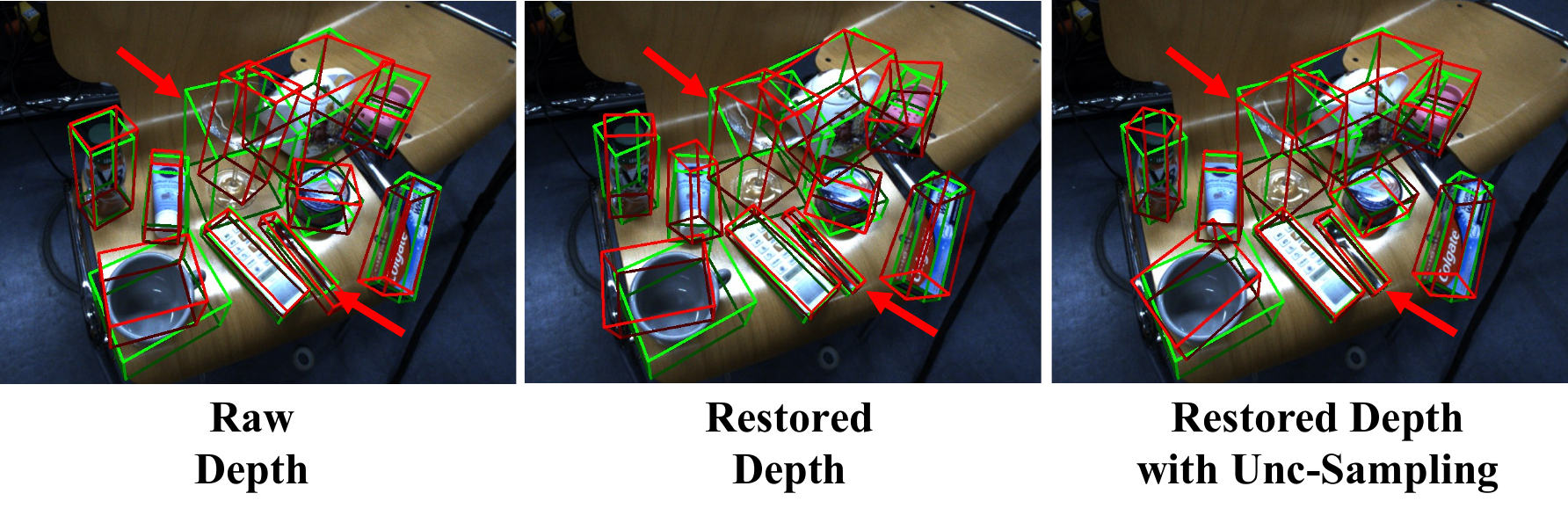}
  \caption{VI-Net~\cite{vi-net} results on HouseCat6D~\cite{housecat6d}. From left to right: using raw sensor depth, our restored depth, and restored depth with uncertainty-guided sampling. GT poses are shown as green 3D bounding boxes and predictions as red boxes. Red arrows highlight improved pose estimation on transparent and reflective objects when using restored depth and uncertainty-guided sampling.
  }
\label{fig:cat_pose_vis}
\end{figure}

\noindent\textbf{Results.}
Table~\ref{tab:cat_6d_pose} summarizes VI-Net~\cite{vi-net} on HouseCat6D~\cite{housecat6d} under different depth inputs and sampling policies. 
Using our restored depth improves over the raw baseline across all pose metrics, with clear gains at stricter thresholds: $\mathrm{IoU}@50$ rises from 52.6 to \rui{60.5}, $\mathrm{IoU}@75$ from 17.2 to \rui{30.3}, and $5^\circ 2\text{cm}$ / $10^\circ 5\text{cm}$ from 6.7 / 25.4 to \rui{13.3 / 36.7}.
\rui{Compared with CDM, our restored depth achieves higher mean scores on six of seven metrics, while CDM retains the highest $\mathrm{IoU}@25$ score.}
\rui{The configuration trained and evaluated with uncertainty-guided sampling achieves the highest mean scores on five of seven metrics}, including \rui{$\mathrm{IoU}@75=31.9$, $5^\circ2\text{cm}=14.3$, $5^\circ5\text{cm}=16.4$, and $10^\circ5\text{cm}=39.7$. These results are consistent with prioritizing high-confidence pixels when constructing the input point cloud, which may benefit pose estimation under tighter criteria.}
Qualitative examples in Fig.~\ref{fig:cat_pose_vis} show corresponding improvements, where restored depth and uncertainty-guided sampling correct misaligned poses, particularly on transparent and reflective objects highlighted by the red arrows.

We also find that category-level pose estimation is sensitive to the depth domain. Restored depth should therefore not be viewed as a universal drop-in replacement for pose estimators trained only on raw sensor depth. Although \methodname\ removes sensor-specific holes and material-induced artifacts, it also changes the depth distribution. \rui{Appendix Table~\ref{tab:cat_6d_pose_full} shows that switching to restored depth only at test time lowers VI-Net's IoU@75 from 17.2 to 11.4, while training and testing with restored depth yields 30.3. Mixed raw/restored-depth training is another possible way to address this domain shift.} Other adaptation strategies, such as distillation or lightweight domain adaptation~\cite{zhao2023domain}, may also help bridge this gap. In summary, training VI-Net with our restored depth improves 6D pose accuracy without additional labeling or data collection, while uncertainty-guided sampling adds no extra supervision and only minimal runtime cost.



\begin{table*}[t]
\centering
\caption{6D grasp synthesis performance of AnyGrasp~\cite{anygrasp} on \rui{GN-Trans} under different depth inputs.}
\label{tab:6d_grasp}
\setlength{\tabcolsep}{6.5pt}

\begin{tabular}{LCCCCCCCCCC}
\toprule
\multirow{2}{*}{Depth Type} 
& \multicolumn{3}{c}{Seen} 
& \multicolumn{3}{c}{Similar} 
& \multicolumn{3}{c}{Novel} 
& \multirow{2}{*}{Mean AP} \\ 
\cline{2-10}
& AP & AP$_{0.8}$ & AP$_{0.4}$ 
& AP & AP$_{0.8}$ & AP$_{0.4}$ 
& AP & AP$_{0.8}$ & AP$_{0.4}$ & \\ 
\midrule
Raw                                   & 23.15 & 27.73 & 19.81 & 17.97 & 21.86 & 15.15 &  6.48 &  8.05 &  3.36 & 15.87 \\
SwinDR~\cite{dai2022domain}           & 28.95 & 35.45 & 22.41 & 22.51 & 27.82 & 17.32 & 10.20 & 12.64 &  4.70 & 20.55 \\
$\text{D}^3\text{RoMa}$ (RGB\text{-}D)~\cite{d3roma}      & 26.52 & 33.86 & 16.19 & 22.45 & 28.38 & 15.02 &  9.53 & 11.59 &  3.79 & 19.50 \\
$\text{D}^3\text{RoMa}$ (RGB\text{-}D Stereo)~\cite{d3roma} & \underline{33.01} & \textbf{40.31} & 26.32 & \underline{25.82} & \underline{31.91} & 19.63 & 10.89 & 13.41 &  4.92 & \underline{23.24} \\
\rowstyle{\color{black}}
CDM~\cite{cdm}              & 28.28 & 35.66 & 20.42 & 24.56 & 29.27 & 17.62 &  9.78 & 11.80 & 5.52 & 20.87 \\
\midrule
Ours (RGB-D)                                   & 32.67 & 39.08 & \underline{27.86} & 25.48 & 31.14 & \underline{20.72} & \underline{11.26} & \underline{13.91} & \underline{5.98} & 23.14 \\
Ours (RGB-D, Unc\text{-}Reweighting)          & \textbf{33.54} & \underline{40.10} & \textbf{28.59} & \textbf{26.62} & \textbf{32.54} & \textbf{21.74} & \textbf{11.89} & \textbf{14.72} & \textbf{6.29} & \rui{\textbf{24.02}} \\
\bottomrule
\end{tabular}

\end{table*}

\subsection{6D Grasp Synthesis}
\label{sec:grasp_syn}
We further examine how our restored depth and predicted uncertainty can benefit downstream grasping. 
We consider 6D grasp synthesis and evaluate on GraspNet-Trans using AnyGrasp~\cite{anygrasp} as the grasp predictor.

\noindent\textbf{Metrics.}
GraspNet-Trans is built on the GraspNet-1Billion benchmark~\cite{graspnet1b}, sharing the same object set, CAD models, and grasp annotations. 
Following the official GraspNet protocol, the testing scenes are divided into three splits: \emph{Seen}, \emph{Similar}, and \emph{Novel}, depending on whether the object instances appear in the training set. 
Each predicted grasp is represented as
$g = [R,\, t,\, \omega,\, s]$,
where $R \in \mathrm{SO}(3)$ and $t \in \mathbb{R}^3$ denote the gripper pose, $\omega$ is the gripper width, and $s$ is the grasp score.
A grasp is counted as correct if it satisfies the official collision and force-closure checks.
For each test scene and camera view, we sort predicted grasps by $s$ and compute average precision (AP) on the top-50 predictions. 
We report these AP values separately for the Seen, Similar, and Novel splits, together with the mean AP over all three.

\noindent\textbf{Uncertainty-guided Reweighting.}
AnyGrasp~\cite{anygrasp} parameterizes each grasp hypothesis by a surface point on the object as its \emph{center}, together with additional grasp parameters such as view rotation, in-plane rotation, offset, and gripper width. 
We make the simple assumption that the reliability of the depth at this center point is a key factor for the success of the corresponding grasp. 
Therefore, we use our confidence map $\mathbf{C}(x,y)$ (Eq.~\eqref{eq:confidence_map}) to reweight the scores of predicted grasps at inference time. 

Specifically, for a grasp $g$ with original AnyGrasp score $s_g$, we sample $C_g=\mathbf{C}(x_g,y_g)$ at its center and compute the reweighted score $s'_g=C_g\,s_g$. 
We apply this to all candidates in a scene and re-rank them by $s'_g$ before computing AP, promoting grasps anchored on reliable geometry and down-weighting those on uncertain depth. 
We do not use uncertainty-guided point sampling here, as it can discard informative edge points and reduce grasp accuracy; instead, we use uncertainty only at the scoring stage.
A more detailed analysis of these design choices is provided in the Appendix~\ref{app:6d_grasp}.
\rui{Alternative uncertainty reweighting strategies are compared in Appendix Table~\ref{tab:unc_grasp}.}

\noindent\textbf{Baselines.}
We freeze the pretrained AnyGrasp~\cite{anygrasp} and vary only the depth input and whether test-time uncertainty is used.
We compare: (i) Raw sensor depth; (ii) SwinDR~\cite{dai2022domain}; (iii) \emph{$\text{D}^3\text{RoMa}$ (RGB-D)}; (iv) \emph{$\text{D}^3\text{RoMa}$ (RGB-D Stereo)}~\cite{d3roma}; (v) \textit{CDM} (CDM-D435)~\cite{cdm}; (vi) \emph{Ours} (restored); and (vii) \emph{Ours (Unc-Reweighting)}, which reweights grasps by our confidence map.
$\text{D}^3\text{RoMa}$ (RGB-D Stereo) consumes stereo image pairs in addition to depth, providing richer geometry than single-depth methods. 
We do not fine-tune AnyGrasp; differences therefore reflect only the depth source and the reweighting.

\noindent\textbf{Results.}
As shown in Table~\ref{tab:6d_grasp}, our restored depth already surpasses SwinDR~\cite{dai2022domain}, $\text{D}^3\text{RoMa}$ (RGB-D)~\cite{d3roma}, and CDM~\cite{cdm} on most splits and metrics, improving Mean AP to 23.14 compared with 20.55, 19.50, and 20.87, respectively.
$\text{D}^3\text{RoMa}$ (RGB-D Stereo)~\cite{d3roma} benefits from extra stereo cues and attains the strongest $\mathrm{AP}_{0.8}$ on the Seen split, exceeding our method without Unc-Reweighting in terms of mean AP.
Adding uncertainty-guided reweighting improves the mean AP, yielding the best mean AP among the evaluated methods.
These gains indicate that, on GraspNet-Trans, using $\mathbf{U}$ to modulate grasp scores can improve candidate ranking.
However, this reweighting only reflects depth reliability and does not directly model contact geometry, or physical grasp stability.
We revisit this limitation through the real-robot failure case in Sec.~\ref{sec:real_world_eval}.

\begin{figure}[t]
  \centering
  \includegraphics[width=\linewidth]{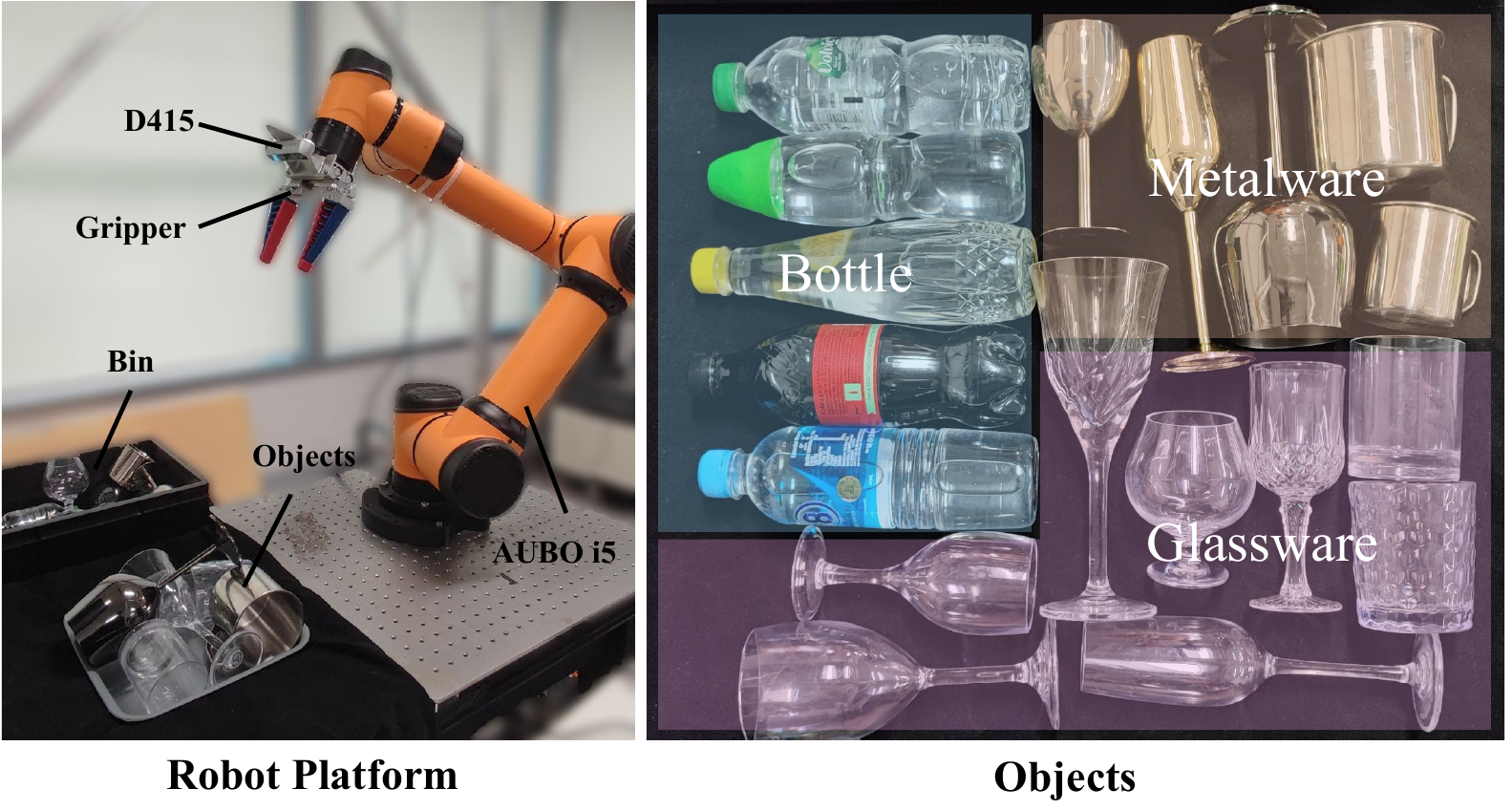}
    \caption{Robot platform and the full set of test objects used for real-world evaluation of 6D grasp synthesis. Our test set emphasizes challenging household items with transparent and reflective surfaces.}
\label{fig:robot_plat}
\end{figure}

\subsection{Real-world Evaluation}
\label{sec:real_world_eval}
\noindent\textbf{Pick-and-place.}
We deploy our mixed-dataset pretrained model on a real robotic pick-and-place system in a zero-shot manner, without any task- or sensor-specific fine-tuning.
The system uses an AUBO--i5 arm, a RobotDigg two-finger gripper, and an eye-in-hand Intel RealSense D415 RGB-D camera (Fig.~\ref{fig:robot_plat}). 
We build cluttered scenes with 5 objects, focusing on challenging materials such as transparent and reflective items, and evaluate all depth sources on the same 20 fixed scenes; each scene is sampled from the same 18-object pool, with object sets and placements matched as closely as possible across methods.
We use AnyGrasp~\cite{anygrasp} for 6D grasp synthesis and keep the grasping pipeline and control parameters identical across methods, varying only the depth used for point-cloud back-projection.

Following the TransCG protocol~\cite{fang2022transcg}, \rui{an object is removed and counted as not cleared once it accumulates three failed grasp attempts.}
We report Success Rate (SR) as successful executions over attempts, and Completion Rate (CR) as cleared objects over target objects; both metrics are first computed per scene and then reported as \rui{scene-level means with 95\% percentile-bootstrap confidence intervals} in Table~\ref{tab:real_robot_grasp}.
\rui{The detailed counting protocol, per-scene execution records, and scene-level bootstrap analysis, including paired confidence intervals, are provided in Appendix~\ref{app:pick-and-place}.}

\begin{table}[t]
\centering
\caption{Real-robot pick-and-place results on the same 20 fixed scenes. \rui{We report scene-level means with 95\% percentile-bootstrap confidence intervals in brackets.} Our method and CDM use RGB-D inputs, while D$^3$RoMa uses RGB-D with active infrared stereo.}
\label{tab:real_robot_grasp}
\resizebox{\linewidth}{!}{
\begin{tabular}{lcc}
\toprule
Input Depth & \rui{SR (\%)} $\uparrow$ & \rui{CR (\%)} $\uparrow$ \\
\midrule
Raw & \rui{54.42 [45.33, 64.31]} & \rui{88.00 [81.00, 94.00]} \\
D$^3$RoMa & \rui{69.05 [62.36, 76.34]} & \rui{96.00 [92.00, 99.00]} \\
CDM  & \rui{72.73 [66.55, 78.69]} & \rui{98.00 [95.00, 100.00]} \\
Ours & \textbf{\rui{76.65 [71.57, 81.82]}} & \textbf{\rui{99.00 [97.00, 100.00]}} \\
Ours (Unc-Reweighting) & \rui{70.25 [62.60, 77.35]} & \rui{93.75 [88.25, 98.00]} \\
\bottomrule
\end{tabular}
}
\end{table}

\begin{figure}[t]
  \centering
  \includegraphics[width=\linewidth]{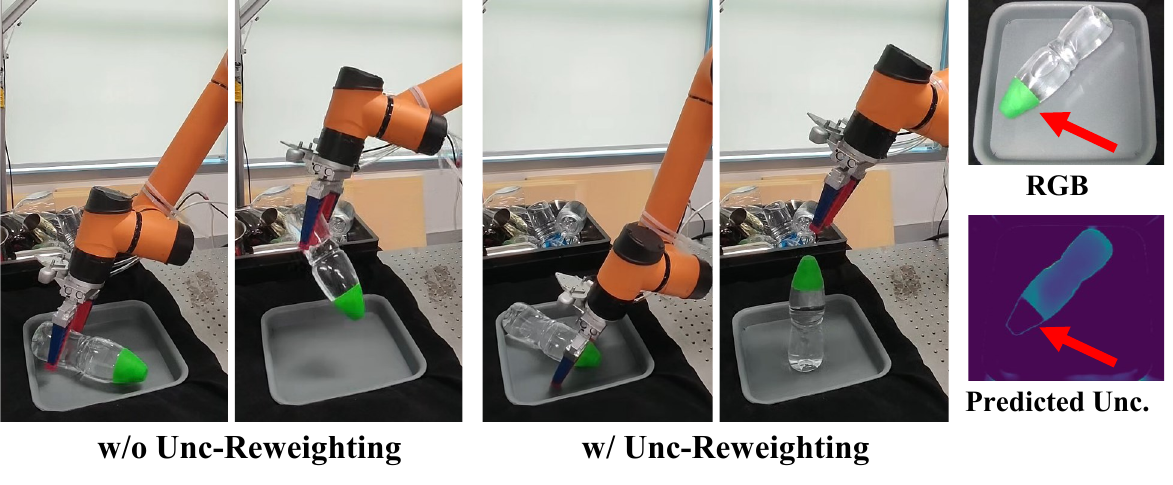}
  \caption{Failure case of uncertainty-guided reweighting: the predicted uncertainty is locally low on an off-center region of a multi-material object (highlighted), biasing the selection toward an unstable off-center grasp.}
  \label{fig:unc_reweight_fail}
\end{figure}

Table~\ref{tab:real_robot_grasp} shows that, in our pick-and-place setup, our restored depth achieves the highest scene-level mean SR and CR among the compared depth sources. It improves SR from 54.42\% to 76.65\% and CR from 88.00\% to 99.00\% over raw depth, and also obtains higher mean SR/CR than D$^3$RoMa and CDM.
We also test uncertainty-guided grasp reweighting as a plug-in at inference time. 
In our setup, it does not provide a clear benefit: 
\rui{the observed mean SR decreases from 76.65\% to 70.25\%, and CR from 99.00\% to 93.75\%.}
As shown in Fig.~\ref{fig:unc_reweight_fail}, a naive reweighting strategy may bias grasps toward locally low-uncertainty regions that are not aligned with stable grasp geometry, leading to off-center and physically unstable grasps.
Taken together, these results provide deployment evidence that \methodname\ improves real-world pick-and-place mainly through restored depth geometry, without task- or sensor-specific fine-tuning. 
They also show that lower depth uncertainty does not necessarily imply a more stable grasp. 
\rui{The paired confidence intervals support improvements over raw sensor depth, while those for comparisons with D$^3$RoMa and CDM, as well as the incremental effect of uncertainty-guided reweighting, include or touch zero (Appendix Table~\ref{tab:real_robot_paired_diff}).}

\begin{figure}[t]
  \centering
  \includegraphics[width=\linewidth]{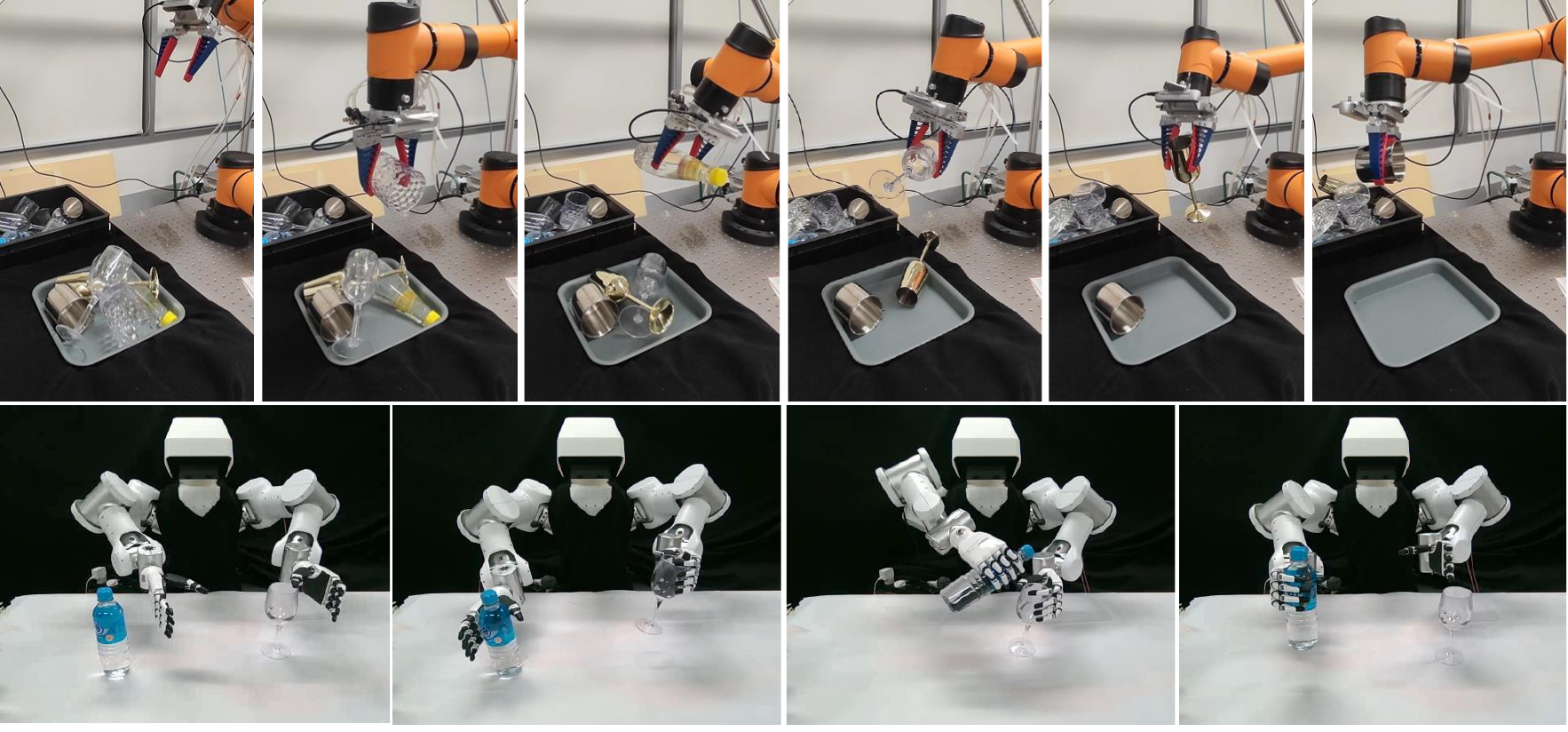}
  \caption{Execution sequences of real-world manipulation tasks: the top row shows pick-and-place, and the bottom row shows a long-horizon bimanual pouring demo.}
  \label{fig:grasp_demo}
\end{figure}

\noindent\textbf{Bimanual manipulation.}
We also include a long-horizon bimanual pouring example on a custom dual-arm platform equipped with Inspire Robotics RH56DFX dexterous hands. A single head-mounted ZED~2i provides RGB-D observations. The task is to grasp a bottle with the right arm, grasp a transparent cup with the left arm, and pour water from the bottle into the cup. Demonstration data are collected via teleoperation following~\cite{cheng2025open}. We use Pi$_{0.5}$~\cite{pi0.5} as the policy model, feeding our restored depth to the policy and predicting joint actions for both arms; the policy is fine-tuned using the officially recommended settings. Fig.~\ref{fig:grasp_demo} and the supplementary video show representative executions of this bimanual pouring task.

\section{Limitation}
\label{sec:limitation}
\rui{First, our failure analysis of zero-shot depth restoration shows that uncertainty feedback can be less effective at correcting depth errors in regions with biased sensor observations or inaccurate initialization, as detailed in Appendix~\ref{app:failure_taxonomy}.
Leveraging consistency across successive refinement steps may help mitigate such performance degradation.}

Second, restoring depth for scenes with stacked transparent objects remains a challenge, because multiple valid surfaces can correspond to a single RGB pixel along the same camera ray. 
%
This violates the single-depth-per-pixel assumption and can cause unstable predictions around transparent boundaries.
Addressing this may require multi-layer depth representations or multi-view cues~\cite{wen2025seeing}.

Third, inference latency is still a bottleneck.
While our iterative co-refinement improves accuracy, the overall runtime is mainly limited by the frozen DAV2 backbone and context extraction (Table~\ref{tab:inference_time}).
To achieve real-time efficiency, we may need lightweight backbones, feature distillation, or reduced-resolution refinement strategies.

Finally, although we exploit predicted uncertainty in downstream tasks via point sampling or score reweighting, these strategies remain heuristic and can be brittle in some cases (e.g., off-center grasps on multi-material objects; Fig.~\ref{fig:unc_reweight_fail}).
A more systematic and task-adaptive integration of uncertainty, for example through learning-based or policy-level coupling, is an important direction for future work.


\section{Conclusion}
\label{sec:conclusion}
We introduce \methodname, a depth restoration framework that uses predicted uncertainty as recurrent feedback to guide metric depth refinement under non-Lambertian failures in commodity RGB-D sensing.
In this framework, DAV2 backbone with ADF provides a metrically grounded dense initialization, Robust-SN stabilizes log-depth relative-to-metric normalization under biased sensor measurements, and UCFC turns the evolving uncertainty estimate into an active refinement signal.
Across three in-domain datasets, \methodname\ substantially improves metric depth accuracy over prior restoration/completion methods and produces uncertainty estimates that correlate with restoration errors.
We also present an eight-dataset benchmark for cross-dataset transfer spanning diverse scenes, materials, sensors, and capture conditions, on which \methodname\ attains the best overall performance among strong baselines.
Beyond depth metrics, the evaluated pose-estimation and grasp-synthesis pipelines show that the predicted uncertainty can support manipulation-oriented perception.
Future work will explore consistency-aware refinement, multi-layer or multi-hypothesis depth representations for layered transparency, lightweight backbones or distillation for lower latency, and more principled integration of uncertainty into manipulation policies.









\nocite{peyre2019computational,rabin2012wasserstein,gretton2012kernel}

\bibliographystyle{IEEEtran}
\bibliography{reference}


\clearpage
\appendices
\setcounter{topnumber}{3}
\setcounter{bottomnumber}{2}
\setcounter{totalnumber}{5}
\setcounter{dbltopnumber}{3}
\renewcommand{\topfraction}{0.9}
\renewcommand{\bottomfraction}{0.8}
\renewcommand{\textfraction}{0.05}
\renewcommand{\floatpagefraction}{0.6}
\renewcommand{\dbltopfraction}{0.9}
\renewcommand{\dblfloatpagefraction}{0.6}
\setcounter{table}{0}
\renewcommand{\thetable}{A-\Roman{table}}
\renewcommand{\theHtable}{appendix.\arabic{table}}
\setcounter{figure}{0}
\renewcommand{\thefigure}{A-\arabic{figure}}
\renewcommand{\theHfigure}{appendix.\arabic{figure}}
\setcounter{algorithm}{0}
\renewcommand{\thealgorithm}{A-\Roman{algorithm}}

\section{Additional Experimental Details}
\label{app:exp_details}

\subsection{\rui{Forward-Pass Details}}
\label{app:forward_pass}
\rui{Algorithm~\ref{alg:ua_depth_forward} summarizes the inference procedure of the final mixed-dataset model using DAV2 and Robust-SN. The in-domain alignment is described in Sec.~\ref{sec:adf}.}

\begin{algorithm*}[t]
\caption{\rui{Inference forward pass of \methodname\ with Robust-SN}}
\label{alg:ua_depth_forward}
\footnotesize
\begin{algorithmic}[1]
\STATE \rui{\textbf{Input:} RGB image $\mathbf I$, sensor depth $\mathbf D_{\mathrm{sen}}$, iterations $K$.}
\STATE \rui{$(\mathbf q,\mathbf F_{\mathrm{img}})\gets\operatorname{DAV2}_{\mathrm{frozen}}(\mathbf I)$; $\mathbf D_{\mathrm{rel}}(x,y)\gets1/\mathbf q(x,y)$ if $\mathbf q(x,y)>0$, otherwise $0$.}
\STATE \rui{$\mathcal V\gets\{(x,y):\mathbf D_{\mathrm{sen}}(x,y)>d_{\min}\}$; set $\varepsilon=10^{-6}$ and clamp depth inputs by $\varepsilon$ before taking logarithms.}
\IF{\rui{$\mathcal V\neq\varnothing$}}
  \STATE \rui{Estimate provisional $a_0$ from selected valid pixels (Sec.~\ref{sec:implementation_details}); set $a_0\gets0$ if fewer than 16 candidates remain.}
  \STATE \rui{$r(x,y)\gets\left|\log\mathbf D_{\mathrm{sen}}(x,y)-(\log\mathbf D_{\mathrm{rel}}(x,y)+a_0)\right|$ for $(x,y)\in\mathcal V$.}
  \STATE \rui{$m_r\gets\operatorname{median}_{\mathcal V}r$; $\tau\gets\kappa\max\!\bigl(\operatorname{median}_{\mathcal V}|r-m_r|,\varepsilon\bigr)$, $\kappa=2.2239$; $w(x,y)\gets\exp(-r(x,y)/\tau)$.}
  \IF{\rui{$\sum_{(x,y)\in\mathcal V}w(x,y)>\varepsilon$}}
    \STATE \rui{$a\gets\operatorname{median}^{w}_{\mathcal V}(\log\mathbf D_{\mathrm{sen}}-\log\mathbf D_{\mathrm{rel}})$; $s_{\mathrm{anc}}\gets\operatorname{median}^{w}_{\mathcal V}(\mathbf D_{\mathrm{sen}})$.}
  \ELSE
    \STATE \rui{$a\gets\operatorname{median}_{\mathcal V}(\log\mathbf D_{\mathrm{sen}}-\log\mathbf D_{\mathrm{rel}})$; $s_{\mathrm{anc}}\gets\operatorname{median}_{\mathcal V}(\mathbf D_{\mathrm{sen}})$.}
  \ENDIF
\ELSE
  \STATE \rui{$a\gets0$; $s_{\mathrm{anc}}\gets\max\!\left(\operatorname{median}\bigl(\max(\mathbf D_{\mathrm{rel}},\varepsilon)\bigr),\varepsilon\right)$.}
\ENDIF
\STATE \rui{$\log\mathbf D'_{\mathrm{rel}}\gets\log\mathbf D_{\mathrm{rel}}+a$; $\mathbf D'_{\mathrm{rel}}\gets\exp(\log\mathbf D'_{\mathrm{rel}})$.}
\STATE \rui{$\tilde{\mathbf D}_{\mathrm{sen}}\gets\log\mathbf D_{\mathrm{sen}}-\log s_{\mathrm{anc}}$; $\tilde{\mathbf D}_{\mathrm{rel}}\gets\log\mathbf D'_{\mathrm{rel}}-\log s_{\mathrm{anc}}$.}
\STATE \rui{$\tilde{\mathbf D}^{0}\gets\mathrm{ADF}(\tilde{\mathbf D}_{\mathrm{sen}},\tilde{\mathbf D}_{\mathrm{rel}};\mathcal V)$, preserving valid sensor depths.}
\STATE \rui{$\mathbf U^0(x,y)\gets|\mathbf D'_{\mathrm{rel}}(x,y)-\mathbf D_{\mathrm{sen}}(x,y)|/\max(\mathbf D'_{\mathrm{rel}}(x,y),\varepsilon)$ on $\mathcal V$, otherwise $1$.}
\STATE \rui{Downsample $\tilde{\mathbf D}^{0}$ and $\mathbf U^0$ to the refinement resolution; denote the depth state by $\mathbf D^0$.}
\STATE \rui{Extract $\mathbf F'_{\mathrm{img},j}$ and $\mathbf F_{\mathrm{sen},j}$; form context $\mathbf F_j\gets[\mathbf F'_{\mathrm{img},j},\mathbf F_{\mathrm{sen},j}]$ and initialize $\mathbf H_j^0\gets\tanh(\mathbf F_j)$, $j\in\{4,8,16\}$.}
\FOR{\rui{$k=0,\ldots,K-1$}}
  \STATE \rui{$\mathbf M_{\mathrm{sen}}^k\gets\sigma(\mathrm{Conv}_{\mathrm{sen}}(\mathbf U^k))$; $\mathbf M_{\mathrm{img}}^k\gets\sigma(\mathrm{Conv}_{\mathrm{img}}(\mathbf U^k))$.}
  \STATE \rui{$\tilde{\mathbf F}_{\mathrm{sen}}^k\gets\mathbf M_{\mathrm{sen}}^k\odot\mathbf F_{\mathrm{sen}}$; $\tilde{\mathbf F}_{\mathrm{img}}^k\gets\mathbf M_{\mathrm{img}}^k\odot\mathbf F'_{\mathrm{img}}$; $\mathbf F^k\gets[\tilde{\mathbf F}_{\mathrm{sen}}^k,\tilde{\mathbf F}_{\mathrm{img}}^k]$.}
  \STATE \rui{$\mathbf X^k\gets[\mathbf D^k,\mathbf U^k]$; update ConvGRU states from $\mathbf H^k$ to $\mathbf H^{k+1}$ using $\mathbf X^k$ and $\mathbf F^k$, then predict $(\Delta\mathbf D^k,\Delta\mathbf U^k)$ from $\mathbf H^{k+1}$.}
  \STATE \rui{$\mathbf D^{k+1}\gets\mathbf D^k+\Delta\mathbf D^k$; $\mathbf U^{k+1}\gets\mathbf U^k+\Delta\mathbf U^k$.}
\ENDFOR
\STATE \rui{Upsample $(\mathbf D^K,\mathbf U^K)$ using the predicted convex mask when available, then resize to input resolution; denote the full-resolution states by $(\tilde{\mathbf D},\tilde{\mathbf U})$.}
\STATE \rui{$\mathbf D\gets s_{\mathrm{anc}}\exp(\tilde{\mathbf D})$ (clamped to the configured depth range); $\mathbf U\gets\max(\operatorname{softplus}(\tilde{\mathbf U}),\mathbf U_{\min})$; \textbf{return} $(\mathbf D,\mathbf U)$.}
\end{algorithmic}
\end{algorithm*}


\subsection{In-domain Depth Estimation Datasets}
\label{app:in_domain_datasets}
We evaluate in-domain fitting ability on three public RGB-D datasets, HAMMER~\cite{jung2023importance}, DREDS/STD~\cite{dai2022domain}, and ClearPose~\cite{chen2022clearpose}. For this setting, models are trained and tested on the official splits of each dataset.

\noindent\textbf{HAMMER.}
HAMMER contains 13 indoor scenes (7{,}200 frames) with polarization-guided RGB-D and high-fidelity rendered depth~\cite{jung2023importance}. 
We use it for both in-domain and zero-shot evaluations.

\noindent\textbf{DREDS and STD.}
DREDS is a domain-randomized synthetic dataset designed for non-Lambertian depth sensing, and it provides two evaluation subsets: DREDS-CatKnown (seen categories) and DREDS-CatNovel (novel categories with material perturbations)~\cite{dai2022domain}. STD is the real-world counterpart following the same known and novel protocol (STD-CatKnown and STD-CatNovel). Unless otherwise stated, we train on DREDS-CatKnown and report results on all four subsets to assess both in-domain performance and category and material generalization.

\noindent\textbf{ClearPose.}
ClearPose provides 350k real images of transparent and translucent objects captured using a RealSense L515 across diverse scenes, backgrounds, occlusions, non-planar supports, covers, and liquids~\cite{chen2022clearpose}. We follow the official split and evaluate on all ClearPose test settings.

\subsection{Zero-shot Generalization Benchmark and Dataset Preparation}
\label{app:zs_benchmark}
To assess zero-shot generalization, we pretrain on a mixture of training splits and evaluate \emph{directly} on the test sets of eight datasets: HAMMER~\cite{jung2023importance}, HouseCat6D~\cite{housecat6d}, TransCG~\cite{fang2022transcg}, GraspNet-Trans \rui{(GN-Trans)}, XYZ-IBD~\cite{xyz-ibd}, ROBI~\cite{robi}, YCB-V~\cite{ycbv}, and T-LESS~\cite{tless} without any fine-tuning. For fair comparison, the pretraining mix follows~\cite{d3roma} and includes DREDS-CatKnown~\cite{dai2022domain}, ClearPose~\cite{chen2022clearpose}, and HISS~\cite{d3roma}. Table~\ref{tab:datasets_overview} in the main paper summarizes the train and test composition and dataset statistics used in this benchmark.

\begin{figure}[t]
\centering
\subfloat{
\includegraphics[width=\linewidth]{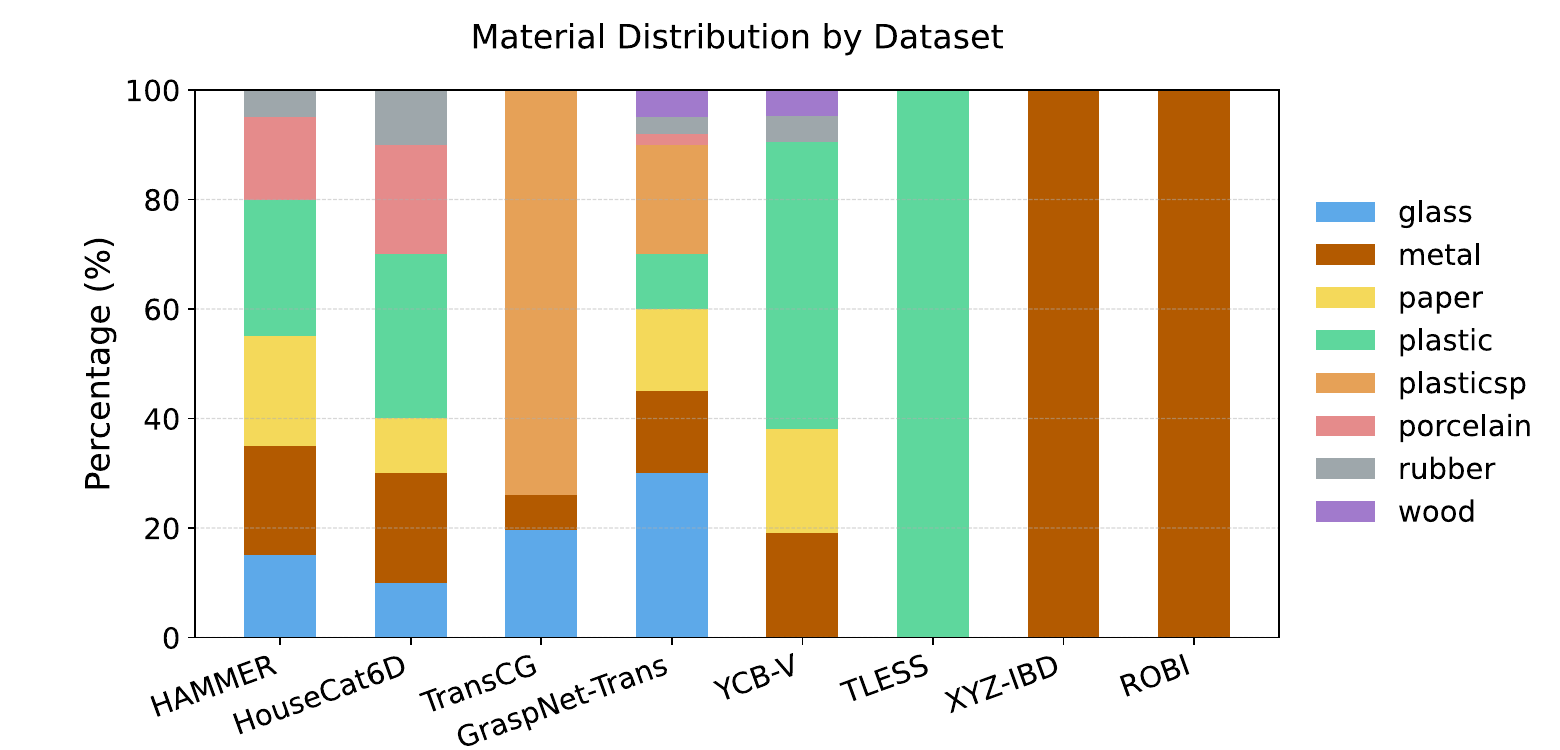}}
\caption{Material distribution across datasets used for zero-shot evaluation. Best view in color.}
\label{fig:mat_dist}
\end{figure}

\begin{table}[!htbp]
\centering
\small
\caption{Rendered depth robustness analysis on YCB-V~\cite{ycbv} and T-LESS~\cite{tless}. We report REL between depths rendered from the \textbf{ground truth} poses and \textbf{perturbed} poses at three pose noise levels. Values are shown as mean / median.}
\label{tab:pose_depth_sensitivity}
\setlength{\tabcolsep}{6pt}
\begin{tabular}{lccc}
\toprule
Dataset & $1^\circ$, 1mm & $2^\circ$, 2mm & $5^\circ$, 5mm \\
\midrule
YCB-V  & 0.0111 / 0.0100 & 0.0190 / 0.0170  & 0.0356 / 0.0314 \\
T-LESS & 0.0136 / 0.0119  & 0.0219 / 0.0191  & 0.0385 / 0.0326 \\
\bottomrule
\end{tabular}
\end{table}

\noindent\textbf{Dataset Selection and Preparation.}
We select datasets that (i) include sensor depth (mostly real-captured; GN-Trans is fully synthetic), (ii) provide high-quality GT metric depth, (iii) cover standard robotic manipulation scenes, and (iv) span both household objects with diverse materials and industrial objects underrepresented in prior work~\cite{dai2022domain,d3roma}.
For YCB-V~\cite{ycbv} and T-LESS~\cite{tless}, which do not supply per-pixel GT metric depth, we render object-region metric depth from the provided 6D poses and CAD models using the camera intrinsics, taking the z-buffer depth (depth along the camera optical axis) of visible surfaces, and evaluate methods on those object pixels.
For HouseCat6D~\cite{housecat6d}, we use the provided object-region GT metric depth.
Fig.~\ref{fig:mat_dist} visualizes the material coverage, showing that our benchmark spans a wide diversity of rigid-body object materials.

\noindent\textbf{Object Transfer Audit.}
We further audit the object-level transfer factors using the training and testing datasets.
All source and target objects are mapped to a unified canonical category taxonomy.
The source inventory is the union of DREDS-CatKnown, ClearPose, and HISS. Canonical categories are assigned from dataset metadata and CAD/asset names, with only unambiguous synonyms merged; optical material regimes are taken from metadata or rendering labels when available and are otherwise assigned by inspecting the object assets and representative RGB observations. Ambiguous cases are conservatively treated as unmatched.
Table~\ref{tab:category_inventory} lists the canonical categories covered by each training and testing dataset.
This taxonomy is used only for the transfer audit and does not affect model training or evaluation.

We report verified exact-instance overlap, category coverage, material coverage, and category--material coverage.
Exact-instance overlap checks whether a target object shares the same CAD/asset identity with the training inventory.
Category coverage counts target objects whose canonical category appears in the training mixture.
Material coverage checks whether the target object's optical material regime is represented in the training mixture, where we use three regimes: diffuse, specular, and transparent.
Category--material coverage is stricter: a target object is counted only if its category appears in training and the same category has also been observed with the target object's material regime.

Table~\ref{tab:object_transfer_audit} shows that the zero-shot benchmark contains multiple transfer regimes rather than one homogeneous shift.
No target instance is shared with the source inventories.
T-LESS, XYZ-IBD, and ROBI have zero category coverage, indicating strong object-category transfer.
TransCG has the largest category coverage, $33/46$ objects (71.7\%), and also relatively high category--material coverage, $29/46$ objects (63.0\%); thus it mainly tests real transparent/specular-object sensing rather than category novelty alone.
HAMMER, HouseCat6D, GN-Trans, and YCB-V lie between these cases, with partial category coverage and lower strict category--material coverage.
The three material regimes are globally covered by the source mixture, so the remaining material-related difficulty mainly comes from unseen category--material combinations and their interaction with new sensors.

\begin{table*}[!tbp]
\centering
\scriptsize
\caption{Canonical category inventory of the training and zero-shot testing datasets.}
\label{tab:category_inventory}
\setlength{\tabcolsep}{4pt}
\resizebox{\textwidth}{!}{

\setlength{\tabcolsep}{3pt}
\renewcommand{\arraystretch}{1.15}
\setlength{\dashlinedash}{0.6pt}
\setlength{\dashlinegap}{2.0pt}

\begin{tabular}{@{}
>{\centering\arraybackslash}m{0.08\linewidth}
>{\raggedright\arraybackslash}m{0.16\linewidth}
>{\centering\arraybackslash}m{0.06\linewidth}
>{\raggedright\arraybackslash}m{0.66\linewidth}
@{}}
\toprule
\textbf{Split} & \textbf{Dataset} & \textbf{\#Cat.} & \textbf{Canonical Categories} \\
\midrule
Train & DREDS-CatKnown & 15 &
airplane, bag, bottle, bowl, camera, can, car, clock, faucet, guitar,
hat, knife, lamp, mug, phone \\
\hdashline
Train & HISS & 11 &
airplane, bag, bottle, bowl, camera, can, car, guitar, hat, mug, phone \\
\hdashline
Train & ClearPose & 20 &
beaker, bottle, bowl, container, cup, drinking glass, dropper, flask, fork, funnel,
graduated cylinder, knife, mug, pan, pitcher, plate, reagent bottle, rod, spoon, syringe \\
\midrule
Test & HAMMER & 12 &
bottle, box, can, cup, jar, knife, remote control, shoe, spoon, teapot, trophy, tube \\
\hdashline
Test & HouseCat6D & 12 &
bottle, box, can, cup, drinking glass, fork, knife, remote control, shoe, spoon, teapot, tube \\
\hdashline
Test & TransCG & 10 &
bottle, bowl, can, cup, drinking glass, fruit dish, jelly container,
soap container, spoon, wash container \\
\hdashline
Test & \rui{GN-Trans} & 33 &
ball, bottle, bowl, box, can, clamp, climbing hold, computer mouse, cup, drill, figurine,
fruit, gearbox, ice cube tray, industrial holder, industrial part, knife, lid cover, mount, mug,
nozzle, padlock, pawn, peeler, pipe connector, puzzle cube, scissors, screwdriver,
soap bar, toy airplane, tube, turbine housing, vase \\
\hdashline
Test & YCB-V & 12 &
block, bottle, bowl, box, can, clamp, drill, fruit, marker, mug, pitcher, scissors \\
\hdashline
Test & T-LESS & 5 &
electrical bracket, electrical connector, electrical enclosure, electrical socket,
electrical terminal block \\
\hdashline
Test & XYZ-IBD & 10 &
ball joint, bearing ring, bracket, eye bolt, flange, gear, link plate,
threaded rod, tube fitting, wrench \\
\hdashline
Test & ROBI & 6 &
bracket, eye bolt, gear, industrial connector, screw, tube fitting \\
\bottomrule
\end{tabular}
}
\end{table*}

\begin{table}[!htbp]
\centering
\scriptsize
\caption{Object-level transfer audit for zero-shot evaluation. Category coverage counts target objects whose canonical category appears in the training mixture. Category--material coverage additionally requires the target object's material regime to be represented for the same category in the training mixture.}
\label{tab:object_transfer_audit}
\setlength{\tabcolsep}{4pt}
\resizebox{\linewidth}{!}{

\begin{tabular}{lccccc}
\toprule
Dataset &
\#Obj. &
Exact Overlap &
Category Cov. &
Material Cov. &
Category--Material Cov. \\
\midrule
HAMMER         & 20 & 0/20 & 7/20 (35.0\%)  & 20/20 (100.0\%) & 4/20 (20.0\%)  \\
HouseCat6D     & 50 & 0/50 & 25/50 (50.0\%) & 50/50 (100.0\%) & 17/50 (34.0\%) \\
TransCG        & 46 & 0/46 & 33/46 (71.7\%) & 46/46 (100.0\%)   & 29/46 (63.0\%) \\
\rui{GN-Trans} & 76 & 0/76 & 15/76 (19.7\%) & 76/76 (100.0\%) & 10/76 (13.2\%) \\
YCB-V          & 21 & 0/21 & 9/21 (42.9\%)  & 21/21 (100.0\%) & 8/21 (38.1\%)  \\
T-LESS         & 30 & 0/30 & 0/30 (0.0\%)   & 30/30 (100.0\%) & 0/30 (0.0\%)   \\
XYZ-IBD        & 15 & 0/15 & 0/15 (0.0\%)   & 15/15 (100.0\%) & 0/15 (0.0\%)   \\
ROBI           & 7  & 0/7  & 0/7 (0.0\%)    & 7/7 (100.0\%)   & 0/7 (0.0\%)    \\
\bottomrule
\end{tabular}
}
\end{table}

\noindent\textbf{Sensor-Depth Transfer Audit.}
We also audit sensor-depth transfer on object pixels.
Each target dataset is compared with the DREDS-CatKnown/ClearPose/HISS training mixture.
The source reference and each target distribution are constructed using fixed weighting rules that are held constant across all comparisons. A sensor-valid object pixel has valid GT depth and $D_{\mathrm{sen}}>0$. For the valid-depth analysis, we use the log residual $e=|\log D_{\mathrm{sen}}-\log D^\ast|$ and compare source and target residual distributions within the same fixed GT-depth bins before averaging across nonempty bins. For hole morphology, we extract fixed-size local patches from the binary sensor-validity mask, vectorize the patches, and use the same patch extraction and random projection set for every dataset. The frame-level sensor signature concatenates missing-depth, boundary-missingness, valid-residual, and local-hole-morphology statistics; all descriptor preprocessing and RBF-kernel settings are fixed before target-domain comparison and shared across datasets.
We use four distribution-shift metrics.
First, \emph{object-invalid Wasserstein-1 distance} (Invalid-ratio W1) compares the per-frame distribution of invalid object-depth ratios using the Wasserstein-1 distance~\cite{peyre2019computational}.
It measures how much the overall amount of missing object depth shifts from source to target.
Second, \emph{depth-conditioned residual Wasserstein-1 distance} (Valid-error W1) compares the log-depth residual distribution on sensor-valid object pixels, conditioned on GT depth bins.
It measures whether the valid sensor returns have a different error pattern, rather than simply more holes.
Third, \emph{local-validity sliced Wasserstein distance} (Hole SWD) compares local valid/invalid patch descriptors using sliced Wasserstein distance~\cite{rabin2012wasserstein,peyre2019computational}.
It measures whether the spatial shape of missing-depth regions changes, e.g., isolated holes versus large connected missing regions.
Fourth, \emph{sensor-signature maximum mean discrepancy} (Sensor MMD) compares per-frame sensor-corruption descriptors with an RBF-kernel MMD~\cite{gretton2012kernel}.
It summarizes the overall sensor-depth corruption shift, including missing-depth ratio, boundary missingness, valid-depth residuals, and hole morphology.

As shown in Table~\ref{tab:sensor_depth_transfer_audit}, ROBI has the strongest sensor-depth shift, with object invalid depth increasing from 12.1\% in the source mixture to 46.8\%.
TransCG and GN-Trans mainly stress missing-depth transfer, with object invalid ratios of 26.8\% and 22.9\%, respectively.
HAMMER-D435 shows a moderate but clear boundary/missingness shift: its boundary invalid ratio increases from 11.2\% to 23.3\%, while its valid-depth residual shift is comparatively small.
HouseCat6D is close to the source mixture in sensor-depth statistics, suggesting that its transfer difficulty is less dominated by sensor-depth corruption.
YCB-V and T-LESS are not missingness-dominated, but still differ in boundary behavior, valid-depth residuals, or local validity structure.
The object and sensor-depth audits quantify category/material overlap and depth-corruption shift. Capture setup---including camera viewpoint, illumination, background, clutter, and the acquisition pipeline---is characterized from dataset metadata as complementary qualitative context rather than collapsed into a separate scalar distance. We therefore do not claim an independent quantitative audit of capture setup.

\begin{table*}[!tbp]
\centering
\scriptsize
\caption{Sensor-depth transfer audit on object pixels. All metrics are computed within object pixels.}
\label{tab:sensor_depth_transfer_audit}
\setlength{\tabcolsep}{5pt}
\begin{adjustbox}{max width=\textwidth}

\begin{tabular}{lcccccc}
\toprule
Dataset &
Object Invalid &
Boundary Invalid &
Invalid W1 &
Valid-Error W1 &
Hole SWD &
Sensor MMD \\
\midrule
Source Mix      & 12.1\% & 11.2\% & --    & --    & --    & --    \\
HAMMER          & 18.4\% & 23.3\% & 0.085 & 0.036 & 0.037 & 0.537 \\
HouseCat6D      & 11.1\% & 14.3\% & 0.040 & 0.041 & 0.009 & 0.473 \\
TransCG         & 26.8\% & 25.1\% & 0.150 & 0.052 & 0.069 & 0.692 \\
GN-Trans  & 22.9\% & 24.1\% & 0.109 & 0.045 & 0.035 & 0.535 \\
YCB-V           & 13.3\% & 31.2\% & 0.044 & 0.054 & 0.032 & 0.512 \\
T-LESS          & 3.1\%  & 6.3\%  & 0.091 & 0.057 & 0.078 & 0.466 \\
XYZ-IBD         & 19.8\% & 20.8\% & 0.078 & 0.059 & 0.057 & 0.579 \\
ROBI            & 46.8\% & 47.2\% & 0.347 & 0.092 & 0.134 & 0.780 \\
\bottomrule
\end{tabular}
\end{adjustbox}
\end{table*}

\noindent\textbf{Robustness Analysis of Rendered Depth.}
To assess how sensitive this rendered-depth proxy is to pose inaccuracies, we conduct a robustness analysis on YCB-V and T-LESS by perturbing the provided 6D poses at three noise levels and re-rendering z-buffer depth.
Note that we do not attempt to estimate the true pose-labeling error. This study instead serves as a sensitivity analysis that quantifies evaluation noise induced by pose perturbations and verifies that our conclusions are not driven by small rendering inaccuracies.
As summarized in Table~\ref{tab:pose_depth_sensitivity}, the induced discrepancy between the original and perturbed renders remains small under plausible noise (e.g., mean REL $\approx$1.1--1.4\% at $1^\circ,1\text{mm}$ and $\approx$1.9--2.2\% at $2^\circ,2\text{mm}$), suggesting that the pose-rendered object-region depth is reasonably stable for evaluation on these datasets.
If we conservatively treat $2^\circ,2\text{mm}$ as a representative pose-labeling error, then improvements below this REL level on YCB-V/T-LESS are within the rendering-induced error band and should be interpreted as comparable performance rather than definitive gains.
Accordingly, we avoid over-interpreting small metric differences on these pose-rendered benchmarks and focus on consistent trends across datasets with sensor-provided dense GT depth.

\noindent\textbf{GraspNet-Trans.}
\rui{We construct GraspNet-Trans (GN-Trans) as a fully synthetic benchmark derived from the official test split of GraspNet-1Billion~\cite{graspnet1b}. We reuse its CAD models, 6D object poses, and scene layouts while modifying object materials to introduce transparent and specular surfaces. Following~\cite{zhang2023close}, we render photorealistic RGB images, dense GT metric depth, and pattern-projected stereo infrared image pairs. Stereo matching is then applied to the rendered infrared images to simulate sensor depth with missing and noisy measurements.

We match the virtual camera intrinsics and extrinsics to the original captures, preserving the object-camera geometry. This allows the original GraspNet-1B grasp annotations to be transferred to the rendered scenes. GN-Trans therefore supports both depth restoration evaluation and downstream 6D grasp synthesis using point clouds reconstructed from restored depth.}

\noindent\textbf{HouseCat6D.}
A real household dataset with diverse everyday objects in natural indoor scenes. It includes ceramics, plastics, metals, and glass under realistic clutter and occlusion, and uses no artificial markers.

\noindent\textbf{TransCG.}
A tabletop dataset centered on transparent and glossy objects. It features refractive containers and specular surfaces among opaque distractors and changing backgrounds, emphasizing transparency-induced distortions.

\noindent\textbf{YCB-V.}
A widely used real benchmark of household items (tools, food, containers). It is dominated by diffuse plastics and paper packaging with some metallic objects, and contains few transparent or highly specular instances.

\noindent\textbf{T-LESS.}
An industrial-style dataset of texture-less, often symmetric components with similar shapes and part–whole relations, reflecting cluttered scenarios with low-texture parts.

\noindent\textbf{XYZ-IBD.}
A real industrial bin-picking dataset of metallic components with complex geometry, heavy clutter, and millimeter-level annotations, highlighting specular metal surfaces common in automated manufacturing.

\noindent\textbf{ROBI.}
To mitigate the limited number of XYZ-IBD samples, we additionally evaluate on ROBI, which features densely packed, highly reflective industrial parts and closely mirrors real-world industrial bin-picking scenarios.

\subsection{Metrics}
\label{app:metrics}
\noindent\textbf{Metric Depth Evaluation.}
Following prior work~\cite{dai2022domain,d3roma}, we report MAE, RMSE, REL, and threshold accuracy $\delta$. Let $d$ and $d^*$ denote predicted and GT depth at a valid pixel, respectively. We compute
\begin{align}
\text{MAE} &= \frac{1}{|V|}\sum_{(x,y)\in V} |d - d^*|, \\
\text{RMSE} &= \sqrt{\frac{1}{|V|}\sum_{(x,y)\in V} (d - d^*)^2}, \\
\text{REL} &= \frac{1}{|V|}\sum_{(x,y)\in V} \frac{|d - d^*|}{d^*}, \\
\delta_\tau &= \frac{1}{|V|}\sum_{(x,y)\in V} \mathds{1}\!\left(\max\!\left(\frac{d}{d^*},\frac{d^*}{d}\right) < \tau\right),
\end{align}
where $V$ is the evaluation pixel set (all pixels for datasets with full GT depth in a defined depth range, or object pixels for datasets with object-region GT depth), and $\tau \in \{1.05,1.10,1.25\}$.
\rui{We compute each metric separately for each image over its evaluation pixels $V$ and give equal weight to each image with a nonempty $V$ when computing the dataset mean.}

\noindent\textbf{Uncertainty Evaluation.}
We evaluate uncertainty quality using the sparsification protocol~\cite{poggi2020uncertainty,hornauer2025revisiting}. Given a per-pixel predicted uncertainty map $u(x,y)$ and an error definition $\epsilon(x,y)$ derived from an evaluation metric (REL, RMSE, or $\delta\!\ge\!1.05$), we rank pixels by $u(x,y)$ and iteratively remove the most uncertain pixels (2\% per step). After removing a fraction $\alpha$, we recompute the metric on the remaining pixels, producing an \emph{actual sparsification curve} $S_{\text{act}}(\alpha)$. We compare it against an \emph{oracle curve} $S_{\text{orc}}(\alpha)$ obtained by ranking pixels using the true error $\epsilon(x,y)$, and a \emph{random curve} $S_{\text{rnd}}(\alpha)$ obtained by random removal (averaged over trials). We then report
\begin{align}
\text{AUSE} &= \int_0^1 \bigl(S_{\text{act}}(\alpha) - S_{\text{orc}}(\alpha)\bigr)\, d\alpha, \\
\text{AURG} &= \int_0^1 \bigl(S_{\text{rnd}}(\alpha) - S_{\text{act}}(\alpha)\bigr)\, d\alpha,
\end{align}
where lower AUSE indicates closer-to-oracle uncertainty ordering, and higher AURG indicates larger gain over random removal.

\section{More Results}
\subsection{In-domain Depth Estimation Performance}
\label{app:in_domain_results}
\noindent\textbf{HAMMER.}
We report additional in-domain results on HAMMER~\cite{jung2023importance} using the RealSense D435 subset, where the valid depth range is cropped to \([0.2, 2.0]\) meters. 
We compare against two depth restoration methods~\cite{dai2022domain,d3roma}, three depth completion baselines (DepthPrompting~\cite{depthprompting}, OGNI-DC~\cite{zuo2024ogni}, and PromptDA~\cite{promptda}), as well as PPFT~\cite{ikemura2024robust}, the SOTA reported on HAMMER. 
For fair comparison, we follow training and evaluation setup of PPFT~\cite{ikemura2024robust}, using an input and evaluation resolution of \(272\times208\). 
We retrain all open-source baselines except PPFT, for which we use the official checkpoint under the same PPFT protocol.
In the main paper (Table~\ref{tab:hammer_results}), metrics are computed over all valid pixels. 
Here, we additionally report object area results in Table~\ref{tab:hammer_obj}, where evaluation is restricted to object pixels. On average, object-pixel metrics are typically worse than all-pixel metrics on HAMMER, since background regions tend to exhibit less sensor noise than object surfaces; nevertheless, \methodname\ maintains consistent advantages under this evaluation.

\begin{table}[!htbp]
\centering
\caption{Quantitative results on HAMMER~\cite{jung2023importance}. Metrics are computed over \textbf{object} pixels.}
\resizebox{\linewidth}{!}{

\begin{tabular}{lcccccc}
\toprule
Method & RMSE$\downarrow$ & REL$\downarrow$ & MAE$\downarrow$
& $\delta_{1.05}\uparrow$ & $\delta_{1.10}\uparrow$ & $\delta_{1.25}\uparrow$ \\
\midrule
SwinDR~\cite{dai2022domain} & 0.0503 & \underline{0.0403} & 0.0321 & 74.23 & 83.61 & 98.70 \\
$\text{D}^3\text{RoMa}$~\cite{d3roma} & 0.0592 & 0.0544 & 0.0348 & 68.43 & 84.95 & 96.81 \\
DP~\cite{depthprompting} & 0.0484 & 0.0445 & 0.0271 & 70.59 & \underline{87.95} & 98.64 \\
OGNI-DC~\cite{zuo2024ogni} & 0.0506 & 0.0515 & 0.0302 & 67.35 & 86.86 & 98.32 \\
PromptDA~\cite{promptda} & 0.0500 & 0.0532 & 0.0308 & 63.05 & 86.84 & 98.42 \\
PPFT~\cite{ikemura2024robust} & \underline{0.0389} & 0.0435 & \underline{0.0241} & \underline{78.36} & 87.76 & \underline{98.80} \\
\midrule
Ours 
& \msbest{0.0332}{0.0018}
& \msbest{0.0312}{0.0023}
& \msbest{0.0186}{0.0012}
& \msbest{82.95}{2.43}
& \msbest{94.98}{0.62}
& \msbest{99.36}{0.10} \\
\bottomrule
\end{tabular}
}
\label{tab:hammer_obj}
\end{table}

\noindent\textbf{DREDS and STD.}
Following the standard protocol~\cite{dai2022domain}, we clip GT depth to \([0.2, 2.0]\) meters and evaluate all methods at \(224\times126\) on object pixels. 
Results for NLSPN~\cite{nlspn}, LIDF~\cite{zhu2021rgb}, SwinDR~\cite{dai2022domain}, and ASGrasp~\cite{shi2024asgrasp} are taken from their respective papers. 
For $\text{D}^3$RoMa~\cite{d3roma}, the numbers reported in the original paper are computed over all valid pixels rather than object pixels\footnote{https://github.com/songlin/d3roma/issues/22}. 
To ensure a fair comparison, we retrain $\text{D}^3$RoMa using its official codebase under the same setting as \methodname: both are trained on the official DREDS-CatKnown training split, and the best checkpoint is selected using the DREDS-CatKnown validation split. 
We additionally report DREDS-CatNovel test results to evaluate generalization to unseen categories within the same synthetic domain. 
As shown in Table~\ref{tab:dreds_catnovel}, \methodname\ \rui{achieves the best REL, MAE, and $\delta_{1.05}$, while remaining competitive on RMSE and the other threshold metrics.}

\begin{table}[!htbp]
\centering
\caption{Quantitative results on DREDS-CatNovel~\cite{dai2022domain}. Metrics are computed over \textbf{object} pixels.}
\resizebox{\linewidth}{!}{

\begin{tabular}{L C C C C C C}
\toprule
Methods & RMSE$\downarrow$ & REL$\downarrow$ & MAE$\downarrow$
& $\delta_{1.05}\uparrow$ & $\delta_{1.10}\uparrow$ & $\delta_{1.25}\uparrow$ \\
\midrule
NLSPN~\cite{nlspn}             & 0.026 & 0.039 & 0.015 & 78.90 & 89.02 & 97.86 \\
LIDF~\cite{zhu2021rgb}         & 0.082 & 0.183 & 0.069 & 23.70 & 42.77 & 75.44 \\
SwinDR~\cite{dai2022domain}    & \textbf{0.022} & \underline{0.034} & \underline{0.013} & \underline{81.90} & \textbf{92.18} & \textbf{98.39} \\
$\text{D}^3$RoMa~\cite{d3roma} & 0.0247 & 0.0358 & 0.0141 & 79.56 & 89.94 & 98.29 \\

PromptDA~\cite{promptda}       & 0.0463 & 0.0983 & 0.0365 & 42.79 & 66.71 & 91.75 \\

\midrule
Ours                           
& \mssecond{0.0220}{0.0003}
& \msbest{0.0322}{0.0004}
& \msbest{0.0123}{0.0001}
& \msbest{82.30}{0.27}
& \mssecond{91.82}{0.12}
& \mssecond{98.37}{0.05} \\
\bottomrule
\end{tabular}
}
\label{tab:dreds_catnovel}
\end{table}

\subsection{Zero-shot Generalization Evaluation}
\label{app:zero_shot_results}
\noindent\textbf{Baselines.}
For zero-shot generalization, we compare \methodname\ with six representative methods spanning robotic RGB-D restoration and generalized depth completion.
SwinDR~\cite{dai2022domain}, D$^3$RoMa~\cite{d3roma}, and CDM~\cite{cdm} are restoration approaches tailored to commodity RGB-D sensing; CDM further learns camera-specific noise models to synthesize large-scale training data.
PromptDA~\cite{promptda} and PriorDA~\cite{priorda} perform prompt-based depth completion/super-resolution from sparse or low-resolution geometric cues, and OMNI-DC~\cite{omni-dc} addresses sparse-LiDAR depth completion across sparse-depth patterns such as LiDAR- and SfM-like inputs.
Although PromptDA, PriorDA, and OMNI-DC are not explicitly developed for manipulation benchmarks, they exhibit strong zero-shot performance on completion/super-resolution tasks.
Together with CDM, they are all built on DAV2~\cite{depthanythingv2}, so we treat them as DAV2-based reference methods in our study.
For OMNI-DC, we use their v1.1 implementation, which takes DAV2 relative depth as an additional input.
For CDM, we use the CDM-D435 model as a unified reference across all zero-shot experiments and report its performance in both Table~\ref{tab:zs_obj_results} and Table~\ref{tab:zs_all_results} for fair comparison.

\noindent\textbf{Evaluation protocol.}
For each method, we use the official checkpoints released by the authors without additional fine-tuning.
Depth values are clipped to \([0.2, 5.0]\) meters before computing metrics.
To ensure a controlled and fair comparison in the main zero-shot protocol, we standardize the \emph{nominal} RGB-D input resolution for all methods: RGB and the corresponding raw depth are resized to \(518\times518\).
All predictions are then resized to \(320\times180\) for evaluation to match the benchmark resolution.
For \methodname, this \(518\times518\) size denotes the nominal RGB-D input size; internally, \methodname\ resizes the input to \(560\times560\) to satisfy both the stride-16 constraint of the sensor-depth context network and the patch-size-14 constraint of DAV2~\cite{depthanythingv2}.
This internal resizing is also used in the latency analysis.

We further record the official/preferred resolution setting of each baseline for the resolution-sensitivity check.
When a paper specifies an inference resolution, we follow the paper setting; otherwise, we follow the released inference code.
Specifically, SwinDRNet is evaluated at \(224\times224\), D$^3$RoMa at \(360\times640\) with its official camera/intrinsics handling, OMNI-DC at \(480\times640\), and PriorDA at \(518\times518\).
CDM follows its released inference setting, which resizes the shorter side to \(518\) while preserving the aspect ratio.
PromptDA uses the original image resolution unless the longer side exceeds \(1008\), in which case it is resized to \(1008\).
In all cases, the final predictions are resized to the same \(320\times180\) grid for metric computation.
In the main paper, Table~\ref{tab:zs_obj_results} reports REL and $\delta_{1.05}$ computed over object pixels, which directly reflects performance on regions most relevant to manipulation.
For datasets that provide dense scene-level ground truth depth, namely HAMMER~\cite{jung2023importance}, TransCG~\cite{fang2022transcg}, \rui{GN-Trans}, XYZ-IBD~\cite{xyz-ibd}, and ROBI~\cite{robi}, we additionally report results over all valid pixels in Table~\ref{tab:zs_all_results}.

\begin{table*}[!tbp]
\centering
\caption{Zero-shot generalization results on \textbf{all valid} pixels. 
We report REL$\downarrow$ and $\delta_{1.05}\uparrow$ on five datasets. 
Bold and underlined entries indicate the best and second-best results, respectively.}
\resizebox{\textwidth}{!}{

\begin{tabular}{l*{10}{c}c}
\toprule
\multirow{2}{*}{Method}
& \multicolumn{2}{c}{HAMMER} & \multicolumn{2}{c}{TransCG} & \multicolumn{2}{c}{\rui{GN-Trans}}
& \multicolumn{2}{c}{XYZ-IBD} & \multicolumn{2}{c}{ROBI}
& \multirow{2}{*}{Mean Rank} \\
\cmidrule(lr){2-3} \cmidrule(lr){4-5} \cmidrule(lr){6-7} \cmidrule(lr){8-9} \cmidrule(lr){10-11}
& REL$\downarrow$ & $\delta_{1.05}\uparrow$
& REL$\downarrow$ & $\delta_{1.05}\uparrow$
& REL$\downarrow$ & $\delta_{1.05}\uparrow$
& REL$\downarrow$ & $\delta_{1.05}\uparrow$
& REL$\downarrow$ & $\delta_{1.05}\uparrow$ & \\
\midrule
SwinDR~\cite{dai2022domain}
& 0.0492 & 72.81
& 0.0138 & 94.26
& 0.0438 & \underline{85.96}
& 0.0283 & 84.01
& 0.0099 & 95.05
& 3.3 \\
$\text{D}^3\text{RoMa}$~\cite{d3roma}
& 0.0804 & 54.46
& 0.0330 & 81.70
& 0.0546 & 81.05
& 0.0664 & 50.28
& 0.0347 & 76.97
& 5.4 \\
PromptDA~\cite{promptda}
& 0.3642 & 33.42
& 0.1250 & 54.73
& 0.2083 & 56.18
& 0.2998 & 32.94
& 0.2000 & 67.97
& 7.0 \\
OMNI-DC~\cite{omni-dc}
& 0.0665 & 68.65
& \underline{0.0121} & \underline{95.54}
& 0.0752 & 81.35
& \textbf{0.0055} & \textbf{98.56}
& \underline{0.0057} & \underline{99.52}
& 2.8 \\
PriorDA~\cite{priorda}
& 0.0670 & 49.79
& 0.0371 & 74.70
& 0.0690 & 76.05
& 0.0312 & 82.69
& 0.0286 & 79.63
& 5.4 \\
CDM~\cite{cdm}
& \textbf{0.0278} & \textbf{87.09}
& 0.0159 & 94.24
& \underline{0.0365} & 85.52
& 0.0238 & 88.53
& 0.0076 & 99.00
& 2.7 \\
\midrule
Ours
& \mssecond{0.0412}{0.0043} & \mssecond{80.30}{0.62}
& \msbest{0.0088}{0.0003} & \msbest{96.68}{0.13}
& \msbest{0.0214}{0.0006} & \msbest{90.91}{0.11}
& \mssecond{0.0109}{0.0020} & \mssecond{95.69}{0.80}
& \msbest{0.0025}{0.0001} & \msbest{99.55}{0.05}
& \textbf{1.4} \\
\bottomrule
\end{tabular}
}
\label{tab:zs_all_results}
\end{table*}

\noindent\textbf{Results on All Valid Pixels.}
Table~\ref{tab:zs_all_results} summarizes zero-shot performance over all valid pixels on datasets. 
On HAMMER, CDM attains the lowest REL and the highest $\delta_{1.05}$, which is consistent with its explicit camera-noise modeling and large-scale synthetic pretraining tailored to the D435 sensor. 
On XYZ-IBD, OMNI-DC achieves the best REL and $\delta_{1.05}$, reflecting its strong capability on industrial bin-picking scenes with challenging sensor noise. 
For the remaining datasets, \methodname\ obtains the best performance on both REL and $\delta_{1.05}$ and achieves the lowest overall mean rank across all methods. 
These observations are consistent with the object-pixel results in Table~\ref{tab:zs_obj_results} of the main paper, and the similar mean ranks across the two tables indicate that the same conclusions hold whether evaluation is restricted to object regions or carried out over all valid pixels.

\begin{table}[!htbp]
\centering
\caption{Comparison of CDM variants and \methodname\ on \textbf{object} pixels for datasets without dedicated CDM models.}
\begin{tabular}{l l c c c}
\toprule
Dataset & Method & RMSE$\downarrow$ & REL$\downarrow$ & $\delta_{1.05}\uparrow$ \\
\midrule
\multirow{4}{*}{XYZ-IBD}
  & CDM-D435   & 0.0128 & 0.0145 & 96.60 \\
  & CDM-L515   & 0.0146 & 0.0172 & 96.47 \\
  & CDM-Kinect & 0.0163 & 0.0182 & 93.73 \\
  & Ours       & \textbf{0.0067} & \textbf{0.0063} & \textbf{99.25} \\
\midrule
\multirow{4}{*}{YCB-V}
  & CDM-D435   & 0.1216 & 0.0575 & 67.17 \\
  & CDM-L515   & 0.1202 & 0.0501 & 75.19 \\
  & CDM-Kinect & 0.1240 & 0.0566 & 68.35 \\
  & Ours       & \textbf{0.1108} & \textbf{0.0259} & \textbf{94.35} \\
\midrule
\multirow{4}{*}{T-LESS}
  & CDM-D435   & 0.0487 & 0.0250 & 88.92 \\
  & CDM-L515   & 0.0474 & 0.0250 & 92.19 \\
  & CDM-Kinect & 0.0473 & 0.0252 & 89.94 \\
  & Ours       & \textbf{0.0392} & \textbf{0.0143} & \textbf{95.83} \\
\midrule
\multirow{4}{*}{ROBI}
  & CDM-D435   & 0.0074 & 0.0146 & 98.04 \\
  & CDM-L515   & 0.0076 & 0.0149 & 97.47 \\
  & CDM-Kinect & 0.0087 & 0.0175 & 96.00 \\
  & Ours       & \textbf{0.0054} & \textbf{0.0096} & \textbf{99.12} \\
\bottomrule
\end{tabular}
\label{tab:zs_cdm_results}
\end{table}

\noindent\textbf{Compared with Different CDM Models.}
For YCB-V~\cite{ycbv}, T-LESS~\cite{tless}, XYZ-IBD~\cite{xyz-ibd}, and ROBI~\cite{robi}, the cameras used in data acquisition are not covered by the camera models provided in CDM. 
To probe how CDM behaves under such cross-camera settings, we evaluate CDM-D435, CDM-L515, and CDM-Kinect on these datasets over object pixels, and report the results for reference in Table~\ref{tab:zs_cdm_results}. 
\rui{Across all four datasets, the strongest CDM variant varies by dataset and metric.} 
In contrast, \methodname\ consistently achieves lower errors and higher $\delta_{1.05}$, indicating that its gains remain clear even when comparing against the best-performing CDM camera model.

\begin{table}[!htbp]
\centering
\caption{Cross-sensor results on HAMMER. Metrics are computed over \textbf{object} pixels.}
\label{tab:hammer_cross_sensor}
\resizebox{\linewidth}{!}{
\begin{tabular}{llccc}
\toprule
Sensor & Method & RMSE$\downarrow$ & REL$\downarrow$ & $\delta_{1.05}$ $\uparrow$ \\
\midrule
\multirow{3}{*}{D435 (IR Stereo)}
& OMNI-DC       & 0.0616 & 0.0586 & 67.47 \\
& CDM-D435      & 0.0497 & 0.0427 & 77.95 \\
& Ours & \textbf{0.0338} & \textbf{0.0324} & \textbf{82.96} \\
\midrule
\multirow{3}{*}{L515 (D-ToF)}
& OMNI-DC       & 0.0661 & 0.0655 & 66.61 \\
& CDM-L515      & 0.0438 & 0.0407 & 77.48 \\
& Ours & \textbf{0.0387} & \textbf{0.0320} & \textbf{83.83} \\
\midrule
\multirow{4}{*}{Helios (I-ToF)}
& OMNI-DC       & 0.1302 & 0.1332 & 37.50 \\
& CDM-L515      & \textbf{0.0496} & \textbf{0.0549} & 62.25 \\
& CDM-D435      & 0.0843 & 0.0991 & 46.78 \\
& Ours & 0.0567 & 0.0637 & \textbf{62.86} \\
\bottomrule
\end{tabular}
}
\end{table}

\noindent\textbf{Results on Sensor-agnostic Generalization.}
Table~\ref{tab:hammer_cross_sensor} reports HAMMER results on object pixels across three depth sensors (D435, L515, and Helios), evaluating each method without sensor-specific fine-tuning.
Our method achieves the best overall performance on D435 and L515, obtaining the lowest RMSE/REL and the highest $\delta_{1.05}$ (82.96\% and 83.83\%, respectively), which indicates strong generalization across stereo and D-ToF sensing.
On the more challenging Helios I-ToF setting, our method remains competitive: it achieves the highest $\delta_{1.05}$ and substantially improves over OMNI-DC, while \rui{CDM-L515 attains lower RMSE/REL}.
Overall, this preliminary evaluation suggests that our approach can handle unseen sensor-captured depth to a certain extent.
In future work, we plan to scale up pretraining with larger and more diverse data to further strengthen sensor-agnostic generalization.

\begin{figure}[t]
\centering
\subfloat{
\includegraphics[width=\linewidth]{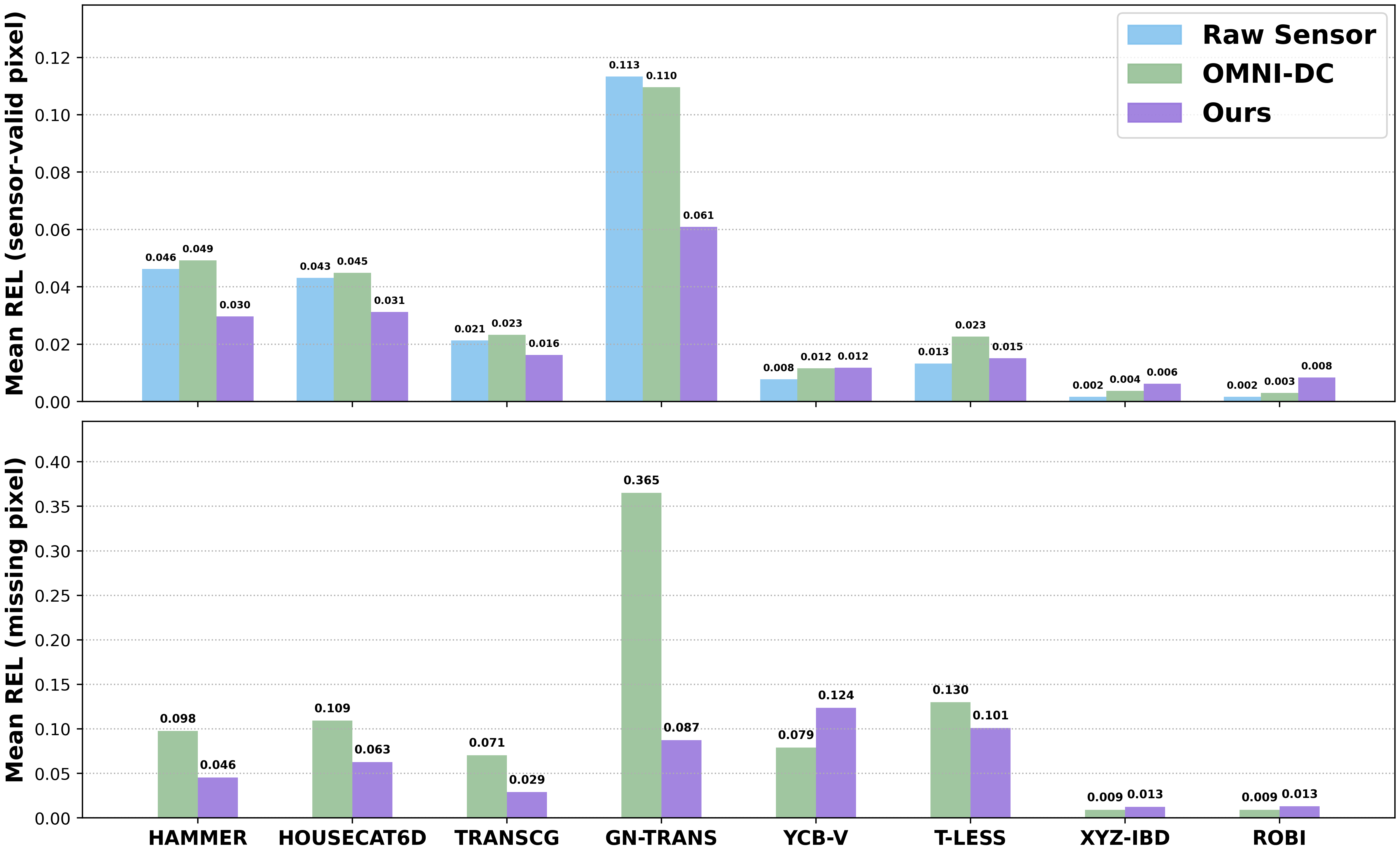}
}
\caption{
Comparison between \methodname\ and OMNI-DC on object regions across eight datasets, reporting REL error separately on \emph{sensor-valid pixels} (pixels with reported depth measurements) and \emph{missing pixels} (pixels where the sensor fails to return depth).
}
\label{fig:vs_omnidc}
\end{figure}

\noindent\textbf{Results under Different Resolution.} 
Because we observe that the performance of D$^3$RoMa varies noticeably with input resolution, we additionally evaluate it at its official setting of 360$\times$640 to quantify its resolution sensitivity, while keeping the evaluation resolution fixed at 320$\times$180 for all methods. As shown in Table~\ref{tab:zs_res_obj}, \rui{the effect of input resizing varies by method and dataset, with improvements in some settings and degradations in others.} \rui{D$^3$RoMa also exhibits dataset-dependent changes, suggesting sensitivity to camera intrinsics and input resizing.} 
We further report a complete official-resolution protocol in Table~\ref{tab:zs_obj_results_official_resolution}, where each baseline is evaluated using its official or preferred inference setting while the final metric computation remains fixed at the same $320\times180$ evaluation grid. This protocol leads to the same overall conclusion as the standardized $518\times518$ protocol in Table~VI. \methodname\ remains the best overall method, with mean rank changing only from $1.4$ to $1.5$. These results indicate that although several baselines are sensitive to input resolution and camera-intrinsics handling, the relative advantage of \methodname\ is not an artifact of the standardized input protocol.

\begin{table*}[!tbp]
\centering
\caption{Effect of input resolution for SwinDR, D$^3$RoMa, and \methodname\ on zero-shot benchmarks on \textbf{object} pixels. 
We report REL$\downarrow$ and $\delta_{1.05}\uparrow$ over object pixels.
All methods are evaluated at 320$\times$180; only the input resizing differs.}
\resizebox{\textwidth}{!}{
\begin{tabular}{l*{17}{c}c}
\toprule
\multirow{2}{*}{Method} &  \multirow{2}{*}{Input Res.}
& \multicolumn{2}{c}{HAMMER} & \multicolumn{2}{c}{HouseCat6D} & \multicolumn{2}{c}{TransCG} & \multicolumn{2}{c}{\rui{GN-Trans}}
& \multicolumn{2}{c}{YCB-V} & \multicolumn{2}{c}{T-LESS} & \multicolumn{2}{c}{XYZ-IBD} & \multicolumn{2}{c}{ROBI}  \\
\cmidrule(lr){3-4} \cmidrule(lr){5-6} \cmidrule(lr){7-8} \cmidrule(lr){9-10}
\cmidrule(lr){11-12} \cmidrule(lr){13-14} \cmidrule(lr){15-16} \cmidrule(lr){17-18} 
& & REL$\downarrow$ & $\delta_{1.05}\uparrow$
& REL$\downarrow$ & $\delta_{1.05}\uparrow$
& REL$\downarrow$ & $\delta_{1.05}\uparrow$
& REL$\downarrow$ & $\delta_{1.05}\uparrow$
& REL$\downarrow$ & $\delta_{1.05}\uparrow$
& REL$\downarrow$ & $\delta_{1.05}\uparrow$
& REL$\downarrow$ & $\delta_{1.05}\uparrow$
& REL$\downarrow$ & $\delta_{1.05}\uparrow$ \\
\midrule
SwinDR &  518$\times$518   & 0.0429 & 76.35 & 0.0463 & 76.84 & 0.0307 & 81.10 & 0.1078 & 61.75 & \textbf{0.0241} & 88.82 & 0.0151 & 93.53 & 0.0247 & 85.19 & 0.0318 & 77.95 \\
D$^3$RoMa  & 518$\times$518 & 0.0655 & 58.43 & 0.0498 & 76.45 & 0.0419 & 67.56 & 0.1231 & 50.27 & 0.0494 & 81.61 & 0.0511 & 73.36 & 0.0650 & 47.27 & 0.0575 & 50.07 \\
Ours  & 518$\times$518      & \textbf{0.0324} & \textbf{82.96} & \textbf{0.0345} & \textbf{86.09} & \textbf{0.0193} & \textbf{89.66} & \textbf{0.0677} & \textbf{67.20} & 0.0259 & \textbf{94.35} & \textbf{0.0143} & \textbf{95.83} & \textbf{0.0063} & \textbf{99.25} & \textbf{0.0096} & \textbf{99.12} \\
\midrule
SwinDR  & 360$\times$640   & 0.0394 & 77.85 & 0.0441 & 76.90 & 0.0242 & 83.30 & 0.1017 & 62.09 & \textbf{0.0169} & 90.62 & \textbf{0.0134} & 93.87 & 0.0257 & 82.34 & 0.0393 & 66.71 \\
D$^3$RoMa  & 360$\times$640 & 0.0393 & \rui{75.18} & 0.0414 & 79.59 & 0.0505 & 58.26 & 0.0744 & 60.34 & 0.0625 & 73.53 & 0.0453 & 77.02 & 0.0869 & 28.59 & 0.0895 & 24.07 \\
Ours  & 360$\times$640 & \textbf{0.0332} & \textbf{82.20} & \textbf{0.0342} & \textbf{86.37} & \textbf{0.0193} & \textbf{89.71} & \textbf{0.0677} & \textbf{66.70} & 0.0265 & \textbf{94.25} & 0.0153 & \textbf{95.56} & \textbf{0.0080} & \textbf{98.33} & \textbf{0.0100} & \textbf{99.04} \\
\bottomrule
\end{tabular}
}
\label{tab:zs_res_obj}
\end{table*}

\begin{table*}[!tbp]
\centering
\small
\caption{
Official-resolution zero-shot generalization results on \textbf{object} pixels.
For each baseline, inference follows the official/preferred input-resolution setting.
All predictions are evaluated at $320\times180$.
$\Delta$Rank denotes the change in mean rank relative to the standardized-resolution protocol in main paper.}
\label{tab:zs_obj_results_official_resolution}
\resizebox{\textwidth}{!}{

\begin{tabular}{l*{16}{c}cc}
\toprule
\multirow{2}{*}{Method}
& \multicolumn{2}{c}{HAMMER}
& \multicolumn{2}{c}{HouseCat6D}
& \multicolumn{2}{c}{TransCG}
& \multicolumn{2}{c}{\rui{GN-Trans}}
& \multicolumn{2}{c}{YCB-V}
& \multicolumn{2}{c}{T-LESS}
& \multicolumn{2}{c}{XYZ-IBD}
& \multicolumn{2}{c}{ROBI}
& \multirow{2}{*}{Mean Rank}
& \multirow{2}{*}{$\Delta$Rank} \\
\cmidrule(lr){2-3}
\cmidrule(lr){4-5}
\cmidrule(lr){6-7}
\cmidrule(lr){8-9}
\cmidrule(lr){10-11}
\cmidrule(lr){12-13}
\cmidrule(lr){14-15}
\cmidrule(lr){16-17}
& REL$\downarrow$ & $\delta_{1.05}\uparrow$
& REL$\downarrow$ & $\delta_{1.05}\uparrow$
& REL$\downarrow$ & $\delta_{1.05}\uparrow$
& REL$\downarrow$ & $\delta_{1.05}\uparrow$
& REL$\downarrow$ & $\delta_{1.05}\uparrow$
& REL$\downarrow$ & $\delta_{1.05}\uparrow$
& REL$\downarrow$ & $\delta_{1.05}\uparrow$
& REL$\downarrow$ & $\delta_{1.05}\uparrow$
& & \\
\midrule
SwinDR
& 0.0401 & 75.53 & 0.0503 & 74.77 & \underline{0.0200} & \textbf{90.41} & 0.0815 & \underline{64.81} & \textbf{0.0182} & 91.41 & \underline{0.0175} & \underline{92.35} & 0.0255 & 84.72 & 0.0346 & 74.21
& 3.1 & -0.06 \\
$\text{D}^3\text{RoMa}$
& \underline{0.0393} & 75.18 & 0.0414 & \underline{79.59} & 0.0505 & 58.26 & \underline{0.0744} & 60.34 & 0.0625 & 73.53 & 0.0453 & 77.02 & 0.0869 & 28.59 & 0.0895 & 24.07
& 4.8 & -0.50 \\
PromptDA
& 0.5383 & 6.70 & 0.3792 & 20.16 & 0.2960 & 14.69 & 0.3474 & 10.99 & 0.4364 & 14.96 & 0.1390 & 59.47 & 0.3373 & 33.47 & 0.7275 & 6.27
& 6.9 & -0.06 \\
OMNI-DC
& 0.0585 & 67.84 & 0.0523 & 72.14 & 0.0356 & 79.14 & 0.1640 & 51.88 & \underline{0.0214} & \textbf{94.59} & 0.0253 & 91.53 & \textbf{0.0047} & \textbf{99.03} & \textbf{0.0065} & \textbf{99.34}
& 3.3 & +0.25 \\
PriorDA
& 0.0762 & 41.41 & 0.0720 & 45.51 & 0.0418 & 71.36 & 0.1492 & 46.55 & 0.0338 & 87.47 & 0.0420 & 73.41 & 0.0205 & 95.26 & 0.0182 & 95.15
& 5.0 & +0.06 \\
CDM
& 0.0425 & \underline{78.41} & \underline{0.0407} & 77.33 & 0.0251 & 87.18 & 0.0914 & 60.81 & 0.0545 & 69.68 & 0.0244 & 89.84 & 0.0139 & 97.56 & 0.0137 & 98.20
& 3.4 & +0.25 \\
\midrule
Ours
& \msbest{0.0325}{0.0002} & \msbest{82.99}{0.71}
& \msbest{0.0347}{0.0014} & \msbest{86.50}{0.39}
& \msbest{0.0198}{0.0004} & \mssecond{89.47}{0.22}
& \msbest{0.0687}{0.0012} & \msbest{67.84}{1.04}
& \makecell[c]{0.0288\\[-0.6ex]{\scriptsize$\pm0.0047$}} & \mssecond{93.73}{0.76}
& \msbest{0.0143}{0.0007} & \msbest{95.73}{0.10}
& \mssecond{0.0070}{0.0009} & \mssecond{98.76}{0.46}
& \mssecond{0.0098}{0.0005} & \mssecond{99.05}{0.28}
& \textbf{1.5} & +0.06 \\
\bottomrule
\end{tabular}
}
\end{table*}

\noindent\textbf{\methodname\ v.s. OMNI-DC.}
OMNI-DC is designed to leverage depth inputs that are either noise-free or subject to relatively low, unstructured noise. While it demonstrates robustness to mild Gaussian perturbations, it is more sensitive to the structured and material-dependent noise introduced by non-Lambertian surfaces, which commonly appears in commodity RGB-D sensing. 

To better understand this behavior, we conduct a pixel-regime analysis on eight datasets by separately reporting the REL error on \emph{sensor-valid pixels} (pixels with depth measurements) and \emph{missing pixels} (pixels where the sensor fails to return depth) within object regions. As shown in Fig.~\ref{fig:vs_omnidc}, on datasets with pronounced non-Lambertian surface (HAMMER, HouseCat6D, TransCG, \rui{GN-Trans}, and T-LESS), OMNI-DC exhibits limited correction on sensor-valid pixels, whereas \methodname\ consistently refines depth estimates in these regions. In contrast, on datasets with relatively low sensor noise (XYZ-IBD, ROBI, and YCB-V), \methodname\ may introduce additional refinement errors on sensor-valid pixels, while OMNI-DC preserves the original measurements more faithfully.

A similar trend is observed on missing pixels. OMNI-DC’s completion quality is also affected by noisy observations, but it performs more favorably on datasets where sensor noise is limited. Taken together, these results indicate that OMNI-DC is better suited to regimes with reliable depth measurements, whereas \methodname\ is more effective when depth observations are dense but corrupted by material-dependent noise.

\begin{table*}[!tbp]
\centering
\small
\caption{\rui{Uncertainty calibration on eight zero-shot datasets.}}
\label{tab:unc_calibration}
\rui{\resizebox{\textwidth}{!}{

\begin{tabular}{lccccccccc}
\toprule
\multirow{2}{*}{\textbf{Dataset}}
& \multirow{2}{*}{\textbf{AURG$\uparrow$}}
& \multirow{2}{*}{\textbf{Spearman$\uparrow$}}
& \multirow{2}{*}{\textbf{NLL-Raw$\downarrow$}}
& \multirow{2}{*}{\textbf{NLL-Temp$\downarrow$}}
& \multicolumn{5}{c}{\textbf{Coverage ECE$\downarrow$ by GT-depth range (m)}} \\
\cmidrule(lr){6-10}
& & & & 
& \textbf{0.2--0.4}
& \textbf{0.4--0.6}
& \textbf{0.6--0.8}
& \textbf{0.8--1.0}
& \textbf{1.0--2.0} \\
\midrule
GN-Trans   & 0.0229 & 0.674 & -0.688 & -0.501 & 0.288 & 0.075 & 0.071 & --    & --    \\
HAMMER     & 0.0108 & 0.431 & -0.850 & -1.670 & 0.039 & 0.191 & 0.324 & 0.397 & 0.290 \\
HouseCat6D & 0.0106 & 0.285 & -0.924 & -1.476 & 0.246 & 0.314 & 0.239 & 0.195 & --    \\
TransCG    & 0.0085 & 0.571 & -0.969 & -2.464 & 0.049 & 0.028 & 0.150 & --    & --    \\
ROBI       & 0.0026 & 0.471 & -0.917 & -2.925 & 0.125 & 0.129 & --    & --    & --    \\
T-LESS     & 0.0075 & 0.439 & -1.316 & -2.977 & --    & --    & 0.085 & 0.029 & --    \\
XYZ-IBD    & 0.0025 & 0.650 & -1.217 & -3.334 & --    & 0.193 & 0.162 & 0.027 & --    \\
YCB-V      & 0.0204 & 0.536 & -1.327 & -2.937 & --    & 0.147 & 0.161 & 0.126 & 0.094 \\
\bottomrule
\end{tabular}
}}
\end{table*}

\noindent\textbf{\rui{Uncertainty Calibration.}}
\rui{We evaluate uncertainty on eight zero-shot datasets using AURG and Spearman correlation for error ranking, NLL before (NLL-Raw) and after (NLL-Temp) temperature scaling, and ECE within GT depth bins. For each evaluated pixel $i$, let $\hat d_i$, $d_i$, and $u_i$ denote predicted depth, ground-truth depth, and raw positive uncertainty, respectively. We use the log-depth residual $e_i=\log\max(\hat d_i,\varepsilon)-\log\max(d_i,\varepsilon)$ with $\varepsilon=10^{-6}$ and clamp $u_i$ below by $\varepsilon$. Spearman correlation is computed between the ranks of $u_i$ and $|e_i|$. AURG measures the gain of uncertainty-based sparsification over random removal using REL (defined earlier in this Appendix); it is computed per image and averaged over evaluated images. Neither metric requires a calibrated uncertainty scale.}

\rui{For the NLL diagnostic, we use a Laplace likelihood for $e_i$ with scale $b_i(\tau)=\max(\tau u_i,\varepsilon)$:}
\begin{equation}
\rui{\mathrm{NLL}(\tau)=\frac{1}{N}\sum_{i=1}^{N}\left(\frac{|e_i|}{b_i(\tau)}+\log(2b_i(\tau))\right).}
\end{equation}
\rui{NLL-Raw uses $\tau=1$. For NLL-Temp, a single positive temperature is fitted by minimizing Laplace NLL over the source-domain validation pixels $\mathcal{S}$, giving $\tau^*=|\mathcal{S}|^{-1}\sum_{i\in\mathcal{S}}|e_i|/u_i$. The same $\tau^*$ is used on all eight zero-shot datasets, without fitting on target-domain data.}

\rui{The reported depth-bin ECE measures predictive interval coverage rather than error in meters. For a GT-depth bin $\mathcal{B}_b$ and nominal coverage level $p\in\mathcal{P}=\{0.50,0.68,0.90,0.95\}$, the Laplace interval half-width is $h_i(p)=-b_i(\tau^*)\log(1-p)$. We compute}
\begin{align}
\rui{\widehat c_b(p)} &= \rui{\frac{1}{|\mathcal{B}_b|}\sum_{i\in\mathcal{B}_b}\mathbb{I}\!\left(|e_i|\le h_i(p)\right)},\\
\rui{\mathrm{ECE}_b} &= \rui{\frac{1}{|\mathcal{P}|}\sum_{p\in\mathcal{P}}|\widehat c_b(p)-p|.}
\end{align}
\rui{The bins are defined by GT depth in meters, as shown in Table~\ref{tab:unc_calibration}; the ECE values themselves are dimensionless. A dash (--) indicates insufficient valid pixels in a bin.}

\rui{As shown in Table~\ref{tab:unc_calibration}, temperature scaling reduces NLL on seven datasets but increases it on GN-Trans from $-0.688$ to $-0.501$. ECE also varies across datasets and depth ranges. Thus, uncertainty can rank unreliable pixels, but its numerical scale is not consistently calibrated across datasets.}

\noindent\textbf{\rui{Uncertainty Thresholding.}}
\rui{We test whether uncertainty identifies pixels with more accurate depth. On GN-Trans and HouseCat6D, we retain $50\%$ of object pixels with the lowest uncertainty (Low-$\mathbf{U}$), the highest uncertainty (High-$\mathbf{U}$), or random selection. Table~\ref{tab:unc_threshold} reports REL, RMSE, and $1-\delta_{1.05}$, where the last metric is the percentage of selected pixels failing the $\delta_{1.05}$ accuracy threshold. Low-$\mathbf{U}$ gives lower errors than random selection on both datasets, while High-$\mathbf{U}$ gives higher errors. For example, REL decreases from 0.0673 to 0.0422 on GN-Trans and from 0.0338 to 0.0221 on HouseCat6D.}

\begin{table}[!htbp]
\centering
\small
\caption{\rui{Depth errors at $50\%$ pixel coverage.}}
\label{tab:unc_threshold}
\rui{\resizebox{\linewidth}{!}{\begin{tabular}{llccc}
\toprule
\textbf{Dataset} & \textbf{Selection} & \textbf{REL$\downarrow$} & \textbf{RMSE$\downarrow$} & $\mathbf{1-\delta_{1.05}\downarrow}$ \\
\midrule
\multirow{3}{*}{GN-Trans}
& Random      & 0.0673 & 0.0364 & 32.5 \\
& Low-$\mathbf{U}$  & \textbf{0.0422} & \textbf{0.0243} & \textbf{21.7} \\
& High-$\mathbf{U}$ & 0.0924 & 0.0432 & 43.3 \\
\midrule
\multirow{3}{*}{HouseCat6D}
& Random      & 0.0338 & 0.0375 & 13.5 \\
& Low-$\mathbf{U}$  & \textbf{0.0221} & \textbf{0.0146} & \textbf{3.5} \\
& High-$\mathbf{U}$ & 0.0454 & 0.0507 & 23.5 \\
\bottomrule
\end{tabular}}}
\end{table}

\subsection{Ablations of UCFC Feedback}
\label{app:ucfc_ablation}
We provide additional ablations to examine whether the improvement of UCFC comes from recurrent uncertainty feedback rather than from uncertainty supervision, additional parameters, residual-based reliability estimation, generic feature fusion, or single-step prediction.
All variants use the same DAV2 backbone, ADF initialization, training data, input resolution, and evaluation protocol.
For the recurrent variants, the recurrent refinement backbone is also kept fixed; for the zero-shot ablation reported in the main paper, Robust-SN is fixed across all variants.
The variants differ only in whether refinement is single-step or recurrent, and in how uncertainty or reliability information is injected into the refinement process.

Table~\ref{tab:ucfc_ablation_def} summarizes the design differences.
\emph{Single-step depth} directly predicts a one-shot depth update without recurrent refinement or uncertainty prediction.
\emph{Single-step depth+unc.} additionally predicts uncertainty with the same uncertainty supervision, but still does not use recurrent refinement or uncertainty feedback.
\emph{Plain refinement} removes both uncertainty supervision and uncertainty feedback while keeping the recurrent refinement backbone.
\emph{Uncertainty-only} predicts uncertainty with the same supervision but does not feed it back into refinement.
\emph{Static learned gate} uses a learned gate that is independent of the predicted uncertainty.
\emph{Validity-mask gate} uses the raw valid-depth mask as the reliability signal.
\emph{Residual gate} replaces the predicted uncertainty with the normalized absolute residual between the current depth estimate $\mathbf{D}^{k}$ and the original sensor depth $\mathbf{D}_{\mathrm{sen}}$ on valid sensor pixels, while invalid sensor pixels are set to the maximum residual value of 1 before gating.
\emph{Simple conv fusion} directly concatenates depth, uncertainty, RGB features, and sensor-depth features, and uses convolutional layers to predict the depth and uncertainty updates.
\emph{UCFC-PM} keeps the same calibration modules as UCFC but replaces the uncertainty map with a constant map, serving as a parameter-matched no-feedback control.
\emph{UCFC} uses the evolving predicted uncertainty as recurrent feedback to recalibrate RGB and sensor-depth context at each step.
\rui{
To separate the benefit of an informative spatial uncertainty map from that of repeatedly refreshing uncertainty feedback, we evaluate \emph{Frozen-U}.
Frozen-U retains the same depth--uncertainty refinement and UCFC modules but reuses the uncertainty state $\mathbf{U}^{1}$ estimated after the first refinement update for subsequent context calibration and the ConvGRU uncertainty input. Depth and uncertainty continue to be updated and supervised at every iteration, and $\mathbf{U}^{1}$ is not detached from the computational graph.}

\begin{table}[!htbp]
\centering
\caption{Design comparison of ablation variants for UCFC feedback.}
\label{tab:ucfc_ablation_def}
\resizebox{\linewidth}{!}{

\begin{tabular}{lcccc}
\toprule
\textbf{\shortstack[c]{Model\\Variant}}
& \rui{\textbf{\shortstack[c]{Recurrent\\Refinement}}}
& \textbf{\shortstack[c]{Uncertainty\\Supervision}}
& \textbf{\shortstack[c]{Feedback\\Signal}}
& \textbf{\shortstack[c]{Fusion\\Type}} \\
\midrule
Single-step depth
& \rui{--} & -- & -- & -- \\
Single-step depth + Unc.
& \rui{--} & \checkmark & -- & -- \\
Plain refinement
& \rui{\checkmark} & -- & -- & -- \\
Uncertainty-only
& \rui{\checkmark} & \checkmark & -- & -- \\
Static learned gate
& \rui{\checkmark} & \checkmark & Learned static & Gate \\
Validity-mask gate
& \rui{\checkmark} & \checkmark & Valid mask & Gate \\
Residual gate
& \rui{\checkmark} & \checkmark & Depth residual & Gate \\
Simple conv fusion
& \rui{\checkmark} & \checkmark & Pred. $U$ & Conv \\
UCFC-PM
& \rui{\checkmark} & \checkmark & Const. map & UCFC \\
\rui{Frozen-U}
& \rui{\checkmark} & \checkmark & \rui{Pred. $U^1$ (frozen)} & UCFC \\
UCFC (Ours)
& \rui{\checkmark} & \checkmark & Pred. $U$ & UCFC \\
\bottomrule
\end{tabular}
}
\end{table}

\begin{table}[!htbp]
\centering
\caption{In-domain HAMMER ablation of UCFC feedback. REL-AUSE is computed using REL as the pixel-wise error.}
\label{tab:ucfc_indomain_hammer_app}
\resizebox{\linewidth}{!}{

\begin{tabular}{lccc|ccc}
\toprule
\multirow{2}{*}{\textbf{Model Variant}}
& \multicolumn{3}{c|}{\textbf{Object Pixels}}
& \multicolumn{3}{c}{\textbf{All-valid Pixels}} \\
\cmidrule(lr){2-4} \cmidrule(lr){5-7}
& \textbf{REL}$\downarrow$
& \textbf{$\delta_{1.05}$}$\uparrow$
& \textbf{REL-AUSE}$\downarrow$
& \textbf{REL}$\downarrow$
& \textbf{$\delta_{1.05}$}$\uparrow$
& \textbf{REL-AUSE}$\downarrow$ \\
\midrule
Single-step depth
& 0.0341 & 80.60 & -- & 0.0324 & 83.67 & -- \\
Single-step depth + Unc.
& 0.0372 & 76.19 & 0.00152 & 0.0346 & 80.95 & 0.00105 \\
Plain refinement
& 0.0320 & 82.55 & -- & 0.0259 & 88.15 & -- \\
Uncertainty-only
& 0.0340 & 80.16 & 0.00132 & 0.0316 & 83.85 & 0.00111 \\
Static learned gate
& 0.0292 & 85.53 & 0.00117 & 0.0264 & 87.69 & 0.00097 \\
Validity-mask gate
& 0.0324 & 82.01 & 0.00104 & 0.0293 & 85.89 & 0.00097 \\
Residual gate
& 0.0294 & 85.01 & 0.00116 & 0.0267 & 87.79 & 0.00098 \\
Simple conv fusion
& 0.0313 & 82.05 & 0.00112 & 0.0264 & 87.85 & 0.00095 \\
UCFC-PM
& 0.0295 & 83.86 & 0.00189 & 0.0264 & 87.58 & 0.00130 \\
\rui{Frozen-U}
& \rui{0.0318} & \rui{82.84} & \rui{0.00111} & \rui{0.0283} & \rui{86.37} & \rui{0.00101} \\
UCFC (Ours)
& \textbf{0.0285} & \textbf{85.75} & \textbf{0.00095} & \textbf{0.0241} & \textbf{89.58} & \textbf{0.00090} \\
\bottomrule
\end{tabular}%
}
\end{table}

\noindent\textbf{\rui{Bootstrap Analysis of UCFC Ablation.}}
\label{app:ucfc_rank_bootstrap}
\rui{We examine how the mean ranks of the six variants vary with the evaluation samples. Within each of the eight datasets, consecutive images are grouped into batches of up to four. We perform 10,000 bootstrap repetitions by sampling batches with replacement within each dataset, using identical batch indices for all six variants and both metrics. We recalculate REL and $\delta_{1.05}$ from the sampled batches, rank the variants separately for each metric and dataset (assigning average ranks to ties), and average the resulting 16 ranks. The 95\% confidence interval of each variant's mean rank is given by the 2.5th and 97.5th percentiles of its bootstrap distribution.

Table~\ref{tab:ucfc_rank_bootstrap} summarizes the results. UCFC achieves the lowest mean rank in all 10,000 resamples. These intervals measure variation due to the evaluation samples for the trained models, not variation across training seeds.}

\begin{table}[!htbp]
\centering
\caption{\rui{Mean ranks and their 95\% bootstrap confidence intervals for the six UCFC ablation variants.}}
\label{tab:ucfc_rank_bootstrap}
\resizebox{\linewidth}{!}{%
\rui{
\begin{tabular}{lcc}
\toprule
Model Variant & Mean Rank $\downarrow$ & 95\% CI of Mean Rank \\
\midrule
Plain refinement & 4.84 & [4.4375, 4.8750] \\
Residual gate & 4.69 & [4.5625, 4.8750] \\
Simple conv fusion & 2.97 & [2.8125, 3.0625] \\
UCFC-PM & 3.28 & [3.0625, 3.5000] \\
Frozen-U & 3.09 & [2.9375, 3.3750] \\
UCFC (Ours) & \textbf{2.13} & \textbf{[1.8750, 2.3750]} \\
\bottomrule
\end{tabular}
}
}
\end{table}

\noindent\textbf{Results.}
Table~\ref{tab:ucfc_indomain_hammer_app} reports the in-domain HAMMER ablation on both object pixels and all-valid pixels.
Uncertainty-only learning does not improve depth restoration over plain refinement, indicating that predicting uncertainty without feeding it back is insufficient.
Static learned gating, validity-mask gating, residual gating, and simple convolutional fusion improve some metrics but remain weaker than UCFC.
UCFC achieves the best REL and $\delta_{1.05}$ on both object pixels and all-valid pixels, and also obtains the lowest REL-AUSE among variants with uncertainty output.
These results show that the predicted uncertainty is most useful when it is recurrently fed back to calibrate refinement context.
\rui{Frozen-U improves over Uncertainty-only but remains below UCFC across both evaluation masks, supporting UCFC's recurrent feedback design and suggesting additional benefits from refreshing the uncertainty signal at each iteration.}
The full zero-shot ablation in Table~\ref{tab:zs_ucfc_ablation_full}, which additionally includes the single-step depth and single-step depth+unc. baselines, shows a similar trend: single-step prediction is weaker than recurrent refinement, and UCFC achieves the best overall mean rank after uncertainty is dynamically fed back into the refinement process.
\rui{With Frozen-U included, UCFC retains the best mean rank (2.25 versus 3.22 for Frozen-U) and achieves lower REL on six of eight datasets and higher $\delta_{1.05}$ on six.}

\noindent\textbf{\rui{Stop-Gradient Ablation.}}
\rui{We compare UCFC with and without stop-gradient in Eq.~\eqref{eq:l_unc_final} on in-domain HAMMER, keeping the loss, initialization, data split, and training settings identical. As shown in Table~\ref{tab:stop_grad_hammer}, stop-gradient improves depth accuracy on both object and all-valid pixels, whereas removing it yields better uncertainty ranking (lower REL-AUSE and higher REL-AURG). We retain stop-gradient to prioritize accurate depth restoration.}

\begin{table}[!htbp]
\centering
\caption{\rui{Effect of stop-gradient on in-domain HAMMER.}}
\label{tab:stop_grad_hammer}
\rui{\resizebox{\linewidth}{!}{\begin{tabular}{llcccc}
\toprule
\textbf{Mask} & \textbf{Stop-grad.}
& \textbf{REL}$\downarrow$
& \textbf{$\delta_{1.05}$}$\uparrow$
& \textbf{REL-AUSE}$\downarrow$
& \textbf{REL-AURG}$\uparrow$ \\
\midrule
\multirow{2}{*}{Object}
& w/ & \textbf{0.0285} & \textbf{85.75} & 0.000951 & 0.001281 \\
& w/o & 0.0296 & 84.28 & \textbf{0.000896} & \textbf{0.001454} \\
\midrule
\multirow{2}{*}{All-valid}
& w/ & \textbf{0.0241} & \textbf{89.58} & 0.000901 & 0.001151 \\
& w/o & 0.0260 & 87.68 & \textbf{0.000812} & \textbf{0.001519} \\
\bottomrule
\end{tabular}
}}
\end{table}

\begin{table*}[!tbp]
\centering
\caption{Zero-shot object-pixel evaluation on the $2\times2$ contribution decomposition of ADF+Robust-SN and UCFC. Mean rank is computed over REL and $\delta_{1.05}$ across eight target datasets \rui{among the four configurations in this decomposition}.}
\label{tab:zs_2x2_decomp_full}
\resizebox{\textwidth}{!}{

\begin{tabular}{cc*{16}{c}c}
\toprule
\multirow{2}{*}{ADF+Robust-SN}
& \multirow{2}{*}{UCFC}
& \multicolumn{2}{c}{HAMMER}
& \multicolumn{2}{c}{HouseCat6D}
& \multicolumn{2}{c}{TransCG}
& \multicolumn{2}{c}{\rui{GN-Trans}}
& \multicolumn{2}{c}{YCB-V}
& \multicolumn{2}{c}{T-LESS}
& \multicolumn{2}{c}{XYZ-IBD}
& \multicolumn{2}{c}{ROBI}
& \multirow{2}{*}{Mean Rank} \\
\cmidrule(lr){3-4}
\cmidrule(lr){5-6}
\cmidrule(lr){7-8}
\cmidrule(lr){9-10}
\cmidrule(lr){11-12}
\cmidrule(lr){13-14}
\cmidrule(lr){15-16}
\cmidrule(lr){17-18}
& & REL$\downarrow$ & $\delta_{1.05}\uparrow$
& REL$\downarrow$ & $\delta_{1.05}\uparrow$
& REL$\downarrow$ & $\delta_{1.05}\uparrow$
& REL$\downarrow$ & $\delta_{1.05}\uparrow$
& REL$\downarrow$ & $\delta_{1.05}\uparrow$
& REL$\downarrow$ & $\delta_{1.05}\uparrow$
& REL$\downarrow$ & $\delta_{1.05}\uparrow$
& REL$\downarrow$ & $\delta_{1.05}\uparrow$
& \\
\midrule
\multirow{2}{*}{--}
& --
& 0.0372 & 78.83
& 0.0355 & 84.25
& 0.0214 & 87.69
& 0.0773 & 64.53
& \textbf{0.0227} & 92.64
& 0.0161 & 95.17
& 0.0060 & 98.51
& 0.0106 & 98.32
& 3.41  \\

& \checkmark
& 0.0349 & 81.50
& 0.0347 & 85.16
& 0.0206 & 88.56
& \textbf{0.0646} & \textbf{69.42}
& 0.0234 & \textbf{94.06}
& 0.0145 & 95.35
& \textbf{0.0054} & \textbf{99.50}
& 0.0108 & 98.59
& 1.94  \\
\midrule
\multirow{2}{*}{\checkmark}
& --
& 0.0334 & 80.56
& 0.0345 & 84.62
& 0.0208 & 87.70
& 0.0665 & 68.32
& 0.0230 & 91.44
& 0.0153 & 95.32
& 0.0060 & 99.43
& 0.0113 & 98.01
& 2.78 \\

& \checkmark
& \textbf{0.0326} & \textbf{82.38}
& \textbf{0.0334} & \textbf{86.86}
& \textbf{0.0199} & \textbf{89.52}
& 0.0701 & 67.28
& 0.0262 & 93.94
& \textbf{0.0137} & \textbf{95.73}
& 0.0069 & 98.35
& \textbf{0.0103} & \textbf{98.74}
& \textbf{1.88} \\
\bottomrule
\end{tabular}
}
\end{table*}

\begin{table*}[!tbp]
\centering
\small
\caption{Full Ablation of UCFC feedback. Metrics are computed over \textbf{Object} pixels. \rui{Mean rank is computed among all eight variants over REL and $\delta_{1.05}$ across eight zero-shot datasets.}}
\resizebox{\textwidth}{!}{
\begin{tabular}{l*{16}{c}c}
\toprule
\multirow{2}{*}{Model Variant} 
& \multicolumn{2}{c}{HAMMER} 
& \multicolumn{2}{c}{HouseCat6D} 
& \multicolumn{2}{c}{TransCG} 
& \multicolumn{2}{c}{\rui{GN-Trans}} 
& \multicolumn{2}{c}{YCB-V} 
& \multicolumn{2}{c}{T-LESS} 
& \multicolumn{2}{c}{XYZ-IBD} 
& \multicolumn{2}{c}{ROBI} 
& \multirow{2}{*}{Mean Rank} \\
\cmidrule(lr){2-3} 
\cmidrule(lr){4-5} 
\cmidrule(lr){6-7} 
\cmidrule(lr){8-9} 
\cmidrule(lr){10-11} 
\cmidrule(lr){12-13} 
\cmidrule(lr){14-15} 
\cmidrule(lr){16-17}
& REL$\downarrow$ & $\delta_{1.05}\uparrow$
& REL$\downarrow$ & $\delta_{1.05}\uparrow$
& REL$\downarrow$ & $\delta_{1.05}\uparrow$
& REL$\downarrow$ & $\delta_{1.05}\uparrow$
& REL$\downarrow$ & $\delta_{1.05}\uparrow$
& REL$\downarrow$ & $\delta_{1.05}\uparrow$
& REL$\downarrow$ & $\delta_{1.05}\uparrow$
& REL$\downarrow$ & $\delta_{1.05}\uparrow$
& \\
\midrule
Single-step depth
& 0.0482 & 70.02
& 0.0396 & 75.75
& 0.0259 & 83.09
& 0.0779 & 62.12
& 0.0409 & 72.35
& 0.0223 & 89.22
& 0.0094 & 98.68
& 0.0147 & 95.83
& \rui{7.31} \\

Single-step depth + Unc.
& 0.0509 & 73.96
& 0.0458 & 81.26
& 0.0261 & 86.16
& 0.0808 & 61.82
& 0.0423 & 83.86
& 0.0248 & 92.30
& 0.0083 & 99.12
& 0.0155 & 96.52
& \rui{7.13} \\

Plain refinement 
& 0.0365 & 78.05
& 0.0346 & 85.14
& 0.0221 & 86.98
& 0.0566 & 72.23
& 0.0295 & 91.30
& 0.0149 & 95.33
& 0.0090 & 97.87
& 0.0124 & 98.17
& \rui{5.03} \\

Residual gate 
& 0.0393 & 75.92
& 0.0391 & 81.85
& 0.0228 & 86.24
& 0.0682 & 66.67
& 0.0267 & 93.31
& 0.0148 & 94.67
& 0.0065 & 99.38
& 0.0116 & 98.48
& \rui{4.69} \\

Simple conv fusion 
& 0.0351 & 82.36
& 0.0330 & 85.80
& 0.0214 & 87.75
& 0.0727 & 64.39
& \textbf{0.0223} & \textbf{94.11}
& 0.0144 & 95.07
& \textbf{0.0057} & \textbf{99.59}
& 0.0114 & 98.47
& \rui{2.97} \\

UCFC-PM
& 0.0345 & 79.46
& \textbf{0.0322} & 86.83
& 0.0220 & 87.43
& \textbf{0.0545} & \textbf{72.47}
& 0.0404 & 79.91
& 0.0146 & 95.40
& 0.0081 & 99.01
& 0.0121 & 98.54
& \rui{3.41} \\

\rui{Frozen-U}
& \rui{0.0351} & \rui{79.17}
& \rui{0.0346} & \rui{85.54}
& \rui{0.0213} & \rui{87.99}
& \rui{0.0603} & \rui{70.93}
& \rui{0.0256} & \rui{93.66}
& \rui{0.0141} & \rui{95.45}
& \rui{0.0079} & \rui{98.66}
& \rui{0.0117} & \rui{98.54}
& \rui{3.22} \\

UCFC (Ours) 
& \textbf{0.0326} & \textbf{82.38}
& 0.0334 & \textbf{86.86}
& \textbf{0.0199} & \textbf{89.52}
& 0.0701 & 67.28
& 0.0262 & 93.94
& \textbf{0.0137} & \textbf{95.73}
& 0.0069 & 98.35
& \textbf{0.0103} & \textbf{98.74}
& \rui{\textbf{2.25}} \\
\bottomrule
\end{tabular}
}
\label{tab:zs_ucfc_ablation_full}
\end{table*}

\subsection{Ablation Studies of ADF and Robust-SN}
\label{app:adf_sn_ucfc_decomposition}
We further decompose the contributions of ADF+Robust-SN and UCFC in the zero-shot setting. 
All four variants are retrained and evaluated with the same mixed-dataset pretraining protocol, DAV2 prior, recurrent refinement backbone, uncertainty modeling objective, input resolution, and evaluation protocol. 
We only toggle two factors: whether ADF+Robust-SN is used for initialization and log-depth normalization, and whether the predicted uncertainty is fed back into refinement through UCFC. 
This decomposition separates the effect of robust metric initialization/normalization from the effect of recurrent uncertainty feedback.

As shown in Table~\ref{tab:zs_2x2_decomp_full}, enabling UCFC improves the mean rank from 3.41 to 1.94 when ADF+Robust-SN is disabled, and from 2.78 to 1.88 when ADF+Robust-SN is enabled. 
\rui{Thus, UCFC improves the aggregate mean rank under both initialization and normalization settings, although its benefits vary across individual datasets.} 
Conversely, ADF+Robust-SN addresses a complementary failure mode by providing a more stable metric initialization under biased sensor depth, but it does not by itself turn uncertainty into an active refinement signal.

\noindent\textbf{\rui{Robustness to Corrupted Sensor Depth.}}
\rui{We simulate biased sensor measurements by multiplying depths in spatially concentrated regions by $1.1$, corrupting $10\%$, $30\%$, or $50\%$ of originally valid pixels while preserving the missing-depth mask. The $0\%$ setting uses the original observations. We first compare least squares (LS) and robust estimation of the log-depth bias, with all other settings fixed. Table~\ref{tab:structured_corruption} reports REL at ADF initialization and after refinement, evaluated over object pixels. Robust estimation improves both metrics on TransCG and GN-Trans, while LS performs slightly better on HAMMER. At $50\%$ corruption on GN-Trans, Robust reduces initialization REL from 0.1999 to 0.1898 and final REL from 0.1210 to 0.1132.}

\begin{table}[!htbp]
\centering
\small
\caption{\rui{Effect of structured depth corruption on ADF initialization and final REL.}}
\label{tab:structured_corruption}
\rui{\resizebox{\linewidth}{!}{\begin{tabular}{llcccccccc}
\toprule
\multirow{2}{*}{Dataset} & \multirow{2}{*}{Variant}
& \multicolumn{4}{c}{Init REL $\downarrow$}
& \multicolumn{4}{c}{Final REL $\downarrow$} \\
\cmidrule(lr){3-6} \cmidrule(lr){7-10}
& & 0\% & 10\% & 30\% & 50\%
& 0\% & 10\% & 30\% & 50\% \\
\midrule
\multirow{2}{*}{HAMMER}
& LS
& \textbf{0.0884} & \textbf{0.0986} & \textbf{0.1225} & \textbf{0.1447}
& \textbf{0.0320} & \textbf{0.0389} & \textbf{0.0622} & \textbf{0.0879} \\
& Robust
& 0.0979 & 0.1091 & 0.1355 & 0.1604
& 0.0324 & 0.0393 & 0.0624 & 0.0880 \\
\midrule
\multirow{2}{*}{TransCG}
& LS
& 0.0767 & 0.0880 & 0.1150 & 0.1367
& 0.0195 & 0.0258 & 0.0508 & 0.0777 \\
& Robust
& \textbf{0.0766} & \textbf{0.0868} & \textbf{0.1117} & \textbf{0.1343}
& \textbf{0.0193} & \textbf{0.0254} & \textbf{0.0493} & \textbf{0.0750} \\
\midrule
\multirow{2}{*}{GN-Trans}
& LS
& 0.1501 & 0.1590 & 0.1801 & 0.1999
& 0.0714 & 0.0770 & 0.0977 & 0.1210 \\
& Robust
& \textbf{0.1382} & \textbf{0.1476} & \textbf{0.1693} & \textbf{0.1898}
& \textbf{0.0677} & \textbf{0.0739} & \textbf{0.0942} & \textbf{0.1132} \\
\bottomrule
\end{tabular}}}
\end{table}

\rui{
We further compare four residual weighting schemes: uniform (Median), Huber, Tukey, and exponential, under the same structured corruption protocol. All variants use the same provisional estimate $a_0$, weighted median estimators, trained checkpoint, and five refinement iterations. 
In addition to final REL, we measure the change in the estimated anchor depth between the original and $50\%$ corrupted observations: $A_{50}=\frac{1}{N}\sum_{i=1}^{N} \bigl|\log(s_{\mathrm{anc},i}^{50\%}/s_{\mathrm{anc},i}^{0\%})\bigr|$,
where $N$ is the number of evaluated frames.
A smaller $A_{50}$ indicates a more stable anchor estimate under corruption. Although the anchor cancels when the initial depth is recovered in metric units, it affects the normalized inputs used for subsequent refinement.
}

\rui{
As shown in Table~\ref{tab:weighting_corruption}, exponential weighting achieves the lowest or jointly lowest final REL in most settings. The exceptions are HAMMER at $50\%$ corruption (0.0880 versus 0.0878 for Tukey) and GN-Trans at $10\%$ corruption (0.0739 versus 0.0738 for Tukey). At $50\%$ corruption on GN-Trans, exponential weighting achieves a final REL of 0.1132, compared with 0.1194 for Tukey and 0.1199 for Median. 
Exponential weighting also produces the smallest $A_{50}$ on TransCG and GN-Trans, and matches Tukey on HAMMER at the reported precision. These results support exponential weighting as a competitive choice, although its advantage is not uniform across datasets and corruption levels. The exponential scheme corresponds to the Robust variant in Table~\ref{tab:structured_corruption}.
}

\begin{table}[!htbp]
\centering
\footnotesize
\caption{\rui{Comparison of residual weighting functions under structured depth corruption.}}
\label{tab:weighting_corruption}
\setlength{\tabcolsep}{2.5pt}
\rui{\resizebox{\columnwidth}{!}{%
\begin{tabular}{llccccc}
\toprule
& & \multicolumn{4}{c}{Final REL $\downarrow$} & Anchor shift $\downarrow$ \\
\cmidrule(lr){3-6}\cmidrule(lr){7-7}
Dataset & Weighting & $0\%$ & $10\%$ & $30\%$ & $50\%$ & $A_{50}$ \\
\midrule
\multirow{4}{*}{HAMMER}
& Median      & 0.0327 & 0.0401 & 0.0638 & 0.0895 & 0.0554 \\
& Huber       & 0.0325 & 0.0397 & 0.0628 & 0.0881 & 0.0421 \\
& Tukey       & 0.0325 & 0.0396 & 0.0625 & \textbf{0.0878} & \textbf{0.0415} \\
& Exponential & \textbf{0.0324} & \textbf{0.0393} & \textbf{0.0624} & 0.0880 & \textbf{0.0415} \\
\midrule
\multirow{4}{*}{TransCG}
& Median      & \textbf{0.0193} & 0.0257 & 0.0494 & 0.0756 & 0.0585 \\
& Huber       & \textbf{0.0193} & 0.0257 & 0.0494 & 0.0756 & 0.0586 \\
& Tukey       & \textbf{0.0193} & 0.0257 & \textbf{0.0493} & 0.0756 & 0.0581 \\
& Exponential & \textbf{0.0193} & \textbf{0.0254} & \textbf{0.0493} & \textbf{0.0750} & \textbf{0.0578} \\
\midrule
\multirow{4}{*}{GN-Trans}
& Median      & 0.0683 & 0.0744 & 0.0956 & 0.1199 & 0.0641 \\
& Huber       & 0.0678 & 0.0739 & 0.0951 & 0.1196 & 0.0627 \\
& Tukey       & \textbf{0.0677} & \textbf{0.0738} & 0.0949 & 0.1194 & 0.0615 \\
& Exponential & \textbf{0.0677} & 0.0739 & \textbf{0.0942} & \textbf{0.1132} & \textbf{0.0589} \\
\bottomrule
\end{tabular}%
}}
\end{table}

\noindent\textbf{\rui{Effect of Metric Affine Alignment.}}
\rui{We compare Robust-SN, which estimates only a scale, with metric affine alignment that estimates both a scale $s$ and a shift $t$ from valid sensor depths. Using the same trained network, we evaluate 256 evenly sampled test frames from each of seven unseen datasets and all 45 XYZ-IBD frames (1,837 frames in total). No ground truth is used for alignment. Table~\ref{tab:metric_affine_alignment} reports the difference in REL (affine minus Robust-SN) on missing object pixels at initialization and after the first refinement iteration, and on all object pixels at the final iteration. Negative values indicate better results with affine alignment.}

\begin{table}[!htbp]
\centering
\small
\caption{\rui{Effect of metric affine alignment. Values are REL differences (affine alignment minus Robust-SN).}}
\label{tab:metric_affine_alignment}
\rui{\resizebox{\linewidth}{!}{\begin{tabular}{lccc}
\toprule
\textbf{Dataset} & \textbf{Init. $\Delta$REL$\downarrow$} & \textbf{Iter. 1 $\Delta$REL$\downarrow$} & \textbf{Final $\Delta$REL$\downarrow$} \\
\midrule
HAMMER       & -0.10888 & +0.00042 & -0.00008 \\
HouseCat6D   & -0.01533 & +0.00035 & -0.00003 \\
TransCG      & -0.14735 & -0.00046 & -0.00004 \\
GN-Trans     & -0.10702 & +0.00795 & -0.00016 \\
YCB-V        & -0.41380 & -0.00232 & -0.00015 \\
T-LESS       & -0.21179 & +0.00688 & +0.00023 \\
ROBI         & -0.09053 & -0.00152 & -0.00058 \\
XYZ-IBD      & -0.05960 & -0.00013 & +0.00005 \\
\midrule
\textbf{Macro avg.} & \textbf{-0.14429} & \textbf{+0.00140} & \textbf{-0.00009} \\
\bottomrule
\end{tabular}
}}
\end{table}

\rui{Affine alignment reduces the mean initialization REL on missing object pixels from 0.28544 to 0.14115 (50.55\%), improving all eight datasets. However, after the first refinement iteration, the mean REL on these pixels increases slightly from 0.05995 to 0.06135. Final object REL changes only from 0.02665 to 0.02656, with small degradations on T-LESS and XYZ-IBD. We therefore retain Robust-SN, since adding the shift does not consistently improve the final depth predictions.}

\subsection{Failure Analysis}
\label{app:failure_taxonomy}
\rui{We compare UCFC with UCFC-PM on HAMMER and GN-Trans to examine where uncertainty feedback improves or degrades depth restoration. UCFC-PM retains the same calibration modules but replaces predicted uncertainty with a constant map. Both methods are evaluated on object pixels at $518\times518$ resolution.}

\rui{We divide object pixels into six groups based on sensor depth, the aligned DAV2 prior, and the initial depth used for refinement. Sensor measurements are valid when positive and finite; valid measurements with relative error above $5\%$ are considered biased, and the others are reliable. For biased measurements, we distinguish whether the prior is accurate. When the prior is also inaccurate, we further check whether it agrees with the sensor, using $|\log D_{\mathrm{sen}}-\log D_{\mathrm{prior}}|\leq\log(1.05)$. For missing measurements, we distinguish whether the initial depth is accurate. Prior and initial depth estimates are considered accurate when their relative error is at most $5\%$.}

\rui{Table~\ref{tab:failure_taxonomy} reports the average percentage of object pixels in each group, Mean REL, and the percentage of failed predictions (pixel REL $>0.05$). Pixels (\%) denotes the mean within-image share of each condition. Mean REL uses the same image weighting as Table~\ref{tab:zs_ucfc_ablation}: each pixel is weighted by the inverse of the number of object pixels in its image. Failed predictions are counted over all pixels within each group. The final column gives each group's contribution to the overall difference in Mean REL (UCFC minus PM), obtained by multiplying its pixel fraction by its Mean REL difference. Positive values indicate higher error for UCFC.}

\begin{table*}[!tbp]
\centering
\scriptsize
\caption{\rui{Failure analysis on HAMMER and GN-Trans.}}
\label{tab:failure_taxonomy}
\label{tab:ucfc_failure_mode_comparison}
\rui{\resizebox{\textwidth}{!}{\begin{tabular}{@{}lrrrrrr@{}}
\toprule
& & \multicolumn{2}{c}{Mean REL $\downarrow$}
    & \multicolumn{2}{c}{Failed (\%) $\downarrow$} & \\
\cmidrule(lr){3-4}\cmidrule(lr){5-6}
Condition & Pixels (\%) & PM & UCFC & PM & UCFC & $\Delta$REL \\
\midrule
\multicolumn{7}{@{}l}{\textit{HAMMER}} \\
Biased sensor, inaccurate prior, disagree & 13.45 & 0.0667 & \textbf{0.0590} & 59.80 & \textbf{47.92} & $-0.001037$ \\
Biased sensor, inaccurate prior, agree    &  3.48 & 0.0530 & \textbf{0.0466} & 48.86 & \textbf{38.44} & $-0.000223$ \\
Biased sensor, accurate prior             &  2.94 & 0.0500 & \textbf{0.0454} & 41.26 & \textbf{37.86} & $-0.000134$ \\
Missing sensor, inaccurate initialization & 13.01 & 0.0538 & \textbf{0.0496} & 39.10 & \textbf{31.54} & $-0.000550$ \\
Missing sensor, accurate initialization   &  5.35 & 0.0298 & \textbf{0.0297} & \textbf{14.60} & 14.98 & $-0.000001$ \\
Reliable sensor & 61.78 & \textbf{0.0221} & 0.0222 & \textbf{4.88} & 6.19 & $+0.000055$ \\
\textbf{All object pixels} & 100.00 & 0.0345 & \textbf{0.0326} & 19.38 & \textbf{17.28} & $\mathbf{-0.001890}$ \\
\midrule
\multicolumn{7}{@{}l}{\textit{\rui{GN-Trans}}} \\
Biased sensor, inaccurate prior, disagree & 22.67 & \textbf{0.1142} & 0.1534 & \textbf{54.43} & 69.75 & $+0.008872$ \\
Biased sensor, inaccurate prior, agree    &  3.80 & \textbf{0.0765} & 0.1085 & \textbf{50.25} & 71.61 & $+0.001214$ \\
Biased sensor, accurate prior             &  5.59 & \textbf{0.0875} & 0.1221 & \textbf{48.82} & 69.05 & $+0.001931$ \\
Missing sensor, inaccurate initialization & 17.66 & \textbf{0.0809} & 0.1055 & \textbf{44.70} & 48.95 & $+0.004340$ \\
Missing sensor, accurate initialization   &  5.90 & \textbf{0.0338} & 0.0356 & 17.08 & \textbf{16.59} & $+0.000104$ \\
Reliable sensor & 44.40 & 0.0103 & \textbf{0.0082} & 3.72 & \textbf{1.73} & $-0.000954$ \\
\textbf{All object pixels} & 100.00 & \textbf{0.0545} & 0.0701 & \textbf{27.81} & 33.14 & $\mathbf{+0.015507}$ \\
\bottomrule
\end{tabular}
}}
\end{table*}

\rui{On HAMMER, UCFC reduces overall Mean REL from $0.03454$ to $0.03265$ and the failure rate from $19.38\%$ to $17.28\%$. Most of the reduction comes from biased sensor measurements with an inaccurate, disagreeing prior and from missing regions with inaccurate initialization. UCFC performs slightly worse on reliable sensor pixels, where Mean REL increases from $0.02211$ to $0.02220$.}

\rui{On GN-Trans, UCFC increases overall Mean REL from $0.05455$ to $0.07006$ and the failure rate from $27.81\%$ to $33.14\%$. Biased sensor measurements account for $77.5\%$ of this increase in Mean REL, with additional degradation in missing regions where initialization is inaccurate. In contrast, UCFC reduces Mean REL on reliable sensor pixels from $0.01031$ to $0.00816$. Thus, the main difficulty on GN-Trans lies in correcting inaccurate depth rather than preserving reliable measurements.}

\begin{figure*}[t]
\centering
\includegraphics[width=\textwidth]{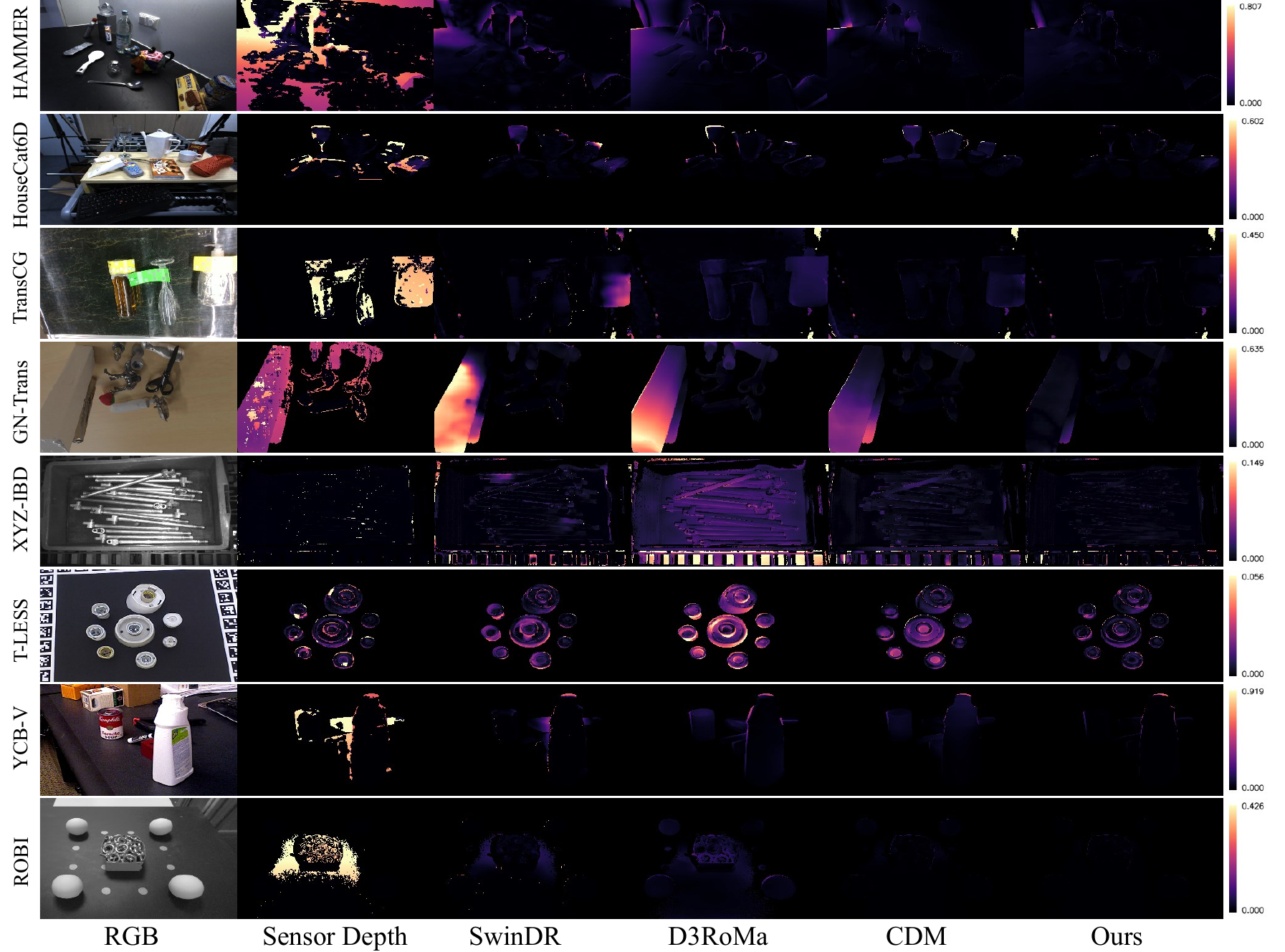}
\caption{Qualitative comparisons on zero-shot datasets. We show RGB, sensor depth, SwinDR, D$^3$RoMa, CDM, and our absolute depth error.}
\label{fig:zs_qualitative}
\end{figure*}

\begin{figure*}[t]
\centering
\includegraphics[width=\textwidth]{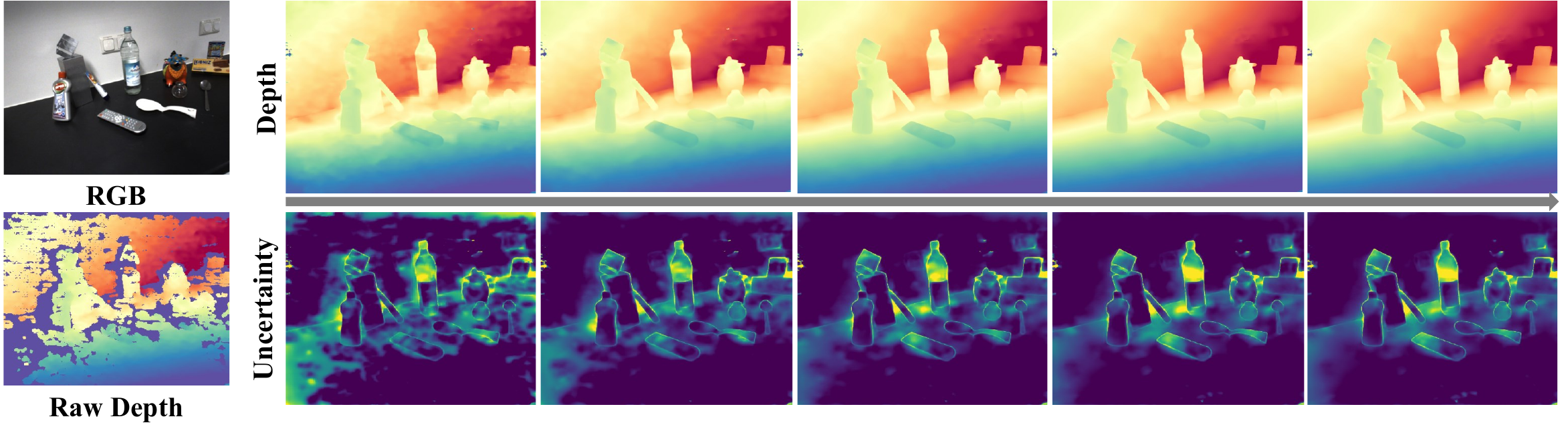}
\caption{\methodname\ iterative refinement visualization. We show RGB with raw depth input and the predicted depth and uncertainty at each refinement step.}
\label{fig:refine_steps}
\end{figure*}

\subsection{Visualization}
Fig.~\ref{fig:zs_qualitative} provides qualitative comparisons on representative zero-shot datasets. Across diverse scenes and sensors, our restored depth is typically more complete on non-Lambertian objects (e.g., transparent or reflective surfaces) and exhibits fewer spurious structures than competing methods, supporting the quantitative improvements reported in the main paper.

Fig.~\ref{fig:refine_steps} visualizes how \methodname\ progressively refines depth and uncertainty. Over refinement steps, the predicted depth becomes more complete and less noisy, while the uncertainty map highlights unreliable regions and stabilizes as the refinement converges.

\subsection{Category-level 6D Pose Estimation}
\label{app:cat6dpose}
\noindent\textbf{Metrics.}
Following the HouseCat6D evaluation protocol, we report both 3D detection/size estimation and 6D pose estimation metrics. For 3D detection and object dimension estimation, we use 3D IoU at different thresholds (IoU@25/50/75). For 6D pose estimation, we report the average precision of object instances whose pose error is below $m$\,cm in translation and $n^\circ$ in rotation (e.g., $5^\circ$2cm, $5^\circ$5cm, $10^\circ$2cm, $10^\circ$5cm). To better isolate pose quality from detection failures, 6D pose evaluation is decoupled from detection by using a low detection threshold (10\% 3D bounding-box overlap between prediction and ground truth) so that most objects are included in the pose evaluation. For symmetric categories (bottle, bowl, and can), rotations around the vertical axis are treated as equivalent, i.e., the predicted 3D bounding box is allowed to freely rotate around the object vertical axis without penalty.


\begin{table*}[!tbp]
\centering
\caption{Category-level 6D pose results under different training/testing depth on HouseCat6D~\cite{housecat6d}. \rui{Results are grouped by training depth type (Raw, GT, and Restored). The final group reports mean $\pm$ sample standard deviation across three depth-model seeds.}}
\label{tab:cat_6d_pose_full}
\setlength{\tabcolsep}{5.5pt}
\resizebox{\linewidth}{!}{
\begin{tabular}{llccccccc}
\toprule
Train Depth & Test Depth & IoU@25$\uparrow$ & IoU@50$\uparrow$ & IoU@75$\uparrow$ & $5^\circ$2cm$\uparrow$ & $5^\circ$5cm$\uparrow$ & $10^\circ$2cm$\uparrow$ & $10^\circ$5cm$\uparrow$ \\
\midrule
Raw                  & Raw                         & 77.3 & \textbf{52.6} & \textbf{17.2} &  \textbf{6.7} & \textbf{8.9} & \textbf{17.6} & \textbf{25.4} \\
Raw                  & Restored                    & 77.9 & 52.2 & 11.4 &  4.4 &  7.1 & 12.4 & 21.9 \\
Raw                  & Restored+Unc-Sampling  & \textbf{78.4} & 52.3 & 11.0 &  4.2 &  7.1 & 12.0 & 22.1 \\
\midrule
GT                   & Raw                         & 60.6 & 27.0 &  3.0 &  1.2 &  2.8 &  3.6 &  8.0 \\
GT                   & Restored                    & 60.7 & 25.5 &  3.3 &  2.7 &  4.2 &  6.6 & 11.9 \\
GT                   & Restored+Unc-Sampling       & \textbf{69.0} & \textbf{43.9} & \textbf{13.7} & \textbf{4.3} & \textbf{5.6} & \textbf{13.4} & \textbf{18.1} \\
\midrule
Restored              & Restored                    & \rui{\makecell[c]{\textbf{79.9}\\[-0.6ex]{\scriptsize$\pm3.1$}}} & \rui{\makecell[c]{\textbf{60.5}\\[-0.6ex]{\scriptsize$\pm4.3$}}} & \rui{\makecell[c]{30.3\\[-0.6ex]{\scriptsize$\pm2.8$}}} & \rui{\makecell[c]{13.3\\[-0.6ex]{\scriptsize$\pm0.6$}}} & \rui{\makecell[c]{15.0\\[-0.6ex]{\scriptsize$\pm0.5$}}} & \rui{\makecell[c]{28.9\\[-0.6ex]{\scriptsize$\pm1.9$}}} & \rui{\makecell[c]{36.7\\[-0.6ex]{\scriptsize$\pm2.1$}}} \\
Restored              & Restored+Unc-Sampling       & \rui{\makecell[c]{79.7\\[-0.6ex]{\scriptsize$\pm2.9$}}} & \rui{\makecell[c]{60.1\\[-0.6ex]{\scriptsize$\pm4.2$}}} & \rui{\makecell[c]{29.2\\[-0.6ex]{\scriptsize$\pm2.8$}}} & \rui{\makecell[c]{13.0\\[-0.6ex]{\scriptsize$\pm0.4$}}} & \rui{\makecell[c]{14.5\\[-0.6ex]{\scriptsize$\pm0.4$}}} & \rui{\makecell[c]{28.3\\[-0.6ex]{\scriptsize$\pm1.4$}}} & \rui{\makecell[c]{35.9\\[-0.6ex]{\scriptsize$\pm2.0$}}} \\
Restored+Unc-Sampling & Restored+Unc-Sampling       & \rui{\makecell[c]{79.4\\[-0.6ex]{\scriptsize$\pm1.1$}}} & \rui{\makecell[c]{60.2\\[-0.6ex]{\scriptsize$\pm1.5$}}} & \rui{\makecell[c]{\textbf{31.9}\\[-0.6ex]{\scriptsize$\pm2.3$}}} & \rui{\makecell[c]{\textbf{14.3}\\[-0.6ex]{\scriptsize$\pm1.8$}}} & \rui{\makecell[c]{\textbf{16.4}\\[-0.6ex]{\scriptsize$\pm1.9$}}} & \rui{\makecell[c]{\textbf{30.8}\\[-0.6ex]{\scriptsize$\pm2.1$}}} & \rui{\makecell[c]{\textbf{39.7}\\[-0.6ex]{\scriptsize$\pm2.8$}}} \\
\bottomrule
\end{tabular}
}
\end{table*}

\rui{The three rows trained with restored depth report the mean and sample standard deviation across three independently trained UA-Depth models with different random seeds, while VI-Net uses a fixed random seed. For the Raw- and GT-trained rows, evaluations involving restored depth or uncertainty-guided sampling use predictions from the first UA-Depth seed.}

Table~\ref{tab:cat_6d_pose_full} provides complete results under different train/test depth types and sampling policies. When the depth type is matched, switching from Raw$\rightarrow$Raw to Restored$\rightarrow$Restored consistently improves all pose metrics, with larger gains under stricter criteria (e.g., IoU@50/75 and $10^\circ$5cm). Moreover, jointly training and testing with Restored+Unc-Sampling achieves the best overall performance, indicating that prioritizing low-uncertainty pixels yields a cleaner point cloud and benefits tight pose thresholds.

Beyond these matched-domain comparisons, Table~\ref{tab:cat_6d_pose_full} further reveals the depth-domain shift \rui{discussed} in the main paper. Training on raw depth but testing on restored depth (Raw$\rightarrow$Restored) leads to noticeable drops, especially at strict thresholds, suggesting that the restored depth distribution is not directly compatible with a model optimized for raw sensor artifacts. A similar effect appears when applying uncertainty-guided sampling only at test time (Restored$\rightarrow$Restored+Unc-Sampling), which slightly underperforms the matched setting; this implies that sampling changes the input statistics and should be included during training for best transfer. Based on these observations, the results reported in the main paper use matched training and testing depth domains to avoid conflating domain shift with the benefit of depth restoration.

\rui{The second block} (training with GT depth) shows that, under this setting, testing on Raw versus Restored depth yields broadly similar performance across metrics, suggesting that depth restoration alone does not significantly change the generalization behavior of a GT-trained model. Interestingly, adding uncertainty-guided sampling at test time consistently improves the results, indicating that prioritizing more reliable pixels can partially compensate for the mismatch when transferring from GT depth to sensor depth.

\begin{table*}[!tbp]
\centering
\caption{Effects of uncertainty utilization strategies on 6D grasp synthesis on \rui{GN-Trans}, following the GraspNet-1B evaluation protocol~\cite{graspnet1b}.}
\label{tab:grasp_6d_unc}
\setlength{\tabcolsep}{5.5pt}
\resizebox{\linewidth}{!}{

\begin{tabular}{lcccccccccccc}
\toprule
\multirow{2}{*}{Depth Type} 
& \multicolumn{3}{c}{Seen} 
& \multicolumn{3}{c}{Similar} 
& \multicolumn{3}{c}{Novel} 
& \multicolumn{3}{c}{Mean} \\
\cmidrule(lr){2-4} \cmidrule(lr){5-7} \cmidrule(lr){8-10} \cmidrule(lr){11-13}
& AP$\uparrow$ & CR$_{10}\downarrow$ & CR$_{50}\downarrow$
& AP$\uparrow$ & CR$_{10}\downarrow$ & CR$_{50}\downarrow$
& AP$\uparrow$ & CR$_{10}\downarrow$ & CR$_{50}\downarrow$
& AP$\uparrow$ & CR$_{10}\downarrow$ & CR$_{50}\downarrow$ \\
\midrule
Ours (Restored) 
& 32.67 & 47.32 & 58.74
& 25.48 & 56.32 & 62.87
& 11.26 & 64.31 & 65.10
& 23.14 & 55.98 & 62.23 \\
Ours (Restored+Unc\text{-}Sampling) 
& 31.65 & 48.60 & 59.84
& 24.58 & 57.73 & 63.62
& 10.55 & 65.76 & \textbf{61.81}
& 22.26 & 57.36 & \textbf{61.76} \\
Ours (Restored+Unc\text{-}Reweighting) 
& \textbf{33.54} & \textbf{45.44} & \textbf{58.41}
& \textbf{26.62} & \textbf{53.98} & \textbf{62.51}
& \textbf{11.89} & \textbf{62.36} & 64.42
& \textbf{24.02} & \textbf{53.93} & 61.78 \\
\bottomrule
\end{tabular}

}
\end{table*}

\subsection{6D Grasp Synthesis}
\label{app:6d_grasp}
\noindent\textbf{Metrics.}
We follow the standard GraspNet-1Billion evaluation protocol~\cite{graspnet1b} to measure grasp accuracy on \rui{GN-Trans}. Each predicted grasp pose is first associated with a target object by checking the scene point cloud enclosed by the gripper. A grasp is counted as a true positive if it satisfies the force-closure criterion under a given friction coefficient $\mu$. For cluttered scenes, where grasping is typically executed using only a few top-ranked candidates, GraspNet reports Average Precision computed over the top-$50$ grasps: $\mathrm{AP}_\mu$ is defined as the average Precision@k for $k\in[1,50]$ at friction $\mu$, and AP further averages $\mathrm{AP}_\mu$ over $\mu\in\{0.2,0.4,0.6,0.8,1.0,1.2\}$, analogous to COCO-style averaging. Following~\cite{graspnet1b}, we apply pose-based NMS before evaluation to reduce redundant near-duplicate grasps and prevent the metric from being dominated by highly similar poses or grasps from a single object.

\noindent\textbf{Why not uncertainty-guided sampling for 6D grasp synthesis.}
Although uncertainty-guided sampling can benefit category-level 6D pose estimation by selecting high-confidence depth for point cloud construction, we do not adopt it as the default for 6D grasp synthesis. As summarized in Table~\ref{tab:grasp_6d_unc}, Unc-Sampling consistently degrades grasp accuracy, reducing the mean AP from 23.14 to 22.26 across the three splits. To better diagnose this drop, we also report the \emph{Collision Rate} (CR): $\text{CR@}K$ denotes the average fraction of the top-$K$ ranked grasps that are flagged as colliding (with the object, table, or surrounding geometry) during collision checking, where lower is better. Under this metric, Unc-Sampling increases collisions in the practically relevant regime: the mean CR@10 rises from 55.98 to 57.36, and this trend holds for seen/similar/novel individually. A plausible reason is that high-uncertainty pixels concentrate near object boundaries, thin structures, and depth discontinuities; sampling them away tends to “shrink” the reconstructed shape and remove boundary support that is crucial for collision checking and contact reasoning, making top-ranked grasps more prone to penetrations or ignoring nearby obstacles. In contrast, Unc-Reweighting retains the full geometry while down-weighting unreliable points, improving mean AP to 24.02 and simultaneously reducing mean CR@10 to 53.93, indicating safer and more accurate grasp ranking.

\noindent\textbf{\rui{Uncertainty-Based Grasp Ranking.}}
\rui{We compare three ways to reweight AnyGrasp scores using uncertainty. The restored point cloud and predicted grasp candidates remain unchanged; only their ranking is updated. Center reweighting uses uncertainty at the projected grasp center. Local ball reweighting averages uncertainty within a 3-cm radius, while gripper region reweighting uses the gripper interaction region. Mean AP denotes mean average precision of predicted grasps across scenes, and CR@10 is the fraction of the top 10 ranked grasps identified as colliding. As shown in Table~\ref{tab:unc_grasp}, all three methods improve Mean AP and reduce CR@10 relative to the original ranking. Center reweighting gives the highest Mean AP (24.02 versus 23.14), while gripper region reweighting gives the lowest CR@10 (51.37 versus 55.98). These scores describe offline grasp ranking; low depth uncertainty alone does not guarantee a stable physical grasp.}

\begin{table}[!htbp]
\centering
\small
\caption{\rui{Effect of uncertainty-based grasp reweighting on GN-Trans.}}
\label{tab:unc_grasp}
\rui{\resizebox{\linewidth}{!}{\begin{tabular}{llcc}
\toprule
\textbf{Variant} & \textbf{Ranking by} & \textbf{Mean AP$\uparrow$} & \textbf{Mean CR@10$\downarrow$} \\
\midrule
No-$\mathbf{U}$ baseline 
& Original AnyGrasp score 
& 23.14 & 55.98 \\
Center reweight 
& Projected grasp-center uncertainty 
& \textbf{24.02} & 53.93 \\
Local-ball reweight 
& Mean uncertainty in 3-cm 3D ball neighborhood 
& 23.62 & 51.78 \\
Gripper-interaction reweight 
& Mean uncertainty in gripper interaction region 
& 23.82 & \textbf{51.37} \\
\bottomrule
\end{tabular}}}
\end{table}

\subsection{Real-world Evaluation and Demonstration}
\label{app:pick-and-place}
We evaluate real-robot grasping using a pick-and-place pipeline that clears cluttered tabletop scenes with 5 objects.
The robot is an AUBO--i5 arm equipped with a RobotDigg two-finger gripper, and an Intel RealSense D415 RGB-D camera is mounted on the gripper in an eye-in-hand configuration (Fig.~\ref{fig:robot_plat}).
For each trial, we capture one RGB-D frame at a fixed view, reconstruct a point cloud using the calibrated intrinsics and extrinsics, and synthesize 6D grasp candidates with AnyGrasp~\cite{anygrasp}.
We keep all grasping parameters identical across compared methods (e.g., point-cloud preprocessing, candidate number, non-maximum suppression, and controller settings), and vary only the depth source used for back-projection: raw sensor depth, D$^3$RoMa (RGB-D Stereo)~\cite{d3roma}, CDM, our restored depth, and our restored depth with uncertainty-guided grasp-score reweighting.

\begin{table}[!htbp]
\centering
\caption{\rui{Scene-level mean SR and CR with 95\% percentile-bootstrap confidence intervals for real-robot pick-and-place.
Results are computed over 20 scenes with 10,000 bootstrap resamples.}}
\label{tab:real_robot_ci_detail}
\resizebox{\linewidth}{!}{
\rui{
\begin{tabular}{lcc}
\toprule
Input Depth & SR (\%) $\uparrow$ & CR (\%) $\uparrow$ \\
\midrule
Raw
& 54.42 [45.33, 64.31]
& 88.00 [81.00, 94.00] \\
D$^3$RoMa (RGB-D-Stereo)
& 69.05 [62.36, 76.34]
& 96.00 [92.00, 99.00] \\
CDM (RGB-D)
& 72.73 [66.55, 78.69]
& 98.00 [95.00, 100.00] \\
Ours (RGB-D)
& \textbf{76.65 [71.57, 81.82]}
& \textbf{99.00 [97.00, 100.00]} \\
Ours + Unc-Reweighting (RGB-D)
& 70.25 [62.60, 77.35]
& 93.75 [88.25, 98.00] \\
\bottomrule
\end{tabular}
}
}
\end{table}

\begin{table}[!htbp]
\centering
\caption{\rui{Paired scene-level differences in real-robot pick-and-place. Differences are computed as \methodname\ minus each baseline on the same scene. 
We report mean differences in percentage points (pp) and 95\% percentile-bootstrap confidence intervals.}}
\label{tab:real_robot_paired_diff}
\resizebox{\linewidth}{!}{
\rui{
\begin{tabular}{lcc}
\toprule
Comparison
& $\Delta$SR (pp) $\uparrow$
& $\Delta$CR (pp) $\uparrow$ \\
\midrule
Ours (RGB-D) $-$ Raw
& +22.23 [12.00, 32.12]
& +11.00 [5.00, 18.00] \\
Ours (RGB-D) $-$ D$^3$RoMa (RGB-D-Stereo)
& +7.60 [-0.15, 14.97]
& +3.00 [-1.00, 7.00] \\
Ours (RGB-D) $-$ CDM (RGB-D)
& +3.92 [-3.09, 12.25]
& +1.00 [0.00, 3.00] \\
Ours (RGB-D) $-$ Ours + Unc-Reweighting (RGB-D)
& +6.40 [-3.05, 15.00]
& +5.25 [0.00, 11.25] \\
\bottomrule
\end{tabular}
}
}
\end{table}

\noindent\textbf{Scene construction.}
We conduct a controlled real-robot evaluation on the same 20 fixed cluttered scenes for all methods.
Each scene is sampled from the same 18-object pool, and the object set and object placement are matched as closely as possible across raw depth, D$^3$RoMa, CDM, \methodname, and \methodname\ with uncertainty-guided reweighting.
This protocol reduces the scene-construction variability of the previous independently sampled setting and allows scene-level paired comparisons across methods.
\rui{Each method was evaluated once per scene, yielding 20 clearing trials per method. Multiple grasp attempts within each trial were recorded separately.}

\noindent\textbf{Uncertainty-guided reweighting.}
We follow Sec.~\ref{sec:grasp_syn} in the main paper: for each grasp candidate, we project its center point to the image plane, sample the confidence map $\mathbf{C}(x,y)=1-\tilde{\mathbf{U}}(x,y)$, and multiply the original AnyGrasp score by this confidence before re-ranking.

\noindent\textbf{Execution and failure handling.}
The selected top-ranked grasp is executed with an open-loop approach-and-close motion followed by lifting and placing into a bin.
We consider an attempt successful if the target object is lifted off the table and placed without dropping.
\rui{
An object is marked as uncleared and manually removed once it accumulates three failed grasp attempts, which need not be consecutive.
When a grasp attempt is directed at the tray, the resulting failure is typically attributed to the target object nearest to the grasp point.
In rare cases, a single successful grasp removes multiple objects simultaneously. To avoid inflating completion statistics, we count such an event as one successful execution and reduce the effective target count by the number of additional objects removed.
Objects manually removed after failed attempts remain in the effective target count and are considered uncleared.
}

\begin{table*}[!tbp]
\centering
\caption{\rui{Per-scene grasp execution counts in the real-robot evaluation.}}
\label{tab:real_robot_scene_counts}
\begingroup
\footnotesize
\setlength{\tabcolsep}{4pt}
\renewcommand{\arraystretch}{1.10}
\rui{
\begin{tabular*}{\textwidth}{@{\extracolsep{\fill}}lccccc@{}}
\toprule
Scene & Raw & D$^3$RoMa & CDM & Ours & Ours + Unc. \\
\midrule
01 & 6/5/5  & 7/4/5  & 7/5/5  & 8/5/5 & 9/4/5 \\
02 & 11/3/5 & 8/5/5  & 6/5/5  & 6/5/5 & 6/5/5 \\
03 & 6/5/5  & 6/5/5  & 6/5/5  & 6/5/5 & 6/5/5 \\
04 & 12/4/5 & 8/5/5  & 6/5/5  & 7/5/5 & 8/4/5 \\
05 & 6/5/5  & 6/5/5  & 6/5/5  & 7/5/5 & 6/3/4 \\
06 & 10/3/5 & 8/4/5  & 8/5/5  & 7/5/5 & 6/5/5 \\
07 & 7/4/5  & 9/5/5  & 7/4/5  & 7/4/5 & 5/5/5 \\
08 & 11/5/5 & 6/5/5  & 6/5/5  & 9/5/5 & 6/5/5 \\
09 & 12/3/5 & 7/4/5  & 6/5/5  & 6/5/5 & 9/3/5 \\
10 & 7/5/5  & 5/5/5  & 12/5/5 & 5/5/5 & 6/5/5 \\
\addlinespace[2pt]
11 & 10/5/5 & 9/4/5  & 9/4/5  & 6/5/5 & 5/4/4 \\
12 & 9/4/5  & 5/5/5  & 7/5/5  & 7/5/5 & 6/5/5 \\
13 & 6/5/5  & 8/5/5  & 7/5/5  & 6/5/5 & 8/5/5 \\
14 & 10/5/5 & 10/5/5 & 6/5/5  & 6/5/5 & 6/5/5 \\
15 & 9/5/5  & 6/5/5  & 4/4/4  & 5/5/5 & 7/5/5 \\
16 & 5/5/5  & 7/5/5  & 8/5/5  & 6/5/5 & 7/5/5 \\
17 & 10/4/5 & 8/5/5  & 8/5/5  & 8/5/5 & 7/4/5 \\
18 & 10/4/5 & 7/4/4  & 7/5/5  & 7/5/5 & 6/5/5 \\
19 & 11/5/5 & 7/5/5  & 6/5/5  & 6/5/5 & 8/5/5 \\
20 & 10/4/5 & 6/5/5  & 7/5/5  & 7/5/5 & 9/5/5 \\
\bottomrule
\end{tabular*}
}
\par\vspace{2pt}
\noindent\begin{minipage}{\textwidth}
\raggedright\scriptsize
\rui{
\emph{Note:} Each entry is $A/S/N$: attempts ($A$),
successful executions and counted cleared objects ($S$),
and effective target count ($N$).
The per-scene rates are $\mathrm{SR}=S/A$ and
$\mathrm{CR}=S/N$.
``Ours + Unc.'' denotes uncertainty-guided reweighting.
}
\end{minipage}
\endgroup
\end{table*}

\noindent\textbf{Metrics and confidence intervals.}
We report Success Rate (SR) as the number of successful grasp executions divided by the total number of executed attempts.
\rui{
We report Completion Rate (CR) as the number of conservatively counted cleared objects divided by
the effective target-object count.
For scene $i$, let $A_i$, $S_i$, and $M_i$ denote the numbers of grasp attempts, successful executions,
and uncleared target objects, respectively.
The effective target count is $N_i=S_i+M_i$. 
Accordingly, the per-scene metrics are $\mathrm{SR}_i=S_i/A_i$ and $\mathrm{CR}_i=S_i/N_i$.
For each method, SR and CR are first computed per scene and then averaged equally over the same 20 matched scenes.
}

\rui{
We estimate 95\% confidence intervals using 10,000 scene-level percentile-bootstrap resamples with replacement (random seed 0). 
Each resample contains 20 scene indices.
The lower and upper bounds are the 2.5th and 97.5th percentiles of the bootstrap estimates.
For paired comparisons, we first compute the per-scene differences between our method and each baseline,
and then bootstrap these paired differences.
The main manuscript reports scene-level means with 95\% percentile-bootstrap confidence intervals, consistent with Table~\ref{tab:real_robot_ci_detail}.
Paired-difference confidence intervals are reported in Table~\ref{tab:real_robot_paired_diff}.
}

\rui{
For reproducibility, Table~\ref{tab:real_robot_scene_counts} reports the recorded per-scene attempt, success, and effective-target counts for all five methods, enabling independent reconstruction of SR, CR, and paired scene-level statistics.
}


\end{document}